\documentclass{lcs2}

\papergithub{https://github.com/PalGitts/SAFE-MEME}  

\usepackage{graphicx}
\usepackage{wrapfig}
\usepackage{pgf-pie}
\usepackage{todonotes}
\usepackage{setspace}
\usepackage{todonotes}
\usepackage{multirow}
\usepackage{svg}
\usepackage{amsmath}
\usepackage{blindtext}
\usepackage{graphicx}
\usepackage{wrapfig}
\usepackage[skip=2pt]{caption}
\usepackage{subcaption}
\usepackage{latexsym}
\usepackage{graphicx}
\usepackage{adjustbox}
\usepackage{soul}
\usepackage{xcolor}
\usepackage{color, colortbl}
\usepackage{tikz}
\usetikzlibrary{patterns}

\usepackage[leftcaption]{sidecap}
\usepackage[linesnumbered,ruled,vlined]{algorithm2e}
\usepackage{adjustbox}
\usepackage[most]{tcolorbox}
\tcbuselibrary{breakable}
\usepackage{subcaption}
\usepackage{pgfplots}
\pgfplotsset{compat=1.18} 

\newcommand{\revised}[1]{\textcolor{black}{#1}}

\definecolor{LightCyan}{rgb}{0.88,1,1}

\newcommand{\model}{\texttt{SAFE-MEME}}
\newcommand{\fvariant}{\model\texttt{-QA}}
\newcommand{\svariant}{\model\texttt{-H}}
\newcommand{\dataone}{\texttt{MHS}}
\newcommand{\datatwo}{\texttt{MHS-Con}}

\usepackage{enumitem}
\definecolor{turquoise}{RGB}{83, 95, 189}
\definecolor{orange_acc2}{RGB}{242, 93, 13}
\definecolor{good_yellow}{RGB}{190, 190, 0}
\definecolor{good_green}{RGB}{0, 100, 0}

\newtcolorbox{keytakeaway}[1][]{
  colback=gray!8,          
  colframe=gray!55,        
  boxrule=0.4pt,
  arc=2pt,
  left=6pt, right=6pt, top=5pt, bottom=5pt,
  breakable,               
  enhanced,
  fonttitle=\bfseries\small,
  coltitle=black,
  title={Key Takeaways#1},
  attach title to upper=\ \ 
}

\newtcolorbox{summOfFindings}[1][]{
  colback=gray!8,          
  colframe=gray!55,        
  boxrule=0.4pt,
  arc=2pt,
  left=6pt, right=6pt, top=5pt, bottom=5pt,
  breakable,               
  enhanced,
  fonttitle=\bfseries\small,
  coltitle=black,
  title={Summary of Findings#1},
  attach title to upper=\ \ 
}

\AtBeginDocument{
  \providecommand\BibTeX{{
    \normalfont B\kern-0.5em{\scshape i\kern-0.25em b}\kern-0.8em\TeX}}}

\papertitle{$\model$: Structured Reasoning Framework for Robust Hate Speech Detection in Memes} 

\papershorttitle{$\model$: Structured Reasoning Framework for Robust Hate Speech Detection in Memes}  

\papershortauthors{P. Nandi, S. Sharma, and T. Chakraborty} 
\paperauthors{%
  Palash Nandi
  \textsuperscript{\tiny 1} \quad
  Shivam Sharma
  \textsuperscript{\tiny 3} \quad
  Tanmoy Chakraborty \textsuperscript{\tiny {1,2}}%
} 

\paperaffil{\textsuperscript{1} Department of Electrical Engineering, Indian Institute of Technology Delhi \\
\textsuperscript{2} Yardi School of Artificial Intelligence, Indian Institute of Technology Delhi\\
\textsuperscript{3} Department of Computer Science, Aalto University,Finland} 
\paperemails{\{eez228472,tanchak\}@iitd.ac.in, shivam.sharma@aalto.fi} 

\paperabstract{ 
Memes act as cryptic tools for sharing sensitive ideas, often requiring contextual knowledge to interpret them correctly. It makes multimodal meme moderation difficult, as existing work either lacks high-quality datasets for nuanced hate categories or relies on low-quality social media visuals. Here, we curate two novel multimodal hate speech datasets comprising English memes -- $\dataone$ and $\datatwo$, which capture fine-grained hateful abstractions in \textit{regular} and \textit{confounding} scenarios, respectively. We benchmark these datasets against several competing baselines. Furthermore, we introduce \model\ (Structured reAsoning FramEwork) with its two variants: a novel multimodal Chain-of-Thought-based framework employing Q\&A-style reasoning ($\fvariant$) and a hierarchical categorization ($\svariant$) to enable robust hate speech detection in memes. $\fvariant$ outperforms the strongest open-source baseline model, showing an improvement of $2\%$ on $\dataone$ and $4.7\%$ on $\datatwo$ and closely follows the closed-source models, GPT-4o and Gemini 2.5. In contrast, $\svariant$ surpasses $\fvariant$, marking a $3\%$ improvement over the best open-source baseline, 
while achieving performance comparable to GPT-4o only on $\dataone$. We show that fine-tuning a single-layer adapter within SAFE-MEME-H outperforms fully fine-tuned models in regular fine-grained hateful meme detection. However, the fully fine-tuning approach with a Q\&A setup is more effective for handling confounding cases. We also systematically examine the error cases, offering valuable insights into the robustness and limitations of the proposed structured reasoning framework for analyzing hateful memes
\footnote{\textcolor{red}{CAUTION: Potentially sensitive content included; viewer discretion is requested.}}.
} 

\begin{document}

\makelabtitle


\section{Introduction}
\begin{figure}[!t]
\centering
\includegraphics[clip, trim={0 0 0 0}, width=1\textwidth]{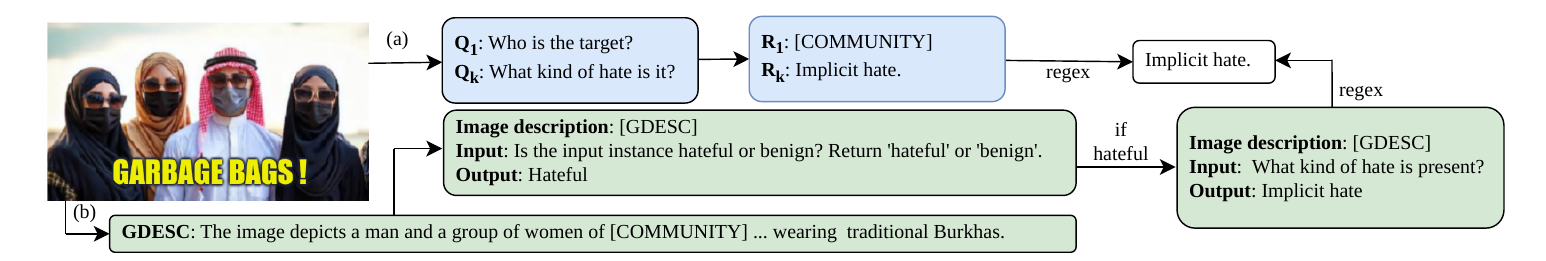} 
\caption{The demonstration of our proposed Chain-of-Thought-based structured reasoning framework for fine-grained hate speech detection in memes (\model) via (a) \textit{Q\&A-style} reasoning ($\fvariant$), and (b) \textit{Hierarchical} categorization ($\svariant$). Given a meme, the proposed variant $\fvariant$ sequentially generates a series of relevant questions-answers while the other proposed variant $\svariant$ opts for a two-level classification approach based on a detailed visual description (\textbf{[GDESC]}), before the final inference.}
\label{fig:MANTIS_exampleAandB}
\end{figure}

In today’s dynamic digital landscape, memes serve as powerful tools for expressing ideas and emotions but can also covertly propagate harmful content. Their blend of visual and textual elements presents unique challenges for analysis and detection of such hateful instances. A key difficulty lies in unraveling the complex factors that contribute to the computational assessment of hate content in memes \cite{cao2023proCap,cao2023prompting}. As illustrated in Figure \ref{fig:MANTIS_exampleAandB}, the visual and verbal cues in a given meme might appear harmless when considered in isolation; however, their combined interpretation may convey harmful intent to a specific community. It implicitly contains subtle elements like \textit{sarcasm} and \textit{irony} to encode hate speech. Therefore, it is important to systematically deconstruct memes and combine information from different semantic levels to understand their intention. Our findings indicate promising outcomes when Chain-of-Thought reasoning-based structured approaches are implemented via \textit{Q\&A-style} intermediate reasoning and hierarchical categorization. 

Existing Vision-Language Models (VLMs) face key limitations, including difficulties with reasoning over implicit content \cite{weng-etal-2026-whats}, multimodal cues integration \cite{DBLP:conf/nips/CampbellRGSG0KF24}, and comprehending out-of-domain cultural contexts \cite{nayak-etal-2024-benchmarking}. These shortcomings compel VLMs to often leverage surface-level features. As a result, they fail to adequately capture the implicit context embedded within the visual-linguistic interplay in memes. They also rely heavily on prompt engineering and often perform worse than specialized traditional classifiers on domain-specific tasks \cite{van2023detectingcorrectinghatespeech}. In response, we introduce a novel framework, $\model$ (Structured reAsoning FramEwork) with its two variants: (a) $\fvariant$, a multimodal Chain-of-Thought-based framework employing Q\&A-style reasoning and (b) $\svariant$, a hierarchical categorization-based framework towards hate speech identification in multimodal memetic context. In addition, we curate two novel datasets, $\dataone$ and $\datatwo$ in English, designed to capture complex hateful abstractions in standard and confounding meme formats. Overall, our contributions can be highlighted as \textit{five} key components\footnote{The source codes and datasets have been made public at \url{https://github.com/PalGitts/SAFE-MEME}.}:

\begin{enumerate}
    \item Introduction of two novel multimodal datasets: $\dataone$, and $\datatwo$ for benchmarking fine-grained hate speech detection and a challenging evaluation.
    \item $\model$, a multimodal framework for hate detection in memes, comprising two variants: a structured reasoning-based approach and a hierarchical categorization-based approach.
    
    \item A performance benchmark comparing $\fvariant$ and $\svariant$ against multiple baselines on both datasets.
    \item An ablation study of $\model$ in terms of intra-variant performance comparison.
    \item Provision of qualitative evaluation and detailed error analysis.
\end{enumerate}


\section{Related Work}
We present a brief overview of key studies on memetic harmfulness, visual question answering, multimodal LLMs, and their implications for meme analysis.   

\paragraph{\textbf{Literature on Hate Studies.}} Research on online hate has evolved from early lexicon and feature-based classifiers \cite{talat2016hateful, DBLP:conf/icwsm/DavidsonWMW17} to transformer-based fine-tuned models \cite{masud-etal-2024-probing} assessing these models' capability to understand and counter hate \cite{hee-etal-2024-recent, masud-etal-2024-hate}. Initially, a line of study explored user behavior on hate content, e.g., modeling the retweet dynamics of hateful content, gaining insights on hate diffusion on Twitter \cite{DBLP:conf/icde/MasudDMJGD021, goel2023hatemongers} and estimating the hate intensity \cite{sahnan2021better, DBLP:conf/kdd/DahiyaSSGCEMB021, DBLP:journals/kbs/MengSL023}. Extending this behavioral lens to multilingual context, \citet{sengupta2022does} highlighted the implicit relationship between five offense traits (aggression, hate, sarcasm, humor, and stance) in Hinglish (Hindi-English) social media code-mixed texts. On the mitigation front, \citet{chakraborty2022nipping} presented a study of methodological challenges for an automated hate mitigation system. \citet{DBLP:conf/kdd/MasudBKA022} addressed the task of hate mitigation in online posts through a three-stage pipeline that estimates hate intensity, identifies the hateful span, and paraphrases it to subdue the toxicity.

Implicit hate speech poses a significantly harder challenge as it hides overt slurs and explicit toxic cues \cite{DBLP:conf/emnlp/ElSheriefZMASCY21, DBLP:conf/naacl/WiegandRE21}. Unlike explicit hate, which can often be captured through lexical cues, implicit hate operates through pragmatic inference. That is, the hateful intent is conveyed not by what is said but by what is implied, causing surface-level classifiers to be particularly brittle in such a setup. To address this challenge, researchers have explored leveraging multiple external knowledge sources to generate implications of hate content \cite{sridhar-yang-2022-explaining, yadav-etal-2024-tox}.

\paragraph{\textbf{Literature on Memetic Studies}}  Various shared tasks on meme analysis focused on semantic roles labeling for entities \cite{Prakash_2023, sharma-etal-2022-findings}, affective characterization \cite{sharma-etal-2020-semeval}, metaphorical feature exploitation \cite{10.1145/3664647.3681060}, and topic learning \cite{10.1145/3581783.3613836}. Additional studies focused on issues such as anti-Semitism, propaganda, and harmfulness \cite{sharma-etal-2022-disarm, Prakash_2023}. Subsequently, \citet{sharma2022detecting} proposed a new classification of harmful memes, including categories such as self-harm \cite{Chen_Robinson_Conway_2025} and extremism \cite{valentini2020onlife}. However, their work did not cover other types of toxicity, such as disinformation, unethical sarcasm, and fake news. Later,  \citet{DBLP:journals/osnm/PandianiSC25} addressed the issue with a revised taxonomy of toxicity that features a uniform set of definitions and notations to represent the multi-dimensional nature of memes. They also pointed out that meme toxicity spans three attributes—target, intent, and tactics (based on rhetorical strategies), but these aspects are underexplored.  Other studies investigated multimodal corroboration and narrative-framing contextualization
\cite{Sharma_Agarwal_Suresh_Nakov_Akhtar_Chakraborty_2023}, affective assessment of memes with hate speech  \cite{cao-etal-2022-prompting}, and contextualization with a semantic repository (knowledge graph of memes) \cite{10.1145/3589335.3651970}. Another study on symbols in hate content highlighted that memes can carry harmful meaning through hidden cultural symbols and mismatched text and images \cite{tripathi2026they}. \citet{martinez2025ontoxkg} formalized the multimodal semantics of toxic symbols by modeling their textual and visual manifestations, capturing structured relationships among symbols, ideologies, and meanings. The corresponding knowledge graph, OnToxKG, includes toxic symbols that link to commonsense knowledge sources to provide a structured semantic representation.

The annotation schemas of the existing datasets have several shortcomings, including coarse-grained abstractions \cite{Prakash_2023}, specialized fine-grained labels like narrative roles \cite{sharma-etal-2022-findings}, protected categories and attack types \cite{mathias-etal-2021-findings}, a lack of nuanced contextualization beyond single-line statements \cite{Sharma_Agarwal_Suresh_Nakov_Akhtar_Chakraborty_2023, hee2023decoding}, and image quality \cite{gomez2019exploringhatespeechdetection} -- along with inconsistencies in how image–text pairs are configured (often appearing separately within social media posts or tweets), all of which together hinder systematic analysis of the multimodal nuances present in image–text memes.

\paragraph{\textbf{Visual Question Answering (VQA).}} 

Recent advancements in VQA tasks have increasingly favored the use of VLMs \cite{li2023blip, wang2024qwen2} over traditional VQA-specific architectures. This shift is largely attributed to the extensive pre-training of VLMs on large-scale image-text pair datasets \cite{chen2024far, hong2024cogvlm2}, which facilitates effective visual-textual alignment within a shared embedding space \cite{lin2024vila, wei2024efficient}. Notably, \citet{ravi2023vlc} introduced a Commonsense Transformer to inject contextual knowledge, while \citet{li2023joint} proposed a plug-and-play knowledge distillation framework that excelled in both answering questions and inferring rationales. In a complementary direction, \citet{ghosal-etal-2023-language} explored leveraging language models to generate rationales for image captions and scene graphs. These developments underscore the growing capability of multimodal LLMs to reason within joint representation spaces. Nonetheless, recent studies \cite{agarwal-etal-2024-mememqa} have identified their limitations when applied to complex, culturally nuanced content such as memes, where coherent multimodal reasoning remains a challenge.

\begin{figure}[t!]
\centering
\includegraphics[clip, trim={0cm 0cm 0cm 0cm},width=1\textwidth]{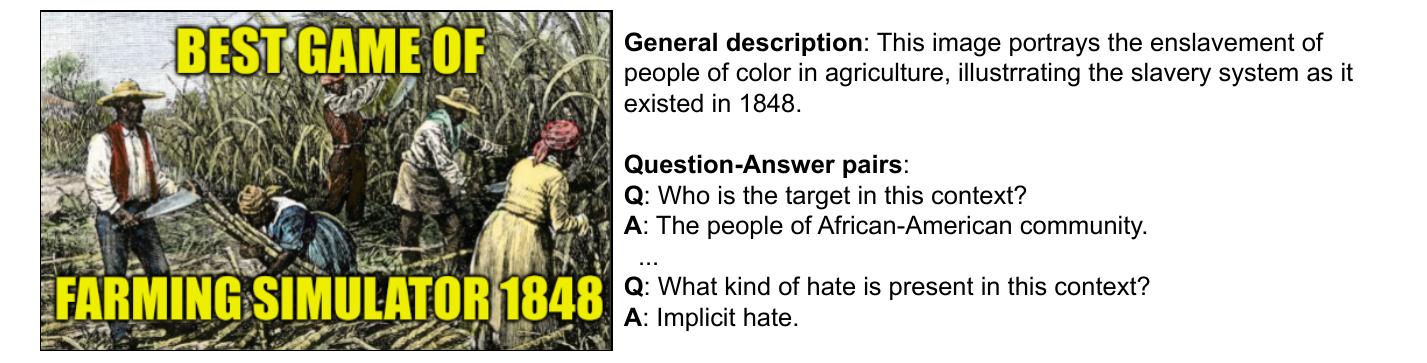}
\caption{An illustration of annotations in the $\dataone$ dataset. Each instance is associated with a hatefulness label -- explicit, implicit, or benign, along with augmented information like \textit{general description} and \textit{Q\&A}. The \textit{general description}  describes important attributes of the entities as well as the inter-entity relationships. Additionally, the context is expressed as a sequence of question and answer pairs in \textit{Q\&A} component.}
\label{fig:MHS_DBExample}
\end{figure}

\paragraph{\textbf{Large Vision Language Models}} The emergence of \text{LLMs} like LLaMA \cite{touvron2023llama}, ChatGPT \cite{chatgpt2022}, Vicuna \cite{vicuna2023}, Bard \cite{bard2023} and GPT-4 \cite{openai2023gpt4} have driven significant progress in NLU-based reasoning, with a growing focus on multimodal augmentation in visual-linguistic tasks. Fusion-based \textit{adapter} layers enhance LLMs' capabilities in multimodal applications like Q\&A and conversational agents \cite{liu2023visual,awadalla2023openflamingo}. Current large vision-language models, including miniGPT-4 \cite{zhu2023minigpt4}, LLaVA \cite{liu2023visual}, and multimodalGPT \cite{gong2023multimodalgpt}, struggle to interpret nuances such as sarcasm and irony within the visual-linguistic incongruity of memes, and perform well only when the content is explicitly conveyed.  Although some recent studies address meme-related tasks \cite{hwang2023memecap}, they focus mainly on \textit{multimodal grounding}-based captioning and Q\&A setups. In complex scenarios, these models face hallucinations. Section \ref{app:limitations_on_existingResources} presents additional details on limitations.

The multimodal modeling and reasoning capacity offered by recent multimodal LLMs is very strong; however, they fall short on tasks involving fine-grained question answering and generating sufficiently detailed explanations. In this work, we strive to enhance the utility of existing multimodal neural frameworks while leveraging the reasoning capabilities of LLMs (and their multimodal counterparts). Our study seeks to enhance the ability of a model to detect hateful content in memes by deconstructing their multifaceted nature. We address these challenges by improving visual-semantic reasoning and generating rich visual information for subsequent stages.

\section{Datasets}

This section outlines the characteristics, curation, and statistics of the datasets used in this study. Existing hate classification datasets often suffer from low-quality, heterogeneous multimedia content \cite{gomez2019exploringhatespeechdetection} (which is arguably representative of the real-world distribution) or offer only coarse-grained labels \cite{kiela_hateful_2020}. In contrast, \citet{mathias-etal-2021-findings} re-annotated the Hateful Memes dataset \cite{kiela_hateful_2020} with binary labels -- `Hateful' or `Not Hateful' -- and added fine-grained tags for protected categories (e.g., race, gender) and attack types (e.g., slurs, dehumanization). \citet{sharma-etal-2022-findings} further introduced HVVMemes, annotating referenced entities as hero, villain, victim, or other. Our proposed dataset, \dataone, differs from the existing ones by providing a two-fold annotation: (i) labeling each meme as explicit hate, implicit hate, or benign, and (ii) supplying question-answer pairs that explain the assigned label and offer contextual background. For example, referring to someone as ``old farm equipment'' may be flagged as implicit hate, with its QA pairs revealing its offensive cultural undertones. 

\begin{figure}[t!]
\centering
\includegraphics[clip, trim={0cm 0cm 1cm 0cm},width=1\textwidth]{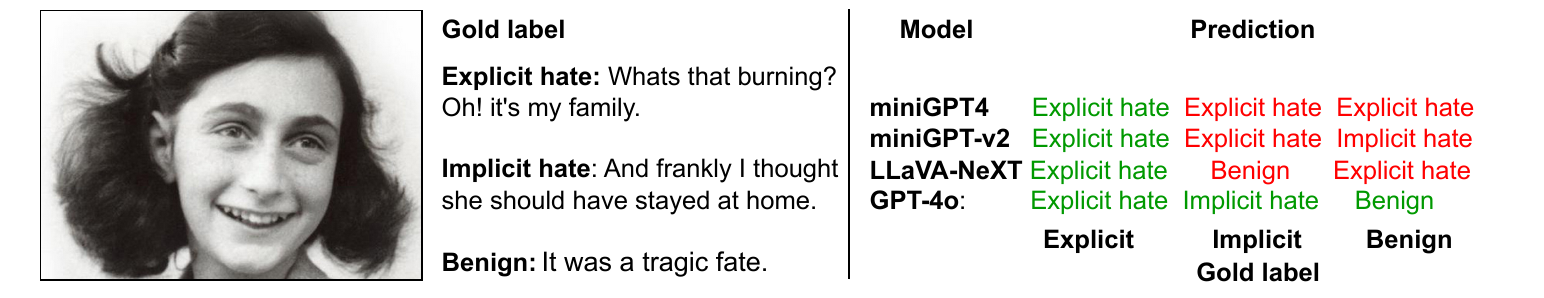} 
\caption{An illustration of instances in the $\datatwo$ dataset. Each unique visual content is associated with three distinct variations of textual hate content: (a) explicit hate, (b) implicit hate, and (c) benign. Notice the given visual instance, when paired with text (a), (b), and (c), emerges as instances of explicit, implicit, and benign cases of hate speech, respectively, in memes. Also shown is a comparison of the category prediction by a closed-source (GPT4-o) and three open-source (miniGPT4,  miniGPT-v2 and LLaVA-NeXT) large VLMs.}
\label{fig:DBCon_example}
\end{figure}

An essential aspect of this work is evaluating model robustness under dynamic multimodal configurations across three fine-grained hate categories, necessitating controlled benchmarking, which is challenging to achieve with free-form, realistic datasets that lack consistent image-text associations. To navigate these constraints, we introduce two new datasets -- $\dataone$, designed explicitly for benchmarking competitive unimodal and multimodal models, and $\datatwo$, exclusively intended to stress-test a given model. Each data instance in the $\dataone$ dataset is categorized with a corresponding hate label -- explicit, implicit, or benign -- based on its characteristics. Additionally, the dataset includes supplementary information, such as \textit{Q\&A details} and a \textit{general description}, where relevant, to improve contextual understanding and applicability (cf. Figure \ref{fig:MHS_DBExample}). In contrast, the main objective $\datatwo$ is to develop a complex stress test, reflecting the nuances required for memetic analysis. Each instance has \textit{three} paraphrased versions corresponding to the explicit, implicit, or benign forms of hate, conditioned on the same visual but different textual content (cf. Figure \ref{fig:DBCon_example}). Here, we also demonstrate the limitations of existing large VLMs, whether closed or open source. As shown in Figure \ref{fig:DBCon_example}, GPT-4o (a closed-source model) correctly labels all three variations. In contrast, open-source models struggle to identify correct forms of hate (or the absence thereof), thereby providing a fair evaluation framework.

\subsection{Dataset Curation}
In this section, we present a comprehensive description of the construction of the $\dataone$ and $\datatwo$ datasets. We begin by describing the data collection process, followed by the annotation protocol and a statistical characterization of both datasets.

\begin{figure}[t!]
\centering
\includegraphics[width=1\textwidth]{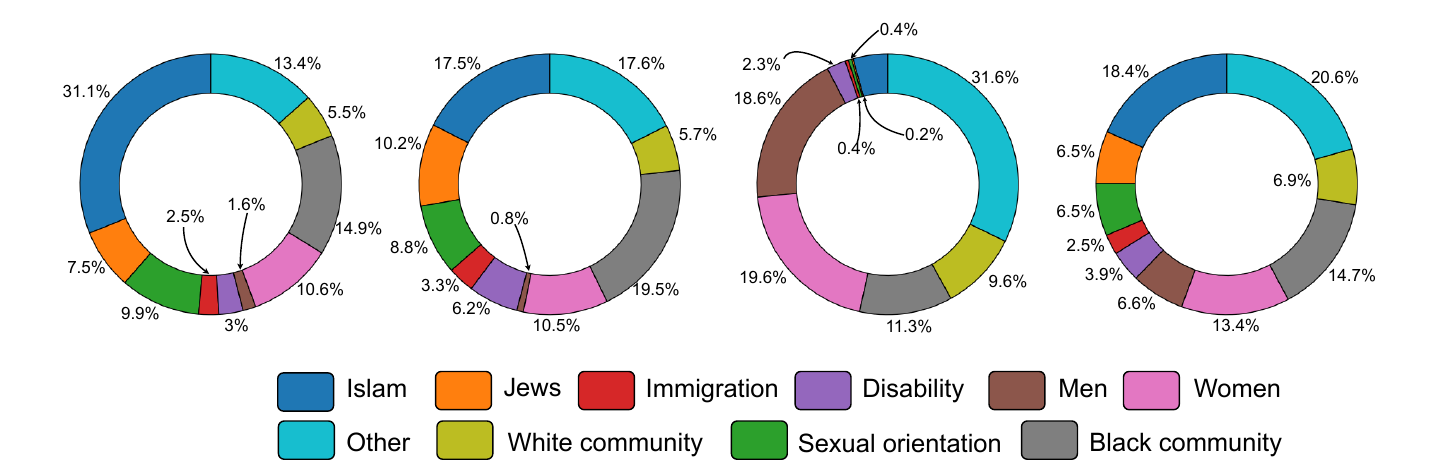} 
\caption{An illustration of the counts (expressed as a percentage) of the various protected groups in the $\dataone$ dataset. The distribution of the protected groups is presented in \textit{four} distinct setups when considering the instances of (a) explicit, (b) implicit, (c) benign and (d) all categories together.}
\label{fig:RMMHS_hate_distribution}
\end{figure}

\subsection*{Data Collection and Deduplication}

We collected memes for the $\dataone$ dataset using \textit{Google Images search}. Initially, we leveraged keywords such as \textit{hateful memes, toxic memes, and dark-humor memes} to collect \textit{harmful} memes. After observing that hateful content predominantly targets protected groups and historical events, we opted for a focused search with particular hateful keywords such as `Islamophobic Memes', `Holocaust Trivialization Memes', or `Memes featuring people of diverse ethnic backgrounds', etc. In contrast, generic keywords, for instance, `funny memes', `hilarious memes', `trending memes', and `viral memes' were utilized to acquire benign memes. We used a {web extension}\footnote{\url{https://download-all-images.mobilefirst.me}} to collect memes in batches. We further web-scraped platforms (\href{https://www.instagram.com/}{Instagram} and \href{https://www.reddit.com/}{Reddit}) to expand our meme collection. We deduplicated\footnote{\url{http://github.com/arsenetar/dupeguru}} the collection and filtered out low-resolution meme images, with a minimum dimension of $220\times319$. We identified \textit{nine} categories of protected groups in the  $\dataone$ dataset (cf. Table \ref{tab:hate_category}). Corresponding keywords were collected from observations made during the preliminary annotation phases. Figure \ref{fig:RMMHS_hate_distribution} illustrates the distribution of hate by categories for four different setups. As observed, hate against Black people and women outnumbers the remaining categories by a large margin. Upon closer inspection, it is noticeable that hate against the Black community, the Jewish community, and women is prominently represented in an implicit form; it implies that hate against the African American and Jewish communities is latent. In contrast, hate targeting Muslims is more explicit. The \dataone~dataset consists of $3,343$ instances, including $795$ explicit, $753$ implicit, and $685$ benign memes, with $2,233$ memes used for training, $305$ for validation, and $805$ for testing. We maintain the exact definition of hate speech to annotate and collect $409$ explicit, $420$ implicit, and $409$ benign memes -- a total of $1,238$ memes in the $\datatwo$ dataset.


The primary motivation behind the creation of $\datatwo$ is to provide a rigorous stress test for fine-grained multimodal hate-speech detection models for memes. Each instance has \textit{three} paraphrased versions of hate conditioned on the same visual content (cf. Figure \ref{fig:DBCon_example}). Initially, approximately 70 distinct memes per hate category were randomly selected from $\dataone$. We then fed the available text from each selected meme into \texttt{zephyr-7b-alpha} as a k-shot prompt for paraphrasing. We then gathered relevant visual content from the web and paired it with each paraphrased text using Canva\footnote{\url{https://www.canva.com}}. Note that $\dataone$ and $\datatwo$ contain no user information. All memes in our datasets were sourced from publicly accessible web pages, with no known copyright issues preventing their use for research purposes.

\begin{figure}[t!]
\centering
\includegraphics[width=1\textwidth]{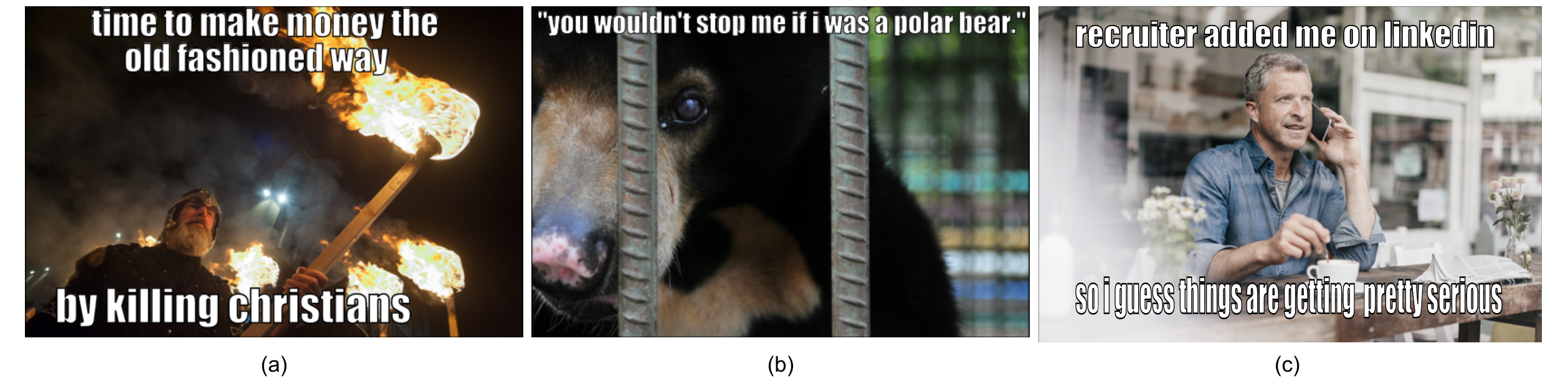} 
\caption{A demonstration of examples from each hate category: (a) Explicit hate — the text directly promotes violence against the Christians, (b) Implicit hate — the meme lacks explicit hateful cues; however, it implicitly targets the Black community, (c) Benign — the content is neutral and devoid of any explicit or implicit hateful content.}
\label{fig:category_wise_examples}
\end{figure}

\subsection*{Annotation Guidelines}
In general, the content of a meme is labeled as `explicit' if it contains hateful words or phrases. Most memes classified under the `implicit' hate category do not overtly exhibit hate on the surface, although it can be recognized through careful analysis and interpretation with relevant background knowledge. Lastly, a meme is labeled as ``benign'' (or \textit{non-hate}) if it exhibits no hostility, implicit or explicit, toward any protected group (cf. Figure \ref{fig:category_wise_examples}). In addition, we aim to provide a comprehensive description of the image, which contains intricate visual information and a set of relevant question-answer pairs essential for exploring the context. During the annotation process, we observed that pre-trained models frequently overlook critical details about the entities. For example, when \texttt{lava-v1.5-13b} was prompted to describe the image presented in Figure \ref{fig:MHS_DBExample}, it generated the following output: \textit{This image depicts a group of individuals engaged in agricultural activities along the riverside}. However, it ignored several critical visual elements essential to understand the context of the meme. At this point, we manually refined and integrated critical details into the previously generated description, resulting in the following revised narrative: \textit{The image depicts a scene of forced labor in an agricultural field, highlighting the oppression of individuals of color}. \footnote{See Section $\ref{app:dataset_construction}$ for additional details on dataset curation and annotation procedures.}


\section{Proposed Methodology: \model}

Both variants of $\model$ follow a two-stage sequential architecture and accept a meme $M$ comprising an image $I$ and an overlaid text $T$ as input to generate a textual label $L$ that indicates whether the meme contains hate content. Inspired by multimodal Chain-of-Thought (MM-CoT) reasoning \cite{zhang2023multimodal}, the first variant, $\fvariant$, incorporates a structured reasoning approach following a two-stage implementation of CoT \cite{agarwal-etal-2024-mememqa}: (i) generation of relevant queries and (ii) answer inference. In the first stage, $\fvariant$ generates a set of context-based queries that serve as inputs in the next stage to obtain a set of contextually relevant responses. In contrast, the second variant, $\svariant$, adopts a hierarchical approach. It first generates a fine-grained visual description that constitutes salient attributes of visual subjects, followed by hierarchical classification.

\begin{figure}[t]
\centering
\includegraphics[width=1\columnwidth]{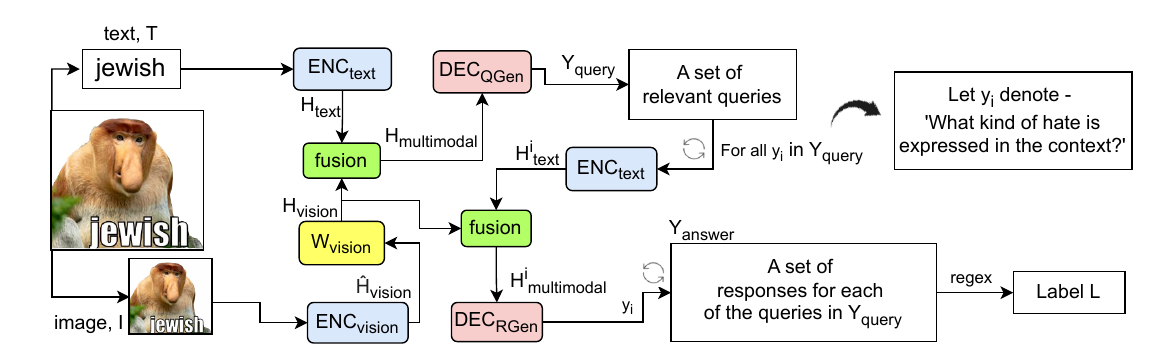} 
\caption{Architectural details of $\fvariant$. Phase 1: Encoders -- textual (\texttt{T5-base}) and visual (\texttt{ViT}) generate a multimodal signal via gated fusion, serving as input to the \texttt{T5-base} decoder (DEC$_{QGen}$) to generate questions, Y$_{query}$. Phase 2: A fused multimodal signal is fed to another \texttt{T5-base} decoder (DEC$_{RGen}$) for each question y$_{i} \in$ Y$_{query}$ to generate responses, Y$_{answer}$. The hate label $L$ of meme $M$ is identified via a regular expression.}
\label{fig:The architecture of MANTIS_NH}
\end{figure}

\subsection{Architecture of \fvariant}

As shown in Figure \ref{fig:The architecture of MANTIS_NH}, the two-stage pipeline for \fvariant\ is characterized by the format and type of the inputs and outputs at each stage. In the first stage, the model takes only the meme ($M$) as an input and generates a set of queries, $Y_{query}$. In the second stage, the model uses the meme ($M$) along with each query ($y_i \in Y_{query}$ ) as input to produce the corresponding set of responses, represented as ($Y_{answer}$). Below, we describe each of these modular steps in detail.

\begin{flushleft}\textbf{Stage 1 - Relevant Question Generation}\end{flushleft} The first stage of \fvariant\ requires the textual and visual information pair, i.e., $I$ and $T$ as inputs and outputs a series of relevant questions $Y_{query}$ in order to \textit{deconstruct} the typically obscure contextual information within memes. For instance, $Y_{queries}$ w.r.t. Figure \ref{fig:The architecture of MANTIS_NH} can be interpreted as follows: a) \textit{What could be the targeted group in this context?} b) \textit{What stereotype or association is implied in this context?} along with other related aspects. The textual input $T$ and the vision input $I$ are transformed into features using a text encoder and a vision encoder, respectively. We use the encoder part of \texttt{T5-base} \cite{raffel2020exploring} as the text encoder, ENC$_{text}$ to encode $T$ (cf. Eq. \ref{eq_visionEncAndProjection}). It returns the hidden states of the last layer as encoded textual features, $H_{text} \in \mathbb{R}^{n \times d}$  where $n$ is the token length and $d$ is the dimension of the hidden states of the \texttt{T5-base} encoder. 
The vision transformer, \texttt{ViT} \cite{dosovitskiy2020image} (ENC$_{vision}$), is employed to encode the patch-level visual information, $H_{vision}  \in \mathbb{R}^{p\times d}$, where $p$ is the number of patches and $d$ is the dimension of the patched embeddings. Subsequently, $H_{vision}$ is linearly projected from the previous dimension $d$ to $\hat{d}$, denoted by $\hat{H}_{vision} \in \mathbb{R}^{p\times\hat{d}}$ (cf. Eq. \ref{eq_visionEncAndProjection}) using a trainable weight $W_{vision}$ to fuse it with $H_{text}$:

\begin{equation}\label{eq_visionEncAndProjection}
    H_{text} = \text{ENC}_{text}(T) \text{,} \quad H_{vision} = \text{ENC}_{vision}(I) \quad \text{and} \quad \hat{H}_{vision} = W_{vision} \cdot H_{vision}
\end{equation}

The correlation between the textual and visual information is obtained in the form of textually conditioned visual features ($H_{attn}$) through a single-headed cross-attention \cite{vaswani2017attention}, where query ($Q$) is conditioned on $H_{text}$, and key ($K$). The value ($V$) are conditioned on $H_{vision}$. Finally, the fused representation, $H_{multimodal}$, is obtained with a gated fusion mechanism \cite{zhang2023multimodal} based on $H_{text}$ and $H_{attn}$ (cf. Eq. \ref{eq_gatedFusion}):
\begin{equation}\label{eq_gatedFusion}
    \lambda = \sigma(W^{fusion}_{text} \cdot H_{text} + W^{fusin}_{vision} \cdot H_{attn}) \quad \text{and} \quad H_{multimodal} = \lambda \cdot H_{attn} + (1- \lambda ) \cdot H_{text}    
\end{equation}
$H_{multimodal}$ is then fed to the decoder of \texttt{T5-base}, $DEC_{QGen}$ that is fine-tuned to generate relevant queries, $y_{i}\in Y_{query}$. We obtain the results by computing the probability of $Y_{query}$ conditioned on $T$ and $I$,
\begin{equation}\label{eq:poutput}
    \begin{gathered}
        P(Y_{query}| \text{ T, I }) = \prod^{N}_{i=1} P(y_{i}|\text{ T, I and } Y_{< i} )
    \end{gathered}
\end{equation}

\begin{flushleft}\textbf{Stage 2 -- Response Generation}\end{flushleft} The second stage of the pipeline follows a similar setup and aims to generate answers iteratively, $y^{i}_{answer}$, for each of the questions $y_{i}\in Y_{query}$ w.r.t. $M$. \revised{For example, w.r.t. Figure \ref{fig:The architecture of MANTIS_NH} and $Y_{query}$ in the previous phase (cf. Stage 1), it can be interpreted as follows: a) \textit{The Jewish community}, b) \textit{The facial attributes}, etc}. Each question $y_i\in Y_{query}$ is encoded into a textual hidden representation, $H^{i}_{text}$, using the encoder of \texttt{T5-base} (cf. eq. \ref{eq_visionEncAndProjection}). 
$H^{i}_{text}$ is then fused with the visual information, $H_{vision}$ (see Eq. \ref{eq_gatedFusion}) in order to generate a single-head, cross-attention-based representation, $H^{i}_{attn}$, where $H^{i}_{text}$ is used to condition the query ($\hat{Q}$)
and $H_{vision}$ conditions the key and value, as in the previous stage.
Finally, the fused multimodal state, $\hat{H}^{i}_{multimodal}$ is formed w.r.t. each question $y_{i}$ and $H_{vision}$ (Eq. \ref{eq:responseGatedFusion}). Notice that the multimodal fused representation $\hat{H}^{i}_{multimodal}$ shares $H_{vision}$ across the question-set $Y_{query}$ but differs for each input question, $y_{i}$.
\begin{equation}\label{eq:responseGatedFusion}
\begin{aligned}
    H^{i}_{text} = \text{ENC}_{text}(y_{i})& \\
    \gamma = \sigma(W^{fusion}_{text} \cdot H^{i}_{text} + W^{fusion}_{vision} \cdot H^{i}_{attn}) \quad \text{and} \quad  & 
    \hat{H}^{i}_{multimodal} = \gamma \cdot H^{i}_{attn} + (1- \gamma) \cdot H^{i}_{text}
\end{aligned}
\end{equation}
Finally, $\hat{H}^{i}_{multimodal}$ is fed to the decoder of \texttt{T5-base}, $DEC_{RGen}$, and fine-tuned to generate the answer $y^{i}_{answer}$ for the query $y_{i}\in Y_{query}$. We obtain $y^{i}_{answer}$ by generating the likely label sequence, $y^{i}_{answer}$, conditioned on $M$ i.e., $I$ and $T$, as demonstrated in Eq. \ref{eq:poutput} (see Section $\ref{app:safeMemeQA}$ for example-based demonstration of steps in \fvariant). 

\subsection{Architecture of \svariant}

\svariant\ aims to generate a classification label utilizing a two-stage pipeline (cf. Figure \ref{fig:The architecture of MANTIS_H}). The characteristics of the stages are described in terms of the format, input types, output types, and corresponding objectives: (i) the first stage attempts to generate a rich visual description ($Y_{gdesc}$) of the input meme involving important contextual cues, and (ii) the second stage performs hierarchical inference based on the $Y_{gdesc}$ from the previous stage, $T$ and $I$ of the input meme, $M$. See Section $\ref{app:safeMemeH}$ for an example-based demonstration of steps in $\svariant$.    

\subsubsection*{Stage 1 - Contextual Description Generation}
\begin{figure}[t]
\centering
\includegraphics[width=1\columnwidth]{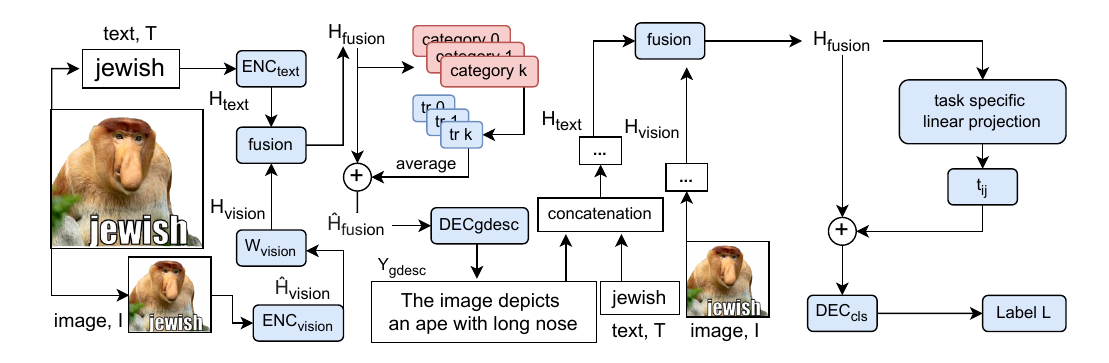}

\caption{Architectural details of $\svariant$. Phase 1: Encoders - textual (\texttt{T5-base}) and visual (\texttt{ViT}) generate a multimodal signal via gated fusion, which further incorporates information from category-specific representations, which serves as input to the \texttt{T5-base} decoder ($DEC_{gdesc}$) to avail a rich contextual description, $Y_{gdesc}$. Phase 2: A similar approach derives a multimodal signal via gated fusion, which collects additional information from  $Y_{gdesc}$. Later, it serves as an input to the \texttt{T5-base} decoder ($DEC_{cls}$) to generate a label (based on the task-specific linear projector). In the first level of the hierarchy, the labels uncover whether a meme is hateful or benign, while in the later stages, it distinguishes between implicit and explicit hate. }
\label{fig:The architecture of MANTIS_H}
\end{figure}

$\svariant$ requires the textual and visual information pair, i.e., $T$ and $I$ as input and outputs a rich visual description ($Y_{gdesc}$) containing fine-grained details of the context. \revised{For example, based on Figure \ref{fig:The architecture of MANTIS_H}, the corresponding $Y_{gdesc}$ can be as follows: \textit{“The image shows an ape with a long nose … along with the overlaid text ‘jewish’”}}. The textual input $T$ and the vision input $I$ are transformed into features using a text encoder (ENC$_{text}$) and a vision encoder (ENC$_{vision}$), respectively (cf. Eq. \ref{eq_gatedFusion}) to encode the patch-level visual information, $H_{vision}$, followed by a linear projection (cf. Eq. \ref{eq_visionEncAndProjection}). The correlation between the textual and visual information is obtained through a single-headed cross-attention, which produces a fused representation, $H_{fusion}$,  with a gated fusion mechanism (cf. Eq. \ref{eq_gatedFusion}). To generate a rich contextual description, the fused representation $H_{fusion}$ is further combined with target category-specific representations (cf. Eq. \ref{eq:fusionCategoryCard}): 

\begin{align}\label{eq:fusionCategoryCard}
    \hat{\text{H}}_{fusion}^{i} &= \text{Proj}_{gdesc}^{i}(\text{i}, H_{fusion}) \quad \text{and} \quad \text{H}_{fusion} = \text{H}_{fusion} + \sum_{i=0}^{C} \hat{\text{H}}_{fusion}^{i} / |C|
\end{align}
In this context, $C$ represents the set of all target groups, Proj$_{gdesc}^{i}$ is a target category-specific linear projector that linearly transforms $H_{fusion}$ keeping the dimension intact (see Section \ref{sec:ablationStudyofSAFE_MEME} for additional details on the linear projector). Following this, ${H}_{fusion}$ is fed to the decoder part of the \texttt{T5-base}, $DEC_{gdesc}$ and fine-tuned to generate the contextual description $y_{gdesc}$. We obtain the outcome by generating the likely text, $Y_{gdesc}$, conditioned on $T$ and $I$:
\begin{equation}\label{eq:poutput_1}
    \begin{gathered}
        P(Y_{gdesc}| \text{ T, I }) = \prod^{N}_{i=1} P(y_{i}|\text{ T, I and } Y_{< i} )
    \end{gathered}
\end{equation}

\subsubsection*{Stage 2 - Hierarchical Classification}

The final stage of $\svariant$ requires $Y_{gdesc}$ and the image-text pair, i.e., $I$ and $T$, as inputs and outputs a label $L$. The first level of the hierarchy is responsible for detecting whether there is hate in the context. If hate is detected, the second stage further categorizes it as implicit or explicit. Following the previous stage, we obtain $H_{fusion}$ with exception that the ENC$_{text}$ now receives the concatenation of $y_{gdesc}$ and $I$ as input. Although, $H_{fusion}$ is further integrated with task-specific information.

\begin{equation}\label{eq:taskSpecificFusion}
    \begin{gathered}
        \text{H}_{fusion} = \text{H}_{fusion} + \text{Proj}_{task}(\text{H}_{fusion})\\
    \end{gathered}
\end{equation}
Here, Proj$_{task}(.)$ is a task-specific linear projector that generates task-specific latent information while maintaining the same dimension. Finally, ${H}_{fusion}$ is fed to the decoder of \texttt{T5-base}, $DEC_{cls}$ and fine-tuned to generate the required label $L$. We obtain the label by generating text ($L$) conditioned on $Y_{gdesc}$, $T$ and $I$:  
\begin{equation}\label{eq:poutput_2}
    \begin{gathered}
        P(L| Y_{gdesc}\text{, T, I }) = \prod^{N}_{i=1} P(L_{i}|\text{ T, I and } Y_{< i} )
    \end{gathered}
\end{equation}


\section{Experimental Setup}
To establish a robust framework for hate content classification in memes, we present our experimental setup, detailing baseline models, ablation studies, and evaluation metrics to assess performance across explicit, implicit, and benign categories.

\subsection{Baselines}
In this work, we systematically utilize a broad spectrum of unimodal and multimodal models to classify memes into three categories: explicit hate, implicit hate, and benign. While the unimodal baselines enable us to quantify the discriminative power of the text and image modalities independently,  the multimodal baselines are formulated to capture the cross-modal interactions of a meme. The details are as follows:

\noindent \paragraph{\textbf{Unimodal Baselines}.} \texttt{T5-large} \cite{raffel2020exploring}, \texttt{ALBERT} \cite{lan2019albert}, and \texttt{BiLSTM} \cite{cornegruta-etal-2016-modelling} to obtain text-based representation of input meme, $M$.

\noindent \paragraph{\textbf{Multimodal Baselines}.} \texttt{RoBERTa} \cite{liu2019roberta} with \texttt{ResNet-152} \cite{he2016deepResnet}, zero-shot prompting on \texttt{BLIP2} \cite{li2023blip}, \texttt{CLIP} \cite{radford2021Clip}, \revised{\texttt{CLIP-XLM-R}}\footnote{\url{https://huggingface.co/M-CLIP/XLM-Roberta-Large-Vit-L-14}}, \texttt{LLaVA} \cite{liu2023llava}, \texttt{LLaVA-Next} \cite{liu2024llavanext}, \texttt{MiniGPT-v2} \cite{chen2023minigptV2}, \revised{\texttt{Qwen-VL}} \cite{bai2023qwen}, Multimodal Chain-of-Thought (\texttt{MM-CoT}) \cite{zhang2023multimodal}, \texttt{Minigpt-4} \cite{zhu2023minigpt4}, \revised{\texttt{GPT-4o}} \cite{achiam2023gpt4},  and \revised{\texttt{Gemini 2.5}} \cite{team2023gemini, comanici2025gemini}.  \texttt{BLIP2} and \texttt{CLIP} are also fine-tuned on \dataone~and employed as multimodal baselines.

\noindent \paragraph{\textbf{ Task Specific Baselines}.} \texttt{\revised{Hate-clipper} (HClipper)} \cite{kumar2022hate}, \texttt{Pro-Cap} \cite{cao2023proCap}, \texttt{CapAlign} \cite{ji2024capalign} and \texttt{PromptHate} \cite{cao-etal-2022-prompting}. 

\subsection{Modular Variations in Proposed Model Architectures} \label{sec:ablationStudyofSAFE_MEME}
We conduct a thorough ablation on the constituent modules to explore the modular capabilities of $\fvariant$ and $\svariant$. To this end, we examine the individual modules for each variant. Each module uses a fully or partially fine-tuned version of the \texttt{T5-base} model, utilizing a combination of annotated information of the $\dataone$ dataset.

\begin{table}[!t]
  \centering
  \small
  \caption{\revised{An overview of unit modules, their inputs and outputs, and corresponding functionalities for both variants of $\model$. Note: \textbf{E}: Explicit hate, \textbf{I}: Implicit hate, and \textbf{B}: Benign}.}
  \label{unitMudulesOf_SM}
  {\color{black} 
  \resizebox{0.8\columnwidth}{!}{%
  \begin{tabular}{|c|c|c|l|}
    \toprule
    \textbf{Name} & \textbf{Input} & \textbf{Output} & \textbf{Functionality}  \\
    \midrule  

    $DGen$ & $I$, and $T$ & $y_{gdesc}$ & It generates a  visual description, $y_{gdesc}$ of $M$ \\

    $CG_{i=0|1}$ & $I$, $T$ (and $y_{gdesc}$ if $i=1$) & \multirow{1}{*}{$y_{context}$} & 
    \multirow{1}{*}{It generates a detailed context, $y_{context}$ of $M$} \\
    

    $QAG_{0|1}$ & $I$, $T$ (and $y_{gdesc}$ if $i=1$) & \multirow{1}{*}{$Y_{qa}$} & 
    \multirow{1}{*}{It generates a set of QA pairs, $Y_{qa}$} \\
    

    $QG_{0|1}$ & $I$, $T$ (and $y_{gdesc}$ if $i=1$) & \multirow{1}{*}{$Y_{query}$} & 
    \multirow{1}{*}{It generates a set of queries, $y_{query}$} \\
    
    $RG_{0|1}$ & $I$, $T$ (and $y_{gdesc}$ if $i=1$) &  \multirow{1}{*}{$Y_{answer}$} & 
     \multirow{1}{*}{It generates a response set $Y_{answer}$} \\

    $CLS$ & $y_{answer}$ / $y_{context}$ & E/I/B & Classifier of $\fvariant$\\

    $HCls_{0}$ & I, T, $y_{gdesc}$ & B/H & First level classifier of $\svariant$\\
    $HCls_{1}$ & I, T, $y_{gdesc}$ & E/I & Second level classifier of $\svariant$\\
    \bottomrule
  \end{tabular}%
  }}
\end{table}

\begin{table}[!t]
  \centering
  \caption{ Configuration of $\model$ variants and their component combinations. Each variant of $\fvariant$ and $\svariant$ integrates different unit modules. Each of the unit modules is either fully or partially fine-tuned with specific objectives. Please refer to Section \ref{sec:ablationStudyofSAFE_MEME} for more details.}
  \label{variantsOf_MANTIS}
  \resizebox{1\columnwidth}{!}{%
  \begin{tabular}{|cc|c|cc|c|}
    \toprule
    \multicolumn{2}{|c|}{\textbf{Variant}}  & \textbf{Combination} & \multicolumn{2}{|c|}{\textbf{Variant}} & \textbf{Combination} \\
    \midrule  
    \multirow{8}{*}{\rotatebox{90}{$\fvariant$}}&M0 & $DGen$ + $QAG_1$ + $CLS_{regex}$ & \multirow{8}{*}{\rotatebox{90}{$\svariant$}} &&\\
    &M1 & $DGen$  + $QG_1$ + $RG_1$ + $CLS_{regex}$ & & M8 & T5-base (full ft.) + $DGen$ + $HCls_0$ + $HCls_1$\\
    &M2 & $DGen$ + $QG_1$ + $RG_1$ + $CLS_{ft}$ & & M9 & gLP with $v_{0}$ + T5-base (full ft.) for $DGen$ + $HCls_0$ + $HCls_1$\\
    &M3 & $CG_0$ + $CLS_{ft.}$ & &M10 & cLPs with $v_{1}$ + T5-base (full ft.) for $DGen$  + $HCls_0$ + $HCls_1$ \\
    &M4 & $DGen$ + $CG_1$ + $CLS_{ft.}$ & & M11 & cLPs with $v_{1}$ + T5-base (pre-trained) for $DGen$ + $HCls_0$ + $HCls_1$\\
    &M5 & $CG_0$ + $QAG_1$ + $CLS_{regex}$   & & M12 & cLPs with $v_{2}$ + T5-base (full ft.) for $DGen$  + $HCls_0$ + $HCls_1$\\
    &M6 & $DGen$ + $CG_1$ + $QAG_1$ + $CLS_{regex}$ & & M13 & cLPs with $v_{2}$ + T5-base (pre-trained) for $DGen$ + $HCls_0$ + $HCls_1$\\
    &M7 & $QG_0$ + $RG_0$ + $CLS_{regex}$ &&&\\

    \bottomrule
  \end{tabular}%
  }
\end{table}

\paragraph{\textbf{Unit Modules of $\model$}.} The variants of $\model$ consist of \textit{seven} distinct modules, each with a specific objective (cf. Table \ref{unitMudulesOf_SM}). The first module, denoted by $\textbf{DGen}$, generates a general visual description for the Q\&A setup. The second module, referred to as the \textbf{C}ontext \textbf{G}enerator ($\textbf{CG}$), is responsible for creating the meme context. The third module, designated as $\textbf{QAG}$, i.e., \textbf{Q}uery-\textbf{A}nswer \textbf{G}enerator, is tasked with generating all question-answer pairs based on a given input meme. The fourth module, known as $\textbf{QG}$, i.e., \textbf{Q}ueries \textbf{G}enerator, generates all important queries required to understand a meme . Finally, the fifth module, referred to as $\textbf{RG}$, i.e., \textbf{R}esponse \textbf{G}enerator, focuses on generating question-specific answers for each query from \textbf{QG}. Except for $\textbf{DGen}$, each of the modules mentioned above has \textit{two} alternative forms: the first one requires the overlaid meme text $\textbf{T}$ and the visual information $\textbf{I}$ of the meme $\textbf{M}$ as input (denoted with subscript `0', e.g., $HCls_0$). In contrast, the second form additionally requires a general description of visual information in \textbf{I} from $\textbf{DGen}$ or a contextual description from  $\textbf{CG}$ (denoted with the subscript `1', e.g., $HCls_1$). At the end of each pipeline, classification is carried out by matching a pattern using a regular expression, denoted by \textbf{CLS}$_{regex}$, or by using a fine-tuned model, denoted by \textbf{CLS}$_{ft}$. 

\paragraph{\textbf{Variants of $\fvariant$}.} There are \textit{eight} variants of $\fvariant$, constructed using the basic unit modules  (cf. Table \ref{variantsOf_MANTIS}). 

\begin{enumerate}[label=(\alph*)]
    \item \revised{\textbf{M0}:
    First, a visual description of $M$ is generated using $DGen$ and used in $QAG_{1}$ for relevant QA pairs, followed by a regex-based classification.}

    \item \revised{\textbf{M1}: First, a visual description of $M$ is generated using $DGen$ and used in $QG_{1}$ to obtain key queries, $Y_{query}$. Each query in $Y_{query}$ is then processed by $RG_{1}$, followed by regex-based classification.}
 
    \item \textbf{M2}: M2 is similar to M1, except that it utilizes a  \textit{fully fine-tuned} based classification module.

    \item \revised{\textbf{M3}: First, the meme context, $CG_{0}$ is retrieved, followed by a fine-tuned classification.}
    
    \item \revised{\textbf{M4}: First, a visual description is obtained from $DGen$ and utilized in $CG_{1}$ to obtain the meme context, followed by a fully fine-tuned classification.}  

    \item \revised{\textbf{M5}: First, a meme context is obtained using $CG_{0}$ and used in $QAG_1$ to generate QA pairs, followed by regex-based classification.}

    \item \revised{\textbf{M6}: First, a meme context, $CG_{1}$ is retrieved and utilized by $QAG_1$ for a series of QA pairs, followed by a regex-based classification.} 

    \item \revised{\textbf{M7}: First, a series of relevant questions is generated using $QG_{0}$, followed by a single response for each of the queries using $RG_{0}$. A regex-based approach is utilized for the classification.} 

\end{enumerate}

\paragraph{\textbf{Variants of $\svariant$}.} The variants of $\svariant$ consist of \textit{two} stages: (a) generation of a general contextual description, $y_{gdesc}$,  by $DGen$, and (b) hierarchical classification, referred to as \textbf{HCls}. Note that both modules, $DGen$ and $HCls$ utilize target-group and task-specific linear projectors. We will briefly discuss the configurations of the linear projectors before delving into the details of $DGen$ and $HCls$.  

\begin{figure}[!t]
\centering
\includegraphics[clip, trim={0cm 0cm 0cm 0cm}, width=0.8\textwidth]{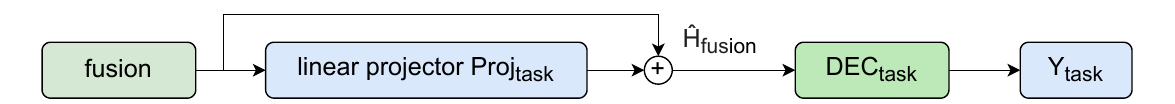} 
\caption{An illustration of the configuration of linear projectors. The fused-multimodal representation, $H_{fusion}$, is subjected to a linear projection. The variation in linear projectors is determined by the selection of tasks and the underlying data. The training process exclusively utilizes instances corresponding to the specific category of interest for target-category-based linear projectors. 
Note that only the linear projector undergoes training, whereas all other pre-trained parameters of the T5-base remain unaltered.  } 
\label{fig:linearProjectorTraining}
\end{figure}
\paragraph{\textbf{Linear Projector}.} Figure \ref{fig:linearProjectorTraining} represents the pipeline of a linear projector.  The multimodal fusion representation, $H_{fusion}$, undergoes a linear transformation before merging with itself via a residual connection. The latent information of linear projectors varies according to the task and underlying data, particularly when the task is specific to a target group. If a target group is specified, then  training is conducted exclusively using the training instances associated with that specific target group. This ensures the category-specific projection alignment for the given task $Y_{task}$. 
\begin{figure}[!t]
\centering
\includegraphics[width=1\textwidth]{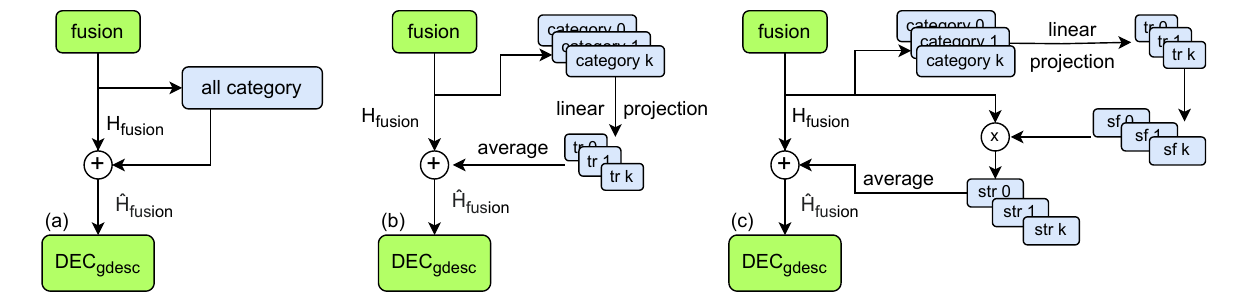} 
\caption{As illustration of the architectural difference in $\svariant$ variants. The key distinction lies in availing and integrating contextual information using linear projectors. The general multimodal representation, denoted as $H_{fusion}$, is subjected to (a) a linear projector that is trained with all instances irrespective of target categories (denoted by $v_{0}$), (b) a linear projection through each target category-specific linear projector (trained with category-specific instances), followed by averaging the results (denoted by $v_{1}$), and (c) a linear projection through each target category-specific linear projector (trained with category-specific instances) with a mechanism that attempts to learn category-specific scaling of $H_{fusion}$ (c.f. Eq. \ref{eq:fusionCategoryCard}) before averaging the results (denoted by $v_{2}$). The representation for $H_{fusion}$ is derived using Eqs. 
\ref{eq_visionEncAndProjection} and \ref{eq_gatedFusion}.} 
\label{fig:architecture_v0Andv1}
\end{figure}

\paragraph{\textbf{Hierarchical Classifiers}.} The classification module $HCls$ can be further divided into two variations, both of which utilize a fine-tuned task-specific linear projection layer: (a) the first one is fine-tuned to distinguish between a hateful and benign context, denoted by $\text{HCls}_0$; (b) the second variation is able to to differentiate between explicit and implicit hate contexts, denoted by $\text{HCls}_1$. Note that the linear projectors employed for classification utilize all training instances.

The pipeline of $DGen$ in $\svariant$ has \textit{three} architectural variations, namely $v_0$, $v_1$ and $v_2$ (cf. Figure \ref{fig:architecture_v0Andv1}). The distinction lies in how contextual information is integrated using linear projection: (a) the first variation, v$_0$ uses a generalized linear projector (gLP) that utilizes all training instances and integrates generalized latent information into $H_{fusion}$, resulting in $\hat{H}_{fusion}$; (b) the variation v$_1$ uses category-specific linear projectors (cLP) to obtain category specific transformed representations (tr$_{i}$). The rationale for using category-specific $\textbf{cLP}_i$, rather than generalized $\textbf{gLP}$, is to better capture the subtle and distinct characteristics associated with each category $i$. Consequently, each $\textbf{cLP}_i$ is trained exclusively with the training instances associated with the target group $i$. It  computes the transformed representation tr$_{i}$ for each category, adds their average to $H_{\text{fusion}}$, and produces $\hat{H}_{\text{fusion}}$; (c) in contrast to $v_1$, the third variation, $v_2$, introduces an additional linear projector that maps each category-specific representation tr$_{i}$ to its corresponding scaling factor sf$_{i}$ that scales $H_{fusion}$ accordingly. It produces another set of intermediate scaled transformed representation (str$_{i}$). Subsequently, the average of all str$_{i}$ is incorporated into $H_{fusion}$, yielding $\hat{H}_{fusion}$. The basic pipeline of $\svariant$ remains the same across all variants. However, the ablation is in the types of underlying modules (cf. Table \ref{variantsOf_MANTIS}). We outline the variations below:
\begin{enumerate}[label=(\alph*)]
    \item \revised{$\textbf{M8}$: Full fine-tune of T5-base.}
    \item \revised{$\textbf{M9}$: A single gLP with $v_0$ along with a fully fine-tuned T5-base model.}
    \item \revised{$\textbf{M10}$: cLPs with $v_1$ along with a fully fine-tuned T5-base.}
    \item \revised{$\textbf{M11}$: cLPs with $v_1$, along with a pre-trained T5-base.}
    \item \revised{$\textbf{M12}$: cLPs with $v_2$ along with a fully fine-tuned T5-base.}
    \item \revised{$\textbf{M13}$: cLPs with $v_2$ along with a pre-trained T5-base.}
\end{enumerate}

\begin{table*}[t!]
\centering
\caption{Benchmarking $\dataone$ using unimodal text (UM-T) baselines, multimodal (MM) baselines, task-specific baselines (TSB) and proposed models, \fvariant\ and \svariant. Note -- Early Fusion: Rob (L) + ResNet-152, I: image, T: text, ZS: zero shot, FT: full fine-tuned, P: precision, R: recall, F: F1 score, $\star$: best performing open-source baseline, $\maltese$:  best performing closed-source baseline. }
\label{tab:benchmarkComparison_MHSDBandBaseline}
\resizebox{\textwidth}{!}
{
\begin{tabular}{@{}clccc|ccc|ccc|ccc@{}}
\toprule
\multirow{2}{*}{\textbf{Mod.}} &
  \multirow{2}{*}{\textbf{Models}} &
  \multicolumn{3}{c}{\textbf{Explicit Hate}} &
  \multicolumn{3}{c}{\textbf{Implicit-Hate}} &
  \multicolumn{3}{c}{\textbf{Benign}} &
  \multicolumn{3}{c}{\textbf{Macro}} \\ \cmidrule(l){3-14} 
 &
   &
  \textbf{P} &
  \textbf{R} &
  \textbf{F} &
  \textbf{P} &
  \textbf{R} &
  \textbf{F} &
  \textbf{P} &
  \textbf{R} &
  \textbf{F} &
  \textbf{P} &
  \textbf{R} &
  \textbf{F} \\ \midrule
\multirow{3}{*}{UM-T} & T5-large (FT) & 0.55 & 0.54 & 0.54 & 0.5  & 0.86 & 0.63 & 0.81                     & 0.1  & 0.18 & 0.62 & 0.5  & 0.45 \\
& ALBERT   & 0.36 & 0.43 & 0.39 & 0.67 & 0.02 & 0.04 & 0.27                     & 0.62 & 0.38 & 0.43 & 0.36 & 0.27 \\

& BiLSTM  & 0.52 & 0.38 & 0.43 & 0.60 & 0.52 & 0.56 & \multicolumn{1}{r}{0.39} & 0.54 & 0.45 & 0.50 & 0.48 & 0.48 \\

                    \midrule

\multirow{16}{*}{MM} & Early Fusion     & 0.21 & 0.01 & 0.02 & 0.42 & 0.97 & 0.58 & 0.00                     & 0.00 & 0.00 & 0.21 & 0.33 & 0.20 \\
& BLIP2 (ZS) & 0.25 & 0.38 & 0.3  & 0.43 & 0.64 & 0.52 & 0.91                     & 0.03 & 0.06 & 0.53 & 0.35 & 0.29 \\
& BLIP2 (FT)   & 0.34 & 0.70 & 0.46        & 0.60  & 0.49 & 0.54       & 0.62 & 0.46 & 0.53        & 0.52 & 0.55 & 0.51 \\
& CLIP (ZS)  & 0.41 & 0.08 & 0.14 & 0.4  & 0.49 & 0.44 & 0.34                     & 0.44 & 0.38 & 0.38 & 0.33 & 0.32 \\
&  CLIP (FT) & 0.47 & 0.58 & 0.52 & 0.52 & 0.73 & 0.61 & 0.62                     & 0.22 & 0.33 & 0.54 & 0.51 & 0.49 \\
& CLIP-XLM-R (FT) & \revised{0.39} & \revised{0.7} & \revised{0.5} & \revised{0.55} & \revised{0.67} & \revised{0.6} & \revised{0.7} & \revised{0.3} & \revised{0.42} & \revised{0.55} & \revised{0.56} & \revised{0.50} \\

& LLaVA-NeXT (ZS)               & 0.49 & 0.23 & 0.32 & 0.45 & 0.82 & 0.58 & 0.60                     & 0.24 & 0.34 & 0.52 & 0.43 & 0.42 \\

& MiniGPT-v2 (ZS) & 0.26 & 0.33 & 0.29 & 0.44 & 0.43 & 0.43 & 0.34                     & 0.29 & 0.32 & 0.35 & 0.35 & 0.35 \\
& LlaVA (ZS) & 0.25 &  0.95 & 0.39 & 0.29 & 0.01 & 0.01 & 0.60 & 0.15 & 0.25 & 0.38 & 0.37 & 0.22 \\

& MiniGPT-4 (ZS)                 & 0.31 & 0.89 & 0.46 & 0.49 & 0.27 & 0.35 & 0.47                     & 0.12 & 0.19 & 0.42 & 0.43 & 0.33 \\ 
& \revised{QwenVL (ZS)} & \revised{0.26} & \revised{0.57} & \revised{0.36} & \revised{0.7} & \revised{0.11} & \revised{0.2} & \revised{0.59} & \revised{0.79} & \revised{0.68} & \revised{0.52} & \revised{0.49} & \revised{0.41}\\

& \revised{QwenVL (FT)} & \revised{0.32} & \revised{0.59} & \revised{0.42} & \revised{0.67} & \revised{0.26} & \revised{0.38} & \revised{0.61} & \revised{0.76} & \revised{0.68} & \revised{0.53} & \revised{0.54} & \revised{0.49} \\

\rowcolor{yellow!25}
& \revised{$\text{GPT-4o}^{\maltese}$} & \revised{0.69} & \revised{0.58} & \revised{0.63} & \revised{0.29} & \revised{0.5} & \revised{0.37} & \revised{0.71} & \revised{0.67} & \revised{0.69} & \revised{0.54} & \revised{0.54} & \revised{0.56} \\

\rowcolor{yellow!25}
& \revised{Gemini 2.5} & \revised{0.62} & \revised{0.44} & \revised{0.51} & \revised{0.27} & \revised{0.68} & \revised{0.39} & \revised{0.74} & \revised{0.71} & \revised{0.72} & \revised{0.52} & \revised{0.59} & \revised{0.54} \\

\multirow{10}{*}{}     & MM-CoT (T5-base)    & 0.42 & 0.04 & 0.06 & 0.42 & 0.65 & 0.51 & 0.49 & 0.43 & 0.46 & 0.45 & 0.38 & 0.35 \\
\multirow{10}{*}{}     & MM-CoT (T5-large)    & 0.31 & 0.74 & 0.43 & 0.57 & 0.13 & 0.21 & 0.51 & 0.58 & 0.54 & 0.46 & 0.48 & 0.39 \\
\cmidrule{2-14} 

\multirow{8}{*}{TSB}  & \revised{HClipper(cross)} & \revised{0.35} & \revised{0.73} & \revised{0.47} & \revised{0.5} & \revised{0.45} & \revised{0.47} & \revised{0.66} & \revised{0.46} & \revised{0.54} & \revised{0.5} & \revised{0.55} & \revised{0.50} \\

\rowcolor{yellow!25} 
& \revised{$\text{HClipper(concat)}^{\star}$ } & \revised{0.45} & \revised{0.49} & \revised{0.47} & \revised{0.57} & \revised{0.41} & \revised{0.47} & \revised{0.57} & \revised{0.7} & \revised{0.63} & \revised{0.53} & \revised{0.53} & \revised{0.52} \\

& \revised{HClipper(align)} & \revised{0.46} & \revised{0.51} & \revised{0.48} & \revised{0.51} & \revised{0.5} & \revised{0.51} & \revised{0.59} & \revised{0.51} & \revised{0.54} & \revised{0.52} & \revised{0.51} & \revised{0.51} \\

& \revised{HClipper(align-concat)} & \revised{0.41} & \revised{0.63} & \revised{0.5} & \revised{0.51} & \revised{0.59} & \revised{0.55} & \revised{0.65} & \revised{0.38} & \revised{0.48} & \revised{0.52} & \revised{0.53} & \revised{0.51} \\
& Pro-Cap (SimpleCls)$^{\dagger}$ & 0.43 & 0.54 & 0.48 & 0.57 & 0.49 & 0.53 & 0.47 & 0.60 & 0.53 & 0.49 & 0.54 & 0.51 \\
& Pro-Cap (SingleCls) & 0.41 & 0.53 & 0.42 & 0.55 & 0.48 & 0.51 & 0.50 & 0.43 & 0.46 & 0.52 & 0.48 & 0.49 \\
& CapAlign & 0.21 & 0.37 & 0.27 & 0.41 & 0.036 & 0.066 & 0.46 & 0.73 & 0.56 & 0.36 & 0.37 & 0.29 \\
& PromptHate & 0.37 & 0.53 & 0.43 & 0.62 & 0.50 & 0.55 & 0.53 & 0.53 & 0.53 & 0.50 & 0.52 & 0.50 \\
\midrule
\multicolumn{2}{c}{\fvariant\ (M7)}    & 0.39 & 0.67 & 0.49 & 0.58 & 0.55 & 0.56 & 0.65 & 0.48 & 0.55 & 0.54 & 0.57 & 0.54 \\ 
\rowcolor{LightCyan} \multicolumn{2}{c}{\svariant\ (M11)$^{\ddagger}$}     & 0.59 & 0.59 & 0.59 & 0.78 & 0.54 & 0.64 & 0.59 & 0.32 & 0.41 & 0.65 & 0.48 & 0.55 \\ \midrule

\multicolumn{2}{c}{$\Delta_{\ddagger -- \star}$}             & {\textcolor{blue}{$\uparrow14\%$}} & {\textcolor{blue}{$\uparrow10\%$}} & {\textcolor{blue}{$\uparrow12\%$}}  
& {\textcolor{blue}{$\uparrow21\%$}}  & {\textcolor{blue}{$\uparrow 13\%$}} & {\textcolor{blue}{$\uparrow17\%$}}
& {\textcolor{blue}{$\uparrow2\%$}}& {\textcolor{red}{$\downarrow38\%$}}& {\textcolor{red}{$\downarrow22\%$}}
& {\textcolor{blue}{$\uparrow12\%$}}& {\textcolor{red}{$\downarrow7\%$}} & {\textcolor{blue}{$\uparrow3\%$}}\\

\multicolumn{2}{c}{$\Delta_{\ddagger - \maltese}$}             & {\textcolor{red}{$\downarrow10\%$}} & {\textcolor{blue}{$\uparrow1\%$}} & {\textcolor{red}{$\downarrow4\%$}}  
& {\textcolor{blue}{$\uparrow49\%$}}  & {\textcolor{blue}{$\uparrow 4\%$}} & {\textcolor{blue}{$\uparrow27\%$}}
& {\textcolor{red}{$\downarrow12\%$}}& {\textcolor{red}{$\downarrow28\%$}}& {\textcolor{red}{$\downarrow35\%$}}
& {\textcolor{blue}{$\uparrow11\%$}}& {\textcolor{red}{$\downarrow6\%$}} & {\textcolor{red}{$\downarrow1\%$}}\\

  \bottomrule
\end{tabular}
}
\end{table*}

\subsection{Evaluation Metrics}

We evaluate model performance using category-wise precision (P), recall (R), and F1 score (F) computed separately for each class, i.e., explicit hate, implicit hate, and benign. We additionally report macro-averaged precision, recall, and F1 score, which provide a more reliable estimate of overall performance. Among these, the macro-F1 score serves as our primary metric for comparing models, as it balances precision and recall across all classes.


\section{Benchmarking and Empirical Analysis}
This section presents the benchmarking results of \dataone\ (c.f. Table \ref{tab:benchmarkComparison_MHSDBandBaseline}) and \datatwo\ (c.f. Table \ref{tab:benchmarkComparison_MHSDBConandBaseline}) using baselines and the proposed models, \fvariant\ and \svariant. 
In addition, it discusses the ablation study results (c.f. Table \ref{tab:benchmarkComparison_FVandSVonMHS_DB}), comparing different variations and components of \fvariant\ and \svariant.

\begin{figure*}[t]
\captionsetup[subfigure]{labelformat=empty, labelsep=none}
    \centering
    \pgfplotsset{
        every axis/.append style={
            legend style={
                at={(0.5,-0.2)},
                anchor=north,
                legend columns=6,
                font=\tiny,
                draw=none,
                fill=none,
                column sep=2pt,
                text=black
            },
            legend image code/.code={\draw[#1] (0cm,-0.05cm) rectangle (0.15cm,0.15cm);},
            width=0.33\linewidth,
            height=3.5cm,
            ybar=0.5pt,
            yticklabel style={font=\tiny},
            bar width=11pt,
            ymin=0.8, ymax=1.1,
            ytick={ 0.85, 0.95, 1.05},
            grid=major,
            xtick=data,
            enlarge x limits=0.5,
            clip=false
        },
HClipper (align-concat)/.style={fill=blue!30, draw=blue!90, pattern=north east lines, pattern color=blue!80},
CLIP-XLM-R (FT)/.style={fill=red!30, draw=red!90, pattern=dots, pattern color=red!80},
CLIP (FT)/.style={fill=green!30, draw=green!90, pattern=crosshatch, pattern color=green!80},
Gemini 2.5/.style={ fill=yellow!40, draw=yellow!80!black, pattern=checkerboard, pattern color=yellow!90!black},
GPT-4o/.style={fill=orange!30, draw=orange!90, pattern=horizontal lines, pattern color=orange!80},
T5-Large (FT)/.style={fill=violet!30, draw=violet!90, pattern=north west lines, pattern color=violet!80},
SAFE-MEME-H/.style={fill=cyan!30, draw=cyan!90, pattern=grid, pattern color=cyan!80},
Llava-Next/.style={fill=brown!30, draw=brown!90, pattern=vertical lines, pattern color=brown!80},
Early-Fusion/.style={fill=magenta!30, draw=magenta!90, pattern=bricks, pattern color=magenta!80},
HClipper (concat)/.style={fill=teal!30, draw=teal!90, pattern=crosshatch dots, pattern color=teal!80},
QwenVL (ZS)/.style={fill=gray!30, draw=gray!90, pattern=north east lines, pattern color=gray!80},
QwenVL (FT)/.style={fill=olive!30, draw=olive!90, pattern=north east lines, pattern color=olive!80},
CLIP-XLM-R (FT)/.style={fill=purple!30, draw=purple!90, pattern=dots, pattern color=purple!80},
ProCap(SingleCls)/.style={fill=red!30, draw=red!90, pattern=vertical lines, pattern color=red!80},
ProCap(SimpleCls)/.style={fill=darkgray!30, draw=darkgray!90, pattern=crosshatch, pattern color=darkgray!80},
PromptHate/.style={fill=lightgray!30, draw=lightgray!90, pattern=grid, pattern color=lightgray!80},
HClipper (cross)/.style={fill=black!15, draw=black!90, pattern=horizontal lines, pattern color=black!80},
HClipper (align)/.style={fill=blue!50!green!30, draw=blue!50!green!90, pattern=dots, pattern color=blue!50!green!80},
BLIP2 (FT)/.style={fill=red!50!orange!30, draw=red!50!orange!90, pattern=north west lines, pattern color=red!50!orange!80},
SAFE-MEME-QA/.style={fill=violet!50!blue!30, draw=violet!50!blue!90, pattern=bricks, pattern color=violet!50!blue!80},
CLIP-XLM-R (FT)0/.style={fill=red!50!violet!30, draw=red!50!violet!90, pattern=crosshatch dots, pattern color=red!50!violet!80},
}
    \subcaptionbox{\label{fig:visOnMHS_exp}}{
    \begin{tikzpicture}
        \begin{axis}[
            symbolic x coords={(a) Explicit},
            ylabel={$\hat{F}_{i } = F_{i } / F_{M11}$},
            ylabel style={font=\tiny},
            legend to name=SharedLegend]
            \addplot[HClipper (align-concat)] coordinates {((a) Explicit,0.84)};  \addlegendentry{HClipper (align-concat)} 
            \addplot[CLIP-XLM-R (FT)] coordinates {((a) Explicit,0.84)}; \addlegendentry{CLIP-XLM-R (FT)} 
            \addlegendimage{Gemini 2.5} coordinates {(Explicit,0.86)}; \addlegendentry{Gemini 2.5} 
            \addplot[CLIP (FT)] coordinates {((a) Explicit,0.88)}; \addlegendentry{CLIP (FT)}
            \addplot[T5-Large (FT)] coordinates {((a) Explicit,0.91)}; \addlegendentry{T5-Large (FT)}
            \addplot[SAFE-MEME-H] coordinates {((a) Explicit,1)}; \addlegendentry{SAFE-MEME-H}
            \addplot[GPT-4o] coordinates {((a) Explicit,1.07)}; \addlegendentry{GPT-4o}

            \addlegendimage{Llava-Next} \addlegendentry{Llava-Next}
            \addlegendimage{Early-Fusion} \addlegendentry{Early-Fusion}
            \addlegendimage{HClipper (concat)} \addlegendentry{HClipper (concat)}
            \addlegendimage{QwenVL (ZS)} \addlegendentry{QwenVL}
            \addlegendimage{QwenVL (FT)} \addlegendentry{QwenVL (FT)}
            \addlegendimage{ProCap(SingleCls)} \addlegendentry{ProCap(SingleCls)}
            \addlegendimage{ProCap(SimpleCls)} \addlegendentry{ProCap(SimpleCls)}
            \addlegendimage{PromptHate} \addlegendentry{PromptHate}
            \addlegendimage{HClipper (cross)} \addlegendentry{HClipper (cross)}
            \addlegendimage{HClipper (align)} \addlegendentry{HClipper (align)}
            \addlegendimage{BLIP2 (FT)} \addlegendentry{BLIP2 (FT)}
            \addlegendimage{SAFE-MEME-QA} \addlegendentry{SAFE-MEME-QA}

        \end{axis}
    \end{tikzpicture}
    }\hfill
    \subcaptionbox{\label{fig:visOnMHS_imp}}{
    \begin{tikzpicture}
        \begin{axis}[symbolic x coords={(b) Implicit}, 
            ymin=0.85, ymax=1.02,
            yticklabel style={font=\tiny},
            ytick={0.85, 0.9, 0.95, 1}]
            \addplot[Llava-Next] coordinates {((b) Implicit,0.9)};  
            \addplot[Early-Fusion] coordinates {((b) Implicit,0.9)}; 
            \addplot[CLIP-XLM-R (FT)] coordinates {((b) Implicit,0.94)}; 
            \addplot[CLIP (FT)] coordinates {((b) Implicit,0.95)}; 
            \addplot[T5-Large (FT)] coordinates {((b) Implicit,0.98)}; 
            \addplot[SAFE-MEME-H] coordinates {((b) Implicit, 1)}; 
        \end{axis}
    \end{tikzpicture}
    }\hfill
    \subcaptionbox{\label{fig:visOnMHS_ben}}{
    \begin{tikzpicture}
        \begin{axis}[symbolic x coords={(c) Benign}, 
            ymin=0.95, ymax=1.8,
            yticklabel style={font=\tiny},
            ytick={0.95, 1.35 , 1.8}]
            \addplot[SAFE-MEME-H] coordinates {((c) Benign, 1)};
            \addplot[HClipper (concat)] coordinates {((c) Benign,1.53)}; 
            \addplot[QwenVL (ZS)] coordinates {((c) Benign,1.65)};
            \addplot[QwenVL (FT)] coordinates {((c) Benign,1.65)};
            \addplot[GPT-4o] coordinates {((c) Benign,1.68)};
            \addplot[Gemini 2.5] coordinates {((c) Benign, 1.75)};
        \end{axis}
    \end{tikzpicture}
    }
    \hfill

\subcaptionbox{\label{fig:visOnMHS_expSkew}}{
\begin{tikzpicture}
    \begin{axis}[
        symbolic x coords={(d) Explicit},
        xtick=data,
        ymin=-0.35, ymax=0.25,
        ytick={-0.3, -0.2, -0.1, 0, 0.1, 0.2},
        yticklabel style={font=\tiny},
        ylabel={$\delta = P - R$},
        ylabel style={font=\tiny},
        width=0.33\linewidth,
        height=3.5cm,
        ybar=2pt,
        bar width=11pt,
        grid=major,
        enlarge x limits=0.4,
        clip=false,
        extra y ticks={0},
        extra y tick style={
            grid=major,
            grid style={black, thick, dashed}
        }
    ]
    \addplot[HClipper (align-concat)]     coordinates {((d) Explicit, -0.1)}; 
    \addplot[CLIP-XLM-R (FT)]     coordinates {((d) Explicit, -0.31)};  
    \addplot[CLIP (FT)]     coordinates {((d) Explicit, -0.11)};  
    \addplot[T5-Large (FT)]     coordinates {((d) Explicit, 0.1)}; 
    \addplot[SAFE-MEME-H]     coordinates {((d) Explicit, 0.0)}; 
    \addplot[GPT-4o]     coordinates {((d) Explicit, 0.21)}; 
    \end{axis}
\end{tikzpicture}
}
    \hfill
\subcaptionbox{\label{fig:visOnMHS_impSkew}}{
\begin{tikzpicture}
    \begin{axis}[
        symbolic x coords={(e) Implicit},
        xtick=data,
        ymin=-0.6, ymax=0.3,
        ytick={-0.6, -0.35, -0.15, 0, 0.3},
        yticklabel style={font=\tiny},
        width=0.37\linewidth,
        height=3.5cm,
        ybar=2pt,
        bar width=11pt,
        grid=major,
        enlarge x limits=0.4,
        clip=false,
        extra y ticks={0},
        extra y tick style={
            grid=major,
            grid style={black, thick, dashed}
        }
    ]
    \addplot[Llava-Next]     coordinates {((e) Implicit, -0.37)}; 
    \addplot[Early-Fusion]     coordinates {((e) Implicit, -0.55)};  
    \addplot[CLIP-XLM-R (FT)]     coordinates {((e) Implicit, -0.12)};  
    \addplot[CLIP (FT)]     coordinates {((e) Implicit, -0.21)}; 
    \addplot[T5-Large (FT)]     coordinates {((e) Implicit, -0.36)}; 
    \addplot[SAFE-MEME-QA]     coordinates {((e) Implicit, 0.03)}; 
    \addplot[SAFE-MEME-H]     coordinates {((e) Implicit, 0.24)}; 
    \end{axis}
\end{tikzpicture}
}
    \hfill
\subcaptionbox{\label{fig:visOnMHS_benSkew}}{
\begin{tikzpicture}
    \begin{axis}[
        symbolic x coords={(f) Benign},
        xtick=data,
        ymin=-0.2, ymax=0.3,
        ytick={-0.2, -0.1, 0, 0.3},
        yticklabel style={font=\tiny},
        width=0.33\linewidth,
        height=3.5cm,
        ybar=2pt,
        bar width=11pt,
        grid=major,
        enlarge x limits=0.4,
        clip=false,
        extra y ticks={0},
        extra y tick style={
            grid=major,
            grid style={black, thick, dashed}
        }
    ]

    \addplot[SAFE-MEME-H] coordinates {((f) Benign, 0.27)};
    \addplot[HClipper (concat)] coordinates {((f) Benign, -0.13)}; 
    \addplot[QwenVL (ZS)] coordinates {((f) Benign, -0.2)};
    \addplot[QwenVL (FT)] coordinates {((f) Benign, -0.15)};
    \addplot[GPT-4o] coordinates {((f) Benign, 0.04)};
    \addplot[Gemini 2.5] coordinates {((f) Benign, 0.03)};

    \end{axis}
\end{tikzpicture}
}

            
            

    \ref{SharedLegend}
    \caption{Performance comparison of baseline models and $\model$ variants on $\dataone$ across categories i.e., explicit hate, implicit hate, and benign. Subplots (a)–(c) show normalized F1 scores relative to $\svariant$ (computed as baseline / $\svariant$), while subplots (d)–(f) illustrate precision–recall (P–R) skew, i.e., deviation from the $P=R$ line (c.f. Table \ref{tab:benchmarkComparison_MHSDBandBaseline}).  We report the top 20 percentile of baseline models, highlighting their comparative performance across scenarios.}
    \label{fig:visOnMHS}
\end{figure*}

\subsection{Comparative Analysis on \dataone} For \textit{explicit hate},  T5-large (FT), \revised{CLIP (FT), CLIP-XLM-R (FT), HClipper (align-concat), GPT-4o exhibit superior performance among baselines (cf. Figure \ref{fig:visOnMHS}(a)) with F1 scores of $0.54$, $0.52$, $0.50$, $0.50$, and 0.63 respectively (cf. Table \ref{tab:benchmarkComparison_MHSDBandBaseline})}. Variant M7 i.e., \fvariant\ (which operates on Q\&A reasoning) and M11 i.e., \svariant\ (which uses a category-specific linear projector with v$_1$) obtain F1 scores of $0.49$ and $0.59$. Both the closed-source models, GPT-4o and Gemini 2.5, score $0.63$ and $0.51$, respectively.  A close analysis of the Q\&A pairs generated in first stage of M7 reveals that occasionally longer responses introduce noise, especially when compared compared to the concise, focused context that unimodal baselines integrate in a single stage. In contrast, M11 gains by $5\%$, $7\%$, $9\%$, $9\%$, and $8\%$ against T5-large (FT), CLIP (FT), CLIP-XLM-R , HClipper (align-concat), and Gemini 2.5, respectively, but falls short by $4\%$ against GPT-4o. In addition, $\svariant$ acts a balanced classifier for this category (cf. Figure \ref{fig:visOnMHS}(a)). Both GPT-4o (P: 0.69, R: 0.58) and Gemini 2.5 (Precision: 0.62, Recall: 0.44) demonstrate a high-precision but low-recall. This indicates a conservative approach in the identification of explicit hate content. If classified as explicit hate, the predictions are generally accurate; however, they fail to detect a substantial number of true positive cases. It might be attributed to alignment training strategies aimed at reducing over-refusal. The prominent verbal cues in the input likely contribute to T5-large (FT)'s superior performance in the explicit hate category. Within other baselines, BiLSTM, BLIP2 (FT), MiniGPT-4 (ZS), QwenVL (FT), MM-CoT (T5-large), HClipper (cross/concat/align), Pro-Cap (SimpleCls/SingleCls), and PromptHate show moderate performance with F1 scores ranging from 0.42 to 0.48.

For the \textit{implicit hate} category, T5-large (FT), CLIP (FT) and PromptHate have superior performance among text-based unimodal, multimodal and task-specific baselines, respectively, with F1 scores of $0.63$, $0.61$, and $0.55$ (cf. Table \ref{tab:benchmarkComparison_MHSDBandBaseline}). \revised{Note that HClipper (align-concat) has similar performance for implicit hate instances (cf. Figure \ref{fig:visOnMHS}(b)). Surprisingly, both closed-source baseline models, GPT-4o and Gemini, exhibit low precision scores. However, their corresponding recall is higher, which indicates a high false positive rate, suggesting fundamental confusion about what constitutes implicit hate.} Variant M7 achieves an F1 score of $0.56$, trailing behind T5-large (FT) and CLIP (FT) by $7\%$ and $5\%$, respectively. Notably, M7's recall for the implicit hate category is $0.55$, significantly lower than the recall scores of $0.86$ for T5-large (FT) and $0.73$ for CLIP (FT). While T5-large (FT) demonstrates a high recall, its precision is close to random, suggesting considerable prediction ambiguity in identifying implicit hate. This imbalance indicates potential overfitting in that category, resulting in a moderate F1 score of $0.63$. \revised{In contrast, $M11$ achieves a score of $0.64$ with a precision and recall of 0.78 and 0.54. It achieves a gain of $1\%$, $3\%$ and $9\%$, respectively, over T5-large (FT), CLIP (FT) and Pro-Cap (SimpleCls) or HClipper (align-concat). Although the recall score of M7 is lower than that of other top-performing models -- yet it shows a balanced score in terms of P-R skewness (cf. Figure \ref{fig:visOnMHS}(e))}.  


\begin{table}[t!]
\centering
\caption{Benchmarking $\datatwo$ using selected baselines and proposed models, $\fvariant$ and $\svariant$. Note -- I: image, T: text, ZS: zero-shot, FT: full fine-tuned, P: precision, R: recall, F: F1 score, $\star$: best performing open-source baseline, $\maltese$: best performing closed-source baseline}
\label{tab:benchmarkComparison_MHSDBConandBaseline}
\resizebox{\textwidth}{!}{
\begin{tabular}{@{}clccc|ccc|ccc|ccc@{}}
\toprule
&  \multirow{2}{*}{\textbf{Models}} &
  \multicolumn{3}{c}{\textbf{Explicit Hate}} &
  \multicolumn{3}{c}{\textbf{Implicit-Hate}} &
  \multicolumn{3}{c}{\textbf{Benign}} &
  \multicolumn{3}{c}{\textbf{Macro}} \\ \cmidrule(l){3-14} 
  && \textbf{P} &
  \textbf{R} &
  \textbf{F} &
  \textbf{P} &
  \textbf{R} &
  \textbf{F} &
  \textbf{P} &
  \textbf{R} &
  \textbf{F} &
  \textbf{P} &
  \textbf{R} &
  \textbf{F} \\ 
\midrule

\multirow{1}{*}{\rotatebox{0}{UM-T}} & T5-large (FT)  & 0.625 & 0.672 & 0.648 & 0.441 & 0.602 & 0.509 & 0.824 & 0.447 & 0.580 & 0.63 & 0.574 & 0.579 \\
\midrule
\multirow{7}{*}{MM} & BLIP2 (FT)   & 0.431 & 0.433 & 0.431        & 0.394  & 0.478 & 0.431       & 0.592 & 0.474 & 0.526        & 0.472 & 0.461 & 0.462 \\

 &  CLIP (FT) & 0.468 & 0.469 & 0.469 & 0.299 & 0.105 & 0.155 & 0.350 & 0.316 & 0.332 & 0.326 & 0.333 & 0.302 \\

&\revised{CLIP-XLM-R (FT)} & \revised{0.381} & \revised{0.657} & \revised{0.482} & \revised{0.399} & \revised{0.726} & \revised{0.514} & \revised{0.741} & \revised{0.403} & \revised{0.534} & \revised{0.599} & \revised{0.503} & \revised{0.510} \\

& \revised{QwenVL (ZS)} & \revised{0.548} & \revised{0.872} & \revised{0.673} & \revised{0.523} & \revised{0.081} & \revised{0.140} & \revised{0.690} & \revised{0.882} & \revised{0.775} & \revised{0.587} & \revised{0.611} & \revised{0.529} \\

& \revised{QwenVL (FT)} & \revised{0.701} & \revised{0.542} & \revised{0.611} & \revised{0.491} & \revised{0.297} & \revised{0.370} & \revised{0.591} & \revised{0.964} & \revised{0.732} & \revised{0.594} & \revised{0.601} & \revised{0.571} \\

\rowcolor{yellow!25}
& \revised{GPT-4o} & \revised{0.652} & \revised{0.743} & \revised{0.694} & \revised{0.314} & \revised{0.555} & \revised{0.401} & \revised{0.871} & \revised{0.852} & \revised{0.861} & \revised{0.612} & \revised{0.717} & \revised{0.652} \\

\rowcolor{yellow!25}
& \revised{$\text{Gemini 2.5}^{\maltese}$} & \revised{0.612} & \revised{0.907} & \revised{0.730} & \revised{0.451} & \revised{0.432} & \revised{0.441} & \revised{0.935} & \revised{0.706} & \revised{0.804} & \revised{0.666} & \revised{0.682} & \revised{0.658} \\

\cmidrule{2-14}


\multirow{5}{*}{\rotatebox{0}{TSB}} & \revised{HClipper(concat)} & \revised{0.680} & \revised{0.447} & \revised{0.540} & \revised{0.444} & \revised{0.474} & \revised{0.459} & \revised{0.592} & \revised{0.755} & \revised{0.664} & \revised{0.572} & \revised{0.559} & \revised{0.554} \\


\rowcolor{yellow!25}
& \revised{$\text{HClipper(align-concat)}^{\star}$} & \revised{0.669} & \revised{0.518} & \revised{0.584} & \revised{0.436} & \revised{0.686} & \revised{0.533} & \revised{0.769} & \revised{0.490} & \revised{0.599} & \revised{0.625} & \revised{0.565} & \revised{0.572} \\

& Pro-Cap (SingleCls) & 0.412 & 0.531 & 0.425 & 0.553 & 0.485 & 0.515 & 0.503 & 0.431 & 0.462 & 0.529 & 0.487 & 0.493 \\

& Pro-Cap (SimpleCls) & 0.431 & 0.545 & 0.487 & 0.573 & 0.494 & 0.532 & 0.471 & 0.604 & 0.531 & 0.496 & 0.549 & 0.518 \\


& PromptHate & 0.372 & 0.535 & 0.431 & 0.629 & 0.503 & 0.558 & 0.532 & 0.537 & 0.539 & 0.502 & 0.521 & 0.505 \\
\midrule
\rowcolor{yellow!25}
&  $\fvariant$ (M7)$^{\dagger}$   & 0.689 & 0.535 & 0.602 & 0.526 & 0.474 & 0.499  & 0.664 & 0.879 & 0.757  & 0.626 & 0.63 & 0.619    \\
& $\svariant$ (M11)  & 0.543 & 0.398 & 0.459 & 0.518 & 0.461 & 0.488  & 0.578 & 0.522 & 0.548  & 0.546 & 0.460 & 0.498\\

\midrule
& \multicolumn{1}{l}{$\Delta_{\dagger -- \star}$} &{\textcolor{blue}{$\uparrow2.0\%$}} & {\textcolor{blue}{$\uparrow1.7\%$}} & {\textcolor{blue}{$\uparrow 1.8\%$}}  
& {\textcolor{blue}{$\uparrow9.0\%$}}  & {\textcolor{red}{$\downarrow 21.2\%$}} & {\textcolor{red}{$\downarrow3.4\%$}}
& {\textcolor{red}{$\downarrow10.5\%$}}& {\textcolor{blue}{$\uparrow38.9\%$}}& {\textcolor{blue}{$\uparrow15.8\%$}}
& {\textcolor{blue}{$\uparrow0.1\%$}} & {\textcolor{blue}{$\uparrow6.5\%$}} & {\textcolor{blue}{$\uparrow4.7\%$}}\\

& \multicolumn{1}{l}{$\Delta_{\maltese -- \dagger}$} &{\textcolor{blue}{$\uparrow7.7\%$}} & {\textcolor{red}{$\downarrow32.7\%$}} & {\textcolor{red}{$\downarrow 12.8\%$}}  
& {\textcolor{blue}{$\uparrow7.5\%$}}  & {\textcolor{blue}{$\uparrow 4.2\%$}} & {\textcolor{blue}{$\uparrow5.8\%$}}
& {\textcolor{red}{$\downarrow27.1\%$}}& {\textcolor{blue}{$\uparrow17.3\%$}}& {\textcolor{red}{$\downarrow4.7\%$}}
& {\textcolor{red}{$\downarrow4.0\%$}} & {\textcolor{red}{$\downarrow5.2\%$}} & {\textcolor{red}{$\downarrow3.9\%$}}\\

\hline
\end{tabular}%
\vspace{-5mm}
}
\end{table}

In \textit{benign} cases, M7 and M11 achieve moderate F1 scores of $0.55$, and $0.41$, respectively. Among all the baseline models, Qwen-VL (ZS/FT), HClipper (concat), GPT-4o, and Gemini 2.5 achieve strong performance (cf. Figure \ref{fig:visOnMHS}(c)), with scores of 0.68, 0.69, 0.72, and 0.63, respectively (cf. Table \ref{tab:benchmarkComparison_MHSDBandBaseline}). Among other baselines, BLIP2 (FT), MM-CoT (T5-large), HClipper(cross/align), Pro-Cap (SimpleCls), CapAlign, and PromptHate exhibit competitive performance, ranging from 0.53 to 0.56 in F1 score. In text-based unimodal baselines, while BiLSTM and ALBERT have F1 scores of $0.45$ and $0.34$, respectively, T5 large (FT) score drops to $0.18$. This result indicates that despite impressive recall rates for the implicit hate category, unimodal hatefulness prediction using T5-large might exhibit bias against valid, harmless samples. Among multimodal baselines, CLIP (ZS), LLaVA-Next (ZS), CLIP (FT), and miniGPT-v2 (ZS) have moderate F1 scores of $0.38$, $0.34$, $0.33$ and $0.32$, followed by LLava (ZS) and miniGPT-4 (ZS) with scores $0.25$, and $0.19$, respectively. 




 



\begin{figure*}[t]
\captionsetup[subfigure]{labelformat=empty}    
    \centering
    \pgfplotsset{
        every axis/.append style={
            legend style={
                at={(0.5,-0.2)},
                anchor=north,
                legend columns=6,
                font=\tiny,
                draw=none,
                fill=none,
                column sep=2pt,
                text=black
            },
            legend image code/.code={\draw[#1] (0cm,-0.05cm) rectangle (0.15cm,0.15cm);},            
            width=0.37\linewidth,
            height=3.5cm,
            ybar=0.5pt,
            bar width=9pt,
            ymin=0.850, ymax=1.215,
            ytick={ 0.95, 1.05, 1.15},
            grid=major,
            xtick=data,
            enlarge x limits=0.5,
            clip=false
        },
QwenVL (FT)/.style={fill=blue!30, draw=blue!90, pattern=north east lines, pattern color=blue!80},
T5-large (FT)/.style={fill=red!30, draw=red!90, pattern=dots, pattern color=red!80},
GPT-4o/.style={fill=green!30, draw=green!90, pattern=crosshatch, pattern color=green!80},
Gemini 2.5/.style={fill=orange!30, draw=orange!90, pattern=horizontal lines, pattern color=orange!80},
Pro-Cap (SingleCls)/.style={fill=darkgray!30, draw=darkgray!90, pattern=crosshatch, pattern color=darkgray!80},
Pro-Cap (SimpleCls)/.style={fill=cyan!30, draw=cyan!90, pattern=grid, pattern color=cyan!80},
HClipper(align-concat)/.style={fill=brown!30, draw=brown!90, pattern=vertical lines, pattern color=brown!80},
HClipper(align)/.style={fill=magenta!30, draw=magenta!90, pattern=bricks, pattern color=magenta!80},
HClipper(cross)/.style={fill=teal!30, draw=teal!90, pattern=crosshatch dots, pattern color=teal!80},
SAFE-MEME-QA/.style={fill=gray!30, draw=gray!90, pattern=north east lines, pattern color=gray!80},
PromptHate/.style={fill=olive!30, draw=olive!90, pattern=north east lines, pattern color=olive!80},
CLIP-XLM-R/.style={fill=purple!30, draw=purple!90, pattern=dots, pattern color=purple!80},
HClipper(concat)/.style={fill=black!30, draw=black!90, pattern=vertical lines, pattern color=black!80},
}
    \subcaptionbox{\label{fig:visOnMHSConf_exp}}{
    \begin{tikzpicture}
        \begin{axis}[symbolic x coords={(a) Explicit}, bar width=11pt,
        ylabel={$\hat{F}_{i } = F_{i } / F_{M7}$},         yticklabel style={font=\tiny},
            ylabel style={font=\tiny},
            legend to name=SharedLegend]

            \addplot[HClipper(concat)] coordinates {((a) Explicit, 0.897)};
            \addlegendentry{HClipper(align)} 
            \addplot[HClipper(align-concat)] coordinates {((a) Explicit, 0.97)};
            \addlegendentry{HClipper(align-concat)} 
            \addplot[SAFE-MEME-QA] coordinates {((a) Explicit, 1)}; \addlegendentry{SAFE-MEME-QA} 
            \addplot[QwenVL (FT)] coordinates {((a) Explicit,1.01)}; \addlegendentry{QwenVL (FT)} 
            \addplot[T5-large (FT)] coordinates {((a) Explicit,1.07)}; \addlegendentry{T5-large (FT)} 
            \addplot[GPT-4o] coordinates {((a) Explicit,1.152)}; \addlegendentry{GPT-4o} 
            \addplot[Gemini 2.5] coordinates {((a) Explicit,1.212)}; \addlegendentry{Gemini 2.5} 
            
            \addlegendimage{Pro-Cap (SingleCls)} \addlegendentry{Pro-Cap (SingleCls)}
            \addlegendimage{Pro-Cap (SimpleCls)} \addlegendentry{Pro-Cap (SimpleCls)}
            \addlegendimage{HClipper(cross)} \addlegendentry{HClipper(cross)}     
            \addlegendimage{PromptHate} \addlegendentry{PromptHate}
            \addlegendimage{CLIP-XLM-R} \addlegendentry{CLIP-XLM-R}
            \addlegendimage{HClipper(concat)} \addlegendentry{HClipper(concat)}
        \end{axis}
    \end{tikzpicture}
    }\hfill
    \subcaptionbox{\label{fig:visOnMHSConf_imp}}{
    \begin{tikzpicture}
        \begin{axis}[symbolic x coords={(b) Implicit},
            ymin=0.99, ymax=1.12, bar width= 11pt, width=0.35\linewidth,         yticklabel style={font=\tiny},
            ytick={1, 1.03, 1.06, 1.09, 1.12}]
            \addplot[SAFE-MEME-QA] coordinates {((b) Implicit,1)}; 
            \addplot[T5-large (FT)] coordinates {((b) Implicit,1.02)}; 
            \addplot[CLIP-XLM-R] coordinates {((b) Implicit,1.03)}; 
            \addplot[Pro-Cap (SingleCls)] coordinates {((b) Implicit,1.032)}; 
            \addplot[Pro-Cap (SimpleCls)] coordinates {((b) Implicit,1.066)}; 
            \addplot[HClipper(align-concat)] coordinates {((b) Implicit,1.068)}; 
            \addplot[PromptHate] coordinates {((b) Implicit,1.118)}; 
            
        \end{axis}
    \end{tikzpicture}
    }\hfill
    \subcaptionbox{\label{fig:visOnMHSConf_ben}}{
    \begin{tikzpicture}
        \begin{axis}[symbolic x coords={(c) Benign}, bar width= 12pt,         yticklabel style={font=\tiny},
            ymin=0.75, ymax=1.15, width=0.33\linewidth,
            ytick={0.75, 0.85, 0.95, 1.05, 1.15}]
            \addplot[T5-large (FT)] coordinates {((c) Benign,0.766)}; 
            \addplot[HClipper(concat)] coordinates {((c) Benign,0.877)}; 
            \addplot[QwenVL (FT)] coordinates {((c) Benign,0.967)}; 
            \addplot[SAFE-MEME-QA] coordinates {((c) Benign,1)}; 
            \addplot[Gemini 2.5] coordinates {((c) Benign,1.062)}; 
            \addplot[GPT-4o] coordinates {((c) Benign,1.137)}; 
        \end{axis}
    \end{tikzpicture}
    }
    \hfill
\subcaptionbox{\label{fig:visOnMHSConf_expSkew}}{
\begin{tikzpicture}
    \begin{axis}[
        symbolic x coords={(d) Explicit},
        xtick=data,
        ymin=-0.3, ymax=0.25,
        ytick={-0.3, -0.15, 0, 1, 0.25},
        yticklabel style={font=\tiny},
        ylabel={$\delta = P - R$},
        ylabel style={font=\tiny},
        width=0.33\linewidth,
        height=3.5cm,
        ybar=2pt,
        bar width=11pt,
        grid=major,
        enlarge x limits=0.4,
        clip=false,
        extra y ticks={0},
        extra y tick style={
            grid=major,
            grid style={black, thick, dashed}
        }
    ]
        \addplot[HClipper(concat)] coordinates {((d) Explicit, 0.223)};  
        \addplot[HClipper(align-concat)] coordinates {((d) Explicit, 0.151)}; 
        \addplot[SAFE-MEME-QA] coordinates {((d) Explicit, 0.154)}; 
        \addplot[QwenVL (FT)] coordinates {((d) Explicit, 0.159)}; 
        \addplot[T5-large (FT)] coordinates {((d) Explicit,-0.047)}; 
        \addplot[GPT-4o] coordinates {((d) Explicit,-0.091)}; 
        \addplot[Gemini 2.5] coordinates {((d) Explicit,-0.295)}; 
    \end{axis}
\end{tikzpicture}
}
    \hfill
\subcaptionbox{\label{fig:visOnMHSConf_impSkew}}{
\begin{tikzpicture}
    \begin{axis}[
        symbolic x coords={(e) Implicit},
        xtick=data,
        ymin=-0.35, ymax=0.15,
        ytick={-0.35, -0.15, 0, 0.15},
        yticklabel style={font=\tiny},
        width=0.33\linewidth,
        height=3.5cm,
        ybar=2pt,
        bar width=11pt,
        grid=major,
        enlarge x limits=0.4,
        clip=false,
        extra y ticks={0},
        extra y tick style={
            grid=major,
            grid style={black, thick, dashed}
        }
    ]
            \addplot[SAFE-MEME-QA] coordinates {((e) Implicit, 0.052)}; 
            \addplot[T5-large (FT)] coordinates {((e) Implicit, -0.161)}; 
            \addplot[CLIP-XLM-R] coordinates {((e) Implicit, -0.327)}; 
            \addplot[Pro-Cap (SingleCls)] coordinates {((e) Implicit, 0.068)}; 
            \addplot[Pro-Cap (SimpleCls)] coordinates {((e) Implicit, 0.079)}; 
            \addplot[HClipper(align-concat)] coordinates {((e) Implicit, -0.25)}; 
            \addplot[PromptHate] coordinates {((e) Implicit, 0.126)}; 
    \end{axis}
\end{tikzpicture}
}
    \hfill
\subcaptionbox{\label{fig:visOnMHSConf_benSkew}}{
\begin{tikzpicture}
    \begin{axis}[
        symbolic x coords={(f) Benign},
        xtick=data,
        ymin=-0.4, ymax=0.4,
        ytick={-0.4, -0.2, 0, 0.2, 0.4},
        yticklabel style={font=\tiny},
        ylabel style={font=\tiny},
        width=0.32\linewidth,
        height=3.5cm,
        ybar=2pt,
        bar width=11pt,
        grid=major,
        enlarge x limits=0.4,
        clip=false,
        extra y ticks={0},
        extra y tick style={
            grid=major,
            grid style={black, thick, dashed}
        }
    ]
            \addplot[T5-large (FT)] coordinates {((f) Benign,0.377)}; 
            \addplot[HClipper(concat)] coordinates {((f) Benign,-0.163)}; 
            \addplot[QwenVL (FT)] coordinates {((f) Benign, -0.373)}; 
            \addplot[SAFE-MEME-QA] coordinates {((f) Benign, -0.215)}; 
            \addplot[Gemini 2.5] coordinates {((f) Benign, 0.229)}; 
            \addplot[GPT-4o] coordinates {((f) Benign, 0.019)}; 

    \end{axis}
\end{tikzpicture}
}
            

    \ref{SharedLegend}
    \caption{\revised{Performance comparison of baseline models and $\model$ variants on $\datatwo$ across categories—explicit hate, implicit hate, and benign. Subplots (a)–(c) show normalized F1 scores relative to $\fvariant$ (computed as baseline / $\fvariant$), while subplots (d)–(f) illustrate precision–recall (P–R) skew, i.e., deviation from the $P=R$ line (c.f. Table \ref{tab:benchmarkComparison_MHSDBConandBaseline}). We report the top 15 percentile of baseline models, highlighting their comparative performance across scenarios.}}
    \label{fig:visOnMHSConf}
    \vspace{-5mm}
    \end{figure*}

\begin{keybox}[Key Takeaways]
\begin{itemize}
    \item $\svariant$ performs better than $\fvariant$ on $\dataone$.
    \item $\svariant$ improves by 3\% over the best performing open-source baseline.
    \item $\svariant$ lies within 1\% of the best performing closed-source baseline.
    \item Closed-source models are conservative on explicit hate,
    over-trigger on implicit hate. 
    \end{itemize}
\end{keybox}
\subsection{\textbf{Comparative Analysis on the \datatwo}} 
We compare the performance of M7 and M11, i.e., \fvariant\ and \svariant\, with the selected baselines: T5-large (FT), BLIP2 (FT), CLIP (FT), \revised{CLIP-XLM-R (FT), QwenVL (FT), GPT-4o, Gemini 2.5, HClipper (concat/align-concat), Pro-Cap (SimpleCls/SingleCls)} and PromptHate on the \datatwo\ dataset. The mentioned baselines are chosen due to their competitive performance in $\dataone$ (cf. Table \ref{tab:benchmarkComparison_MHSDBandBaseline}). The outcome of benchmarking the selected baselines along with M7 and M11 is presented in Table \ref{tab:benchmarkComparison_MHSDBConandBaseline}.

In \textit{explicit hate} cases, T5-large (FT), QwenVL (FT) and HClipper (concat/align-concat) achieve F1 scores of $0.648$, $0.611$, $0.540$ and $0.584$, respectively (cf. Table \ref{tab:benchmarkComparison_MHSDBConandBaseline}). M7 acts as a conservative classifier (cf. Figure \ref{fig:visOnMHSConf}(a)) and scores $0.602$, which creates a lag of $4.6\%$, $0.9\%$, $9.2\%$, $12.8\%$ against T5-large (FT), Qwen-VL (FT), GPT-4o, and Gemini 2.5, respectively but gains $1.8\%$ over HClipper (align-concat) (cf. Figure \ref{fig:visOnMHSConf}(a)). Among the variants of $\svariant$, M11 achieves an F1 score of $0.459$, and scores lower than most of the high-performing baselines. Note that T5-large (FT) serves as the most balanced classifier in this category (cf. Figure \ref{fig:visOnMHSConf}(d)).

In case of \textit{implicit hate}-based test cases, PromptHate scores the highest, $0.558$. It is followed by T5-large (FT), CLIP-XLM-R (FT), HClipper(align-concat), and Pro-Cap (SimpleCls) (cf. Figure \ref{fig:visOnMHSConf}(b)) have scores of $0.509$, $0.514$, $0.533$, and $0.532$, respectively (cf. Table \ref{tab:benchmarkComparison_MHSDBConandBaseline}). The proposed variant M7 from $\fvariant$ achieves an F1 score of $0.499$ and M11 of \svariant\ scores $0.488$. While M7 lags behind T5-large (FT), CLIP-XLM-R, HClipper (align-concat) and Pro-Cap(SimpleCls) by $1\%$, $1\%$, $3.4\%$, and $8.9\%$, respectively, it gains over GPT-4o and Gemini by $9.8\%$ and $5.5\%$, respectively. It is observed that the fine-tuned TSB models perform on par with, or better than, the pre-trained VLMs. In addition, their corresponding P-R skew scores are closer to zero (cf. Figure \ref{fig:visOnMHSConf}(e)) which suggests VLMs have an inherent limitation in effectively understanding implicit hate contexts. Similar to the performance on the explicit category, the recall scores of M7 and M11 for detecting implicit hate samples lag behind T5-large (FT) by $12.8\%$ and $14.1\%$, respectively, against T5-large (FT). Despite this lag, it is worth noting that their respective precision rates are distinctly better than those of T5-large (FT), while maintaining a competitive F1 score. 

For test cases of the \textit{benign} category, the T5-large (FT) score is $0.58$, while the scores of Qwen-VL (FT), GPT-4o (most balanced), Gemini, and HClipper (concat) are $0.732$, $0.861$, $0.804$, and $0.664$, respectively (cf. Figure \ref{fig:visOnMHSConf}(c) and (f)). M7, with an F1-score of $0.757$, is ahead of T5-large (FT), QwenVL (FT), and HClipper (concat) by $17.7\%$, $2.5\%$, and $9.3\%$, respectively, but falls behind GPT-4o and Gemini 2.5 by $10.4\%$ and $4.7\%$ (cf. Table \ref{tab:benchmarkComparison_MHSDBConandBaseline}). Variant M11 with, an F1 score of $0.548$, fails to outperform the top-scoring baselines.  Note that while the T5-large (FT) model achieves high precision ($0.824$), its low recall ($0.447$) results in a lower F1 score of $0.580$. In contrast, the proposed model variants M7 and M11 demonstrate a comparatively balanced \textit{overall} performance, with the better variant M7 (\fvariant) exhibiting decent precision and recall scores of $0.664$ and $0.879$, respectively, yielding a good F1 score of $0.757$.

Among various implementations of $\fvariant$ and $\svariant$, only the M7 variant demonstrates superior performance compared to the selected baseline models on \datatwo. In contrast, M11, the variant of $\svariant$, outperforms the multimodal baseline, highlighting their limited comparative advantage in this setup. There are \revised{\textit{three}} key observations to consider. First, M7 exhibits performance comparable to M11 when evaluated on dataset $\dataone$. \revised{Second}, under a rigorous stress test using \datatwo, M7 significantly outperforms M11 \revised{and lastly, despite its unimodal nature, T5-large (fine-tuned) outperforms other multimodal baseline models. This competitive performance is attributed to balanced results in the explicit hate category, along with moderately high recall for implicit hate and high precision for the benign category. Note that 64\% of the test cases of $\datatwo$ contain explicit hate expressions or demographic groups (in contrast to $\dataone$ with 32.4\%), and the model correctly classifies 85.8\% on these instances. This indicates that the textual content alone provides sufficient cues for a correct classification. In contrast, for test samples involving implicit hate, it demonstrates a disparity between a higher recall (0.602) and a lower precision (0.441).  Approximately 95.8\% of the error cases in benign test cases where T5-large (FT) assigned an implicit label, contain at least one mention of target-group(s) or hate-related keywords i.e., a lexical bias, due to which the model tends to over-predict the implicit hate class, thereby sacrificing precision to achieve higher recall. It is validated for benign test cases by the exceptionally high precision (0.824) and a comparatively low recall of 0.447. It confirms that T5-large(FT) is conservative and highly accurate for unambiguously positive or neutral text. Further details regarding the performance analysis of T5-large(FT) are presented in Section \ref{app:analysisOfT5_onMHS}}. 

\begin{keybox}[Key Takeaways]
\begin{itemize}
    \item $\fvariant$ outperforms $\svariant$ across all three categories on $\datatwo$.
    \item $\fvariant$ improves by 4.7\% macro-F1 over the best open-source baseline.
    \item $\fvariant$ trails behind the closed-source models, GPT-4o and Gemini 2.5.
    \item T5-large (FT) stays competitive due to the presence of explicit lexical cues in $\datatwo$.
\end{itemize}
\end{keybox}
 

\begin{table*}[t!]
\centering
\small
\caption{Ablation Study: It compares the performance of different variations and components of $\fvariant$ and $\svariant$ on $\dataone$ (P: precision, R: recall and F: F1 score).}
\label{tab:benchmarkComparison_FVandSVonMHS_DB}
\resizebox{\textwidth}{!}{
\begin{tabular}{@{}ccccc|ccc|ccc|ccc@{}}
\toprule
\multirow{2}{*}{\textbf{Model}} &
  \multirow{2}{*}{\textbf{Variant}} &
  \multicolumn{3}{c}{\textbf{Explicit Hate}} &
  \multicolumn{3}{c}{\textbf{Implicit-Hate}} &
  \multicolumn{3}{c}{\textbf{Benign}} &
  \multicolumn{3}{c}{\textbf{Macro}} \\ \cmidrule(l){3-14} 
 &
   &
  \textbf{P} & \textbf{R} & \textbf{F} & \textbf{P} & \textbf{R} & \textbf{F} & \textbf{P} & \textbf{R} & \textbf{F} & \textbf{P} & \textbf{R} & \textbf{F} \\ \midrule

\multirow{8}{*}{\rotatebox{90}{$\fvariant$}}  &  M0     & 0.316 & 0.696 & 0.435 & 0.492 & 0.536 & 0.513 & 0.615 & 0.222 & 0.329 & 0.474 & 0.485 & 0.425\\     
  & M1   & 0.343 & 0.759 & 0.472 & 0.544 & 0.428 & 0.479 & 0.599 & 0.404 & 0.482 & 0.495 & 0.53 & 0.478\\
  &  M2 & 0.356 & 0.692 & 0.470 & 0.53 & 0.535 & 0.532 & 0.629 & 0.349 & 0.449 & 0.505 & 0.525 & 0.488 \\
  &  M3   & 0.387 & 0.581 & 0.465 & 0.537 & 0.388 & 0.451 & 0.542 & 0.573 & \textbf{0.557} & 0.489 & 0.514 & 0.491 \\
  & M4  & 0.391 & 0.581 & 0.467 & 0.535 & 0.388 & 0.450 & 0.541 & 0.573 & 0.556 & 0.489 & 0.514 & 0.490 \\
  &  M5   & 0.390 & 0.589 & 0.469 & 0.538 & 0.383 & 0.447 & 0.541 & 0.575 & \textbf{0.557} & 0.490 & 0.515 & 0.491 \\
  & M6   & 0.391 & 0.595 & 0.472 & 0.540 & 0.382 & 0.448 & 0.539 & 0.573 & 0.555 & 0.49 & 0.517 & 0.492 \\
  & M7    & 0.391 & 0.669 & 0.494 & 0.58 & 0.554 & \textbf{0.571} & 0.65 & 0.483 & 0.554 & 0.543 & 0.569 & \textbf{0.539} \\ 
\midrule

\multirow{6}{*}{\rotatebox{90}{$\svariant$}}  & M8    & 0.425 & 0.689 & 0.525 & 0.813 & 0.419 & 0.552 & 0.554 & 0.264 & 0.358 & 0.597 & 0.457 & 0.478 \\
 & M9   & 0.378 & 0.264 & 0.37 & 0.714 & 0.678 & \textbf{0.696} & 0.519 & 0.206 & 0.295 & 0.537 & 0.382 & 0.433 \\
 & M10   & 0.450 & 0.547 & 0.494 & 0.784 & 0.536 & 0.636 & 0.332 & 0.401 & 0.363 & 0.522 & 0.495 & 0.497 \\
 & M11    & 0.588 & 0.594 & \textbf{0.587} & 0.784 & 0.536 & 0.636 & 0.589 & 0.323 & 0.417 & 0.653 & 0.485 & \textbf{0.546} \\
   &  M12    & 0.489 & 0.581 & 0.530 & 0.799 & 0.527 & 0.635 & 0.566 & 0.341 & \textbf{0.426} & 0.617 & 0.483 & 0.530 \\
 &   M13   & 0.426 & 0.703 & 0.531 & 0.818 & 0.419 & 0.554 & 0.568 & 0.267 & 0.364 & 0.603 & 0.462 & 0.482 \\
\bottomrule

\end{tabular}
}

\end{table*}

\subsection{Ablation Results: Insights from Intra-Variant Performance}

\noindent \paragraph{\textbf{Variants of $\fvariant$}} Among the variants of $\fvariant$, M7 is the strongest, achieving a macro-F1 of 0.539 ( macro-F1 of 0.494, 0.571, and 0.554 in explicit, implicit, and benign categories,  respectively) on $\dataone$ (cf. Table \ref{tab:benchmarkComparison_FVandSVonMHS_DB}) and 0.619 ( macro-F1 of 0.602, 0.499, and 0.757 in explicit, implicit, and benign categories,  respectively) on $\datatwo$ (cf. Table \ref{tab:benchmarkComparison_MHSDBConandBaseline}). Variant M7 has a lead of up to
11.4\% over the weakest variant, i.e., M0.

\noindent \paragraph{\textbf{Variants of $\svariant$}} Among the $\svariant$ variants, M11 performs best with a macro-F1 score of 0.546 on $\dataone$ (cf. Table \ref{tab:benchmarkComparison_FVandSVonMHS_DB}). Notably, M11 uses a pre-trained T5-base with a category-specific linear projector, yet matches or surpasses its
fully fine-tuned counterpart M10 on $\datatwo$, which fine-tunes the T5-base with a category-specific linear projector, outperforming it by 9.3\% on explicit hate and 5.4\% on benign. This suggests that a lightweight single-layer adapter can rival fully fine-tuning for fine-grained meme classification. This trend holds on $\datatwo$, where M11 beats M10 on implicit and benign categories and by 4.1\% on macro-F1, though fully fine-tuning (M10)
retains an edge on explicit hate.\footnote{See Section $\ref{app:modelAblation_onMHS}$ for more details on intra-variant analysis of $\model$.} 

\begin{keybox}[Key Takeaways]
\begin{itemize}
    \item $\fvariant$ (M7) is the strongest variant on both datasets.
    \item $\svariant$ (M11) is the strongest variant on $\dataone$.
    \item The single-layer adapter of M11 (over a pre-trained T5-base) attains performance comparable or superior to the fully fine-tuned M10.
\end{itemize}
\end{keybox}



\section{Qualitative Analysis}

This section provides a systematic analysis of the key factors of the proposed model, $\model$, across both of its variants. We first examine how the interplay between the visual and textual modalities contributes to the observed outcomes, and subsequently investigate the  error cases, thereby offering the strengths and limitations of our approach.

\paragraph{\textbf{Qualitative Analysis for $\fvariant$}} 

Note that the distribution of target category-specific hate is not uniform in the $\dataone$ dataset (cf. Figure \ref{fig:RMMHS_hate_distribution}), which constrains the generalization capacity of $\fvariant$. Therefore, factors like the distribution of hate categories in \dataone\ play a prominent role in our qualitative assessment of $\fvariant$. 
\begin{figure}[t!]
\centering
\includegraphics[width=1\textwidth]{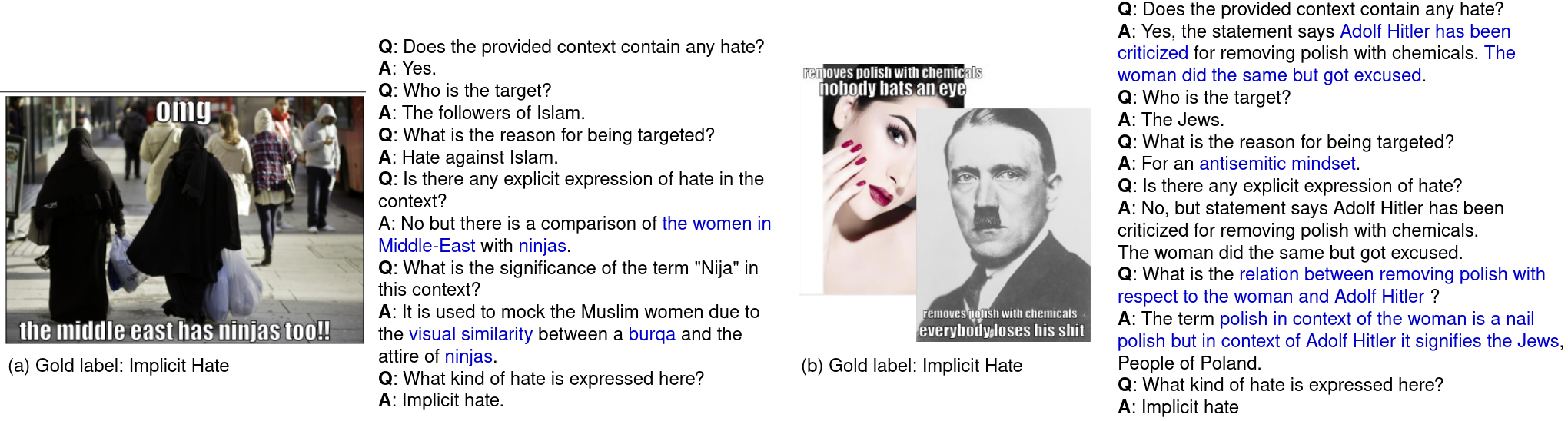}
\caption{An illustration of the generated outcomes of $\fvariant$: (a) for the \textit{high} and (b) for \textit{medium} representation content target categories. Text in \textcolor{blue}{blue} indicates important contextual information generated by the model.}  
\label{fig:qualitativeAnalysis_HAndL}
\end{figure}
We focus on the precision of the generated questions and responses for this study. Test cases are selected from to each of the \textit{high}, \textit{medium}, and \textit{low} representation categories in \dataone. This approach ensures a balanced observation across the distribution spectrum of target categories. Figure \ref{fig:qualitativeAnalysis_HAndL}(a) represents an outcome generated by $\fvariant$. It belongs to \textit{Islam} category, having \textit{high} representation within the training corpus. The context metaphorically addresses a group of Muslim women in \textit{burqa} as \textit{ninja}s. 
Initially, the model identifies a visual reference to mockery targeting the women, despite the absence of explicit cues. Subsequently, it reveals targeted hostility toward Muslim women, indicating an underlying Islamophobic intent. Finally, the required label is extracted using a regular expression.


Figure \ref{fig:qualitativeAnalysis_HAndL}(b) represents an analysis of $\fvariant$ targeting the \textit{Jewish} community which belongs to the \textit{medium} representation category. 
The example consists several ambiguous points that need clarification to understand the actual context. While the context highlights a woman and Adolf Hitler, it implicitly references a third entity: Polish citizens. The analysis highlights the dual reading of polish and remove: literally, removing nail polish for the woman; and, through the reference to Adolf Hitler, the removal of Polish Jews during the Holocaust. The target group is revealed to be the Jewish community, reflecting antisemitic sentiment. This instance is classified as \textit{implicit hate} due to the absence of explicit hate cues.


\begin{figure}[!t]
\centering
\includegraphics[width=1\textwidth]{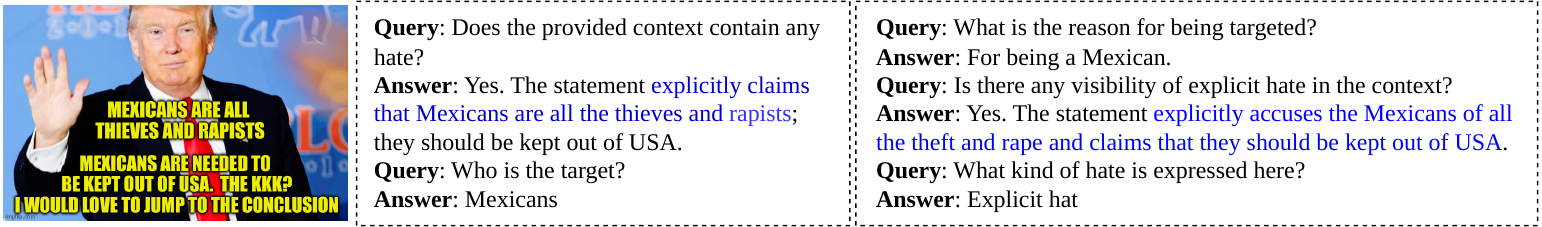}
\caption{Outcomes of $\fvariant$ for the low representation content target categories. Text in \textcolor{blue}{blue} indicates important contextual information generated by the model.}
\label{fig:qualitativeAnalysis_lowResource}
\end{figure}

Figure \ref{fig:qualitativeAnalysis_lowResource} represents an outcome of $\fvariant$ for the \textit{Immigrant} target group in the low-representation category. It features Donald Trump with overlaid text accusing Mexicans of various crimes. $\fvariant$ effectively identifies the target group, claims, and provocative remarks. Finally, the instance is declared as explicit hate due to the presence of explicit accusations. Despite the notable output quality, $\fvariant$ demonstrates a critical limitation: it hallucinates, irrespective of the underlying hate-target category distribution. Although the impact is minimal for categories with higher representation, we notice unrelated information in the outputs. Unfortunately, in low-representation scenarios, the prominence of irrelevant insights distorts the essence of the context.

\paragraph{\textbf{Qualitative analysis for $\svariant$}}
\begin{figure}[t!]
\centering
\includegraphics[width=1\textwidth]{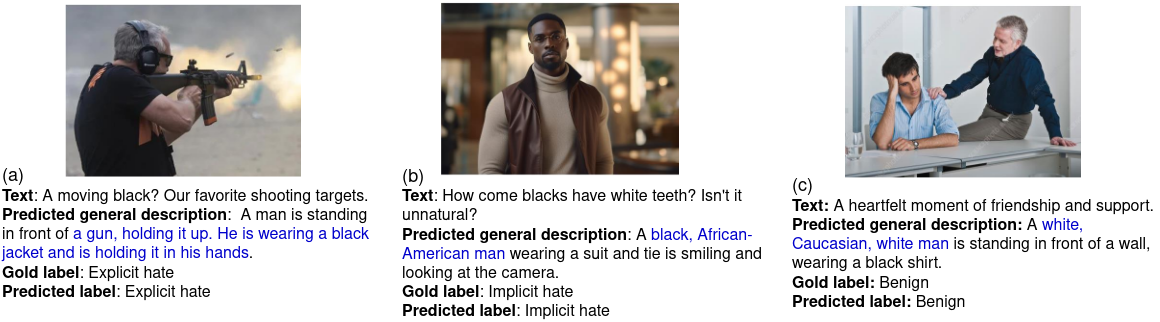}
\caption{An illustration of predictions of M11 from $\svariant$ for all categories of hatefulness -- (a) explicit, (b) implicit and (c) benign. The text in \textcolor{blue}{blue} indicates a correct, important visual aspect of the corresponding entity.}
\label{fig:qualitativeAnalysis_forSV}
\end{figure}
The primary objective of $\svariant$ is to generate a detailed visual description followed by a hierarchical classification. Figure \ref{fig:qualitativeAnalysis_forSV} illustrates \textit{three} cases -- explicit, implicit, and benign. Figure \ref{fig:qualitativeAnalysis_forSV}(a) illustrates an explicit expression of hate in which people of the African-American community are targeted. Also, the visual description accurately notes that an individual wearing a black shirt is depicted holding a firearm. This results in a correct label prediction. Figure \ref{fig:qualitativeAnalysis_forSV}(b) serves as an example where no explicit expressions or hints of hate are found. Still, the context implicitly directs sarcastic remarks toward individuals of the African-American community based on their skin color. The model can correctly describe the visual attributes of the individual and predicts the correct label. Likewise, in Figure \ref{fig:qualitativeAnalysis_forSV}(c), $\svariant$ correctly identifies the individuals as ``white'' and ``Caucasian.''  


\begin{figure}[t!]
\centering
\includegraphics[width=1\textwidth]{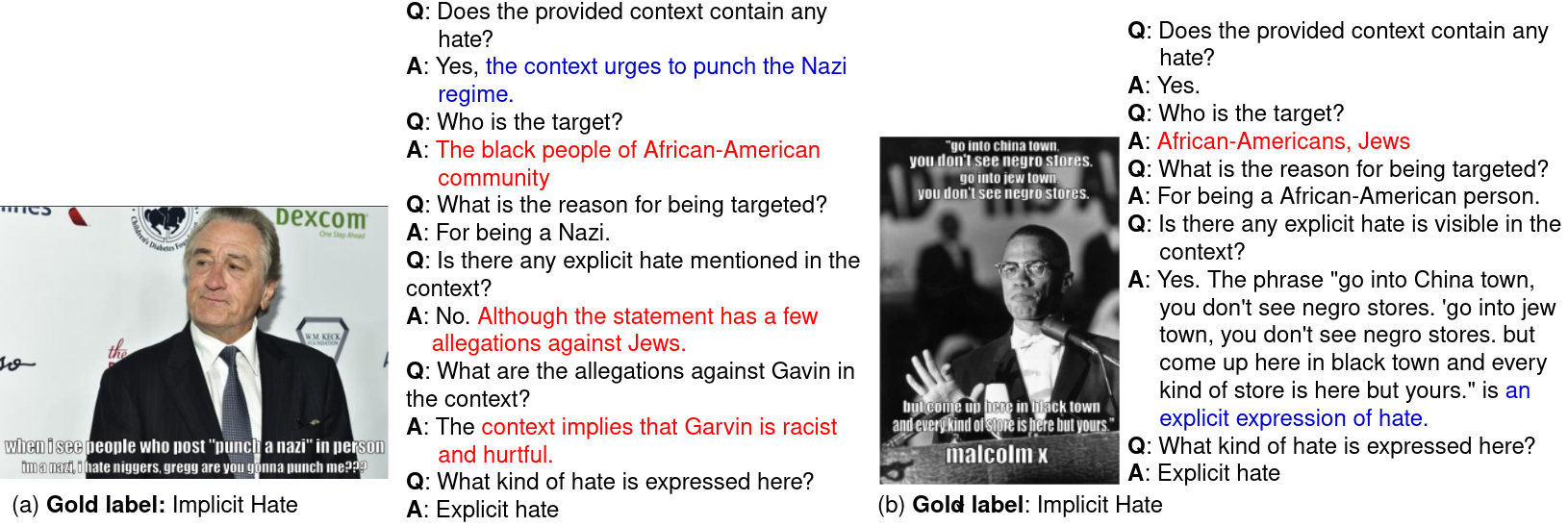}
\caption{An illustration of errors in classification by $\fvariant$: (a) $\fvariant$ incorrectly identifies it as an explicit hate instance for overt advocacy for violent actions against the Nazi regime and fails to recognize the actual target group the \textit{Jewish} community, (b) $\fvariant$ identifies African Americans and Jews as the target groups, with African Americans depicted as victims. In both cases, the offensive racial slurs are highly correlated with the corresponding communities, causing the misclassification.} 
\label{fig:err_QA_14_15_merged}
\end{figure}



\paragraph{\textbf{Error Analysis of $\fvariant$}} Figure \ref{fig:err_QA_14_15_merged}(a) illustrates a case where $\fvariant$ retains contextual cues but misattributes hate toward Nazis rather than the Black community. Here, the model exhibits a fragmented reasoning path, shifting focus across loosely connected entities -- from Nazi supporters to the African-American community to speculative references about Jews and an unrelated individual (``Gavin''). This inconsistency reflects a breakdown in maintaining referential coherence and in correctly identifying the target of hate. This is compounded by the fact that the model disregards statistical patterns in the dataset -- where implicit hate more frequently involves the Black community and associated slurs -- by overemphasizing peripheral cues like ``Nazi'' and inferred allegations against Jews. These conflated associations lead the model to misclassify the meme as expressing explicit hate, despite statistically stronger indicators of implicit hate. The error underscores the model's limitations in reasoning about nuanced, indirect hate expressions and its insensitivity to contextual and distributional cues.

Figure \ref{fig:err_QA_14_15_merged}(b) further exemplifies the sensitivity of $\fvariant$ to the phrases associated with a target group. The context highlights the marginalization of the African-American community by contrasting with other communities like Jewish and Chinese. As in the example shown in Figure \ref{fig:err_QA_14_15_merged}(a), although $\fvariant$ recognizes the references to different religious and ethnic communities like Chinese, Jews and African Americans, it erroneously conflates their reference to explicit hate speech, when it answers the question of whether explicit hate is apparent in the context. Besides demonstrating good capabilities of the proposed approach in recognizing the concepts and entities being referred to within the meme, such examples highlight the gap in their reasoning capabilities, suggesting a suboptimal balance between the breadth of contextual vectors and the depth of logical compositionality, especially within complex artifacts like memes.

The error analysis highlights several limitations of the proposed approaches (M7 and M11), including incoherent reasoning and training corpus dependencies, as well as architectural constraints specific to M11 on certain protected categories. These limitations underscore the outstanding challenges in advanced multimodal reasoning for memetic analysis.\footnote{See Section~\ref{app:limitations} for more details on the limitations of the proposed methodology.}

\begin{figure}[t!]
\centering
\includegraphics[width=1\textwidth]{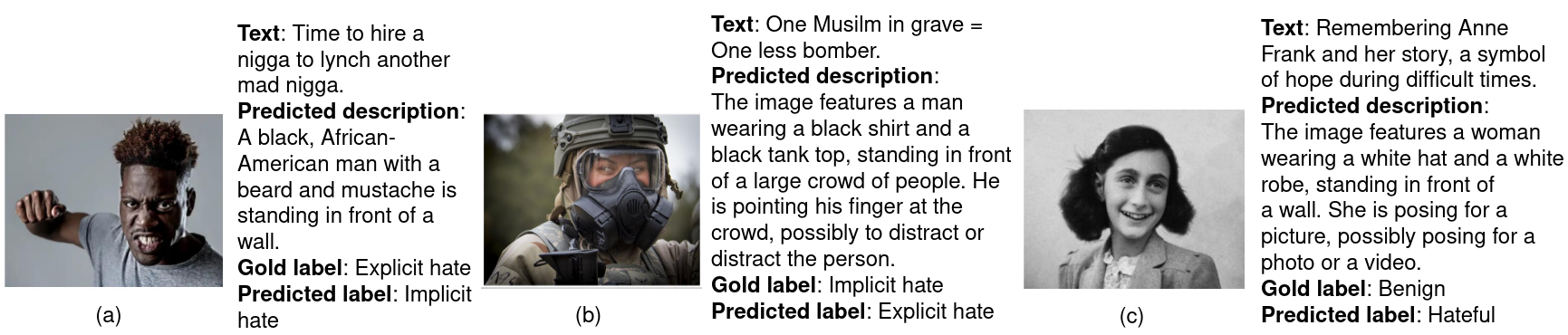}
\caption{An illustration of error cases in classification by $\svariant$. Subfigures (a) and (b) show that content targeting African Americans (more frequently represented within the implicit hate category) and Islamic communities (predominantly associated with the explicit hate category) is mislabeled due to differing distributions of explicit and implicit hate in the $\dataone$ dataset. Subfigure (c) demonstrates a similar error for non-hateful content about Anne Frank, attributed to its frequent association with explicit and implicit instances in $\dataone$.} 
\label{fig:errorAnalysisSV_1}
\end{figure}

\paragraph{\textbf{Error Analysis of $\svariant$}} Figure \ref{fig:errorAnalysisSV_1} presents an analysis of error cases attributed to $\svariant$. Figure  \ref{fig:errorAnalysisSV_1}(a) highlights a meme explicitly targeting the African-American community with accusations of violence, while Figure \ref{fig:errorAnalysisSV_1}(b) shows content implicitly aiming to humiliate the Islamic community. In both instances, $\svariant$ fails to identify the hate category, leading to misclassification. Upon closer analysis, it seems that in the case of the Black community, $\svariant$ attempts to assign the label `implicit hate', whereas, for hate targeting Islam, it predominantly attempts to assign the label `explicit hate'; one reason could be the underlying training instances of $\dataone$ where the number of instances associated with Islam exhibit a higher prevalence of explicit hate than implicit hate. The opposite trend is observed for the Black community. A similar phenomenon is observed in Figure \ref{fig:errorAnalysisSV_1}(c), where non-hateful content related to Anne Frank is misclassified. This misprediction arises from the higher occurrence of Anne Frank-related content in $\dataone$ under both explicit and implicit categories, leading $\svariant$ to incorrectly label such content as hateful.

\begin{keybox}[Key Takeaways]
\begin{itemize}
    \item $\fvariant$, and $\svariant$ inherit the training distribution's bias.
    
    \item $\fvariant$ hallucinates on rare categories.

    \item $\fvariant$ conflates slurs with explicit hate.
    
    \item $\svariant$ tracks category priors, and high-frequency entities.    
\end{itemize}
\end{keybox}

\section{Conclusion}
We introduced two datasets, $\dataone$ and $\datatwo$, for benchmarking fine-grained hate speech detection in high-quality memes and stress-testing the robustness of competing methods. Our proposed model, \model, leverages a novel multimodal CoT-based framework with two modules: a Q\&A-style structured reasoning module ($\fvariant$) and a hierarchical categorization module ($\svariant$), optimized for $\dataone$. Our experiments show that structured reasoning enables $\fvariant$ to outperform open-source baselines while closely tracking the closed-source models, GPT-4o and Gemini~2.5. Notably, a lightweight single-layer adapter within $\svariant$ outperforms the other variant $\fvariant$ on $\dataone$, whereas the Q\&A-based $\fvariant$ proves more effective under the confounding stress test on $\datatwo$. An interesting observation across both datasets is the strong performance of the text-only T5-large (FT) baseline, especially on $\datatwo$, where explicit textual cues are more common. This underscores the value of structured, context-aware approaches such as $\model$ that aim to reason over hidden context rather than surface-level group mentions. However, both approaches reflect training bias. Future research can address the limitations highlighted in this study by expanding and diversifying the dataset to include a broader range of hate categories, involving annotators from varied backgrounds, and improving the model by grounding the reasoning process with verifiable external evidence and incorporating human-in-the-loop feedback.

\begin{keybox}[Summary of Findings]
\begin{itemize}
    \item $\fvariant$ outperforms open-source baselines on both $\dataone$, and $\datatwo$.

    \item T5-large (FT) stays competitive on $\datatwo$ due to explicit lexical cues.
    
    \item Closed-source VLMs over-predict implicit hate, yielding high recall but low precision.

    \item Both variants of $\model$ inherit bias from the training data.
\end{itemize}
\end{keybox}

\section{Limitations} \label{app:limitations}

Despite the strong performance demonstrated by $\fvariant$ and \svariant, a few limitations persist. We will systematically discuss the shortcomings of the proposed model and dataset $\dataone$.

\paragraph{\textbf{Dataset}} Although $\dataone$ consist of nine distinct categories, it does not provide exhaustive coverage of all potential target groups. Therefore, a relevant target-group may be underrepresented or entirely absent in the dataset. It may limit the generalizability of the models. It is also not comprehensive in terms of breadth of data instances. This flaws underscore the requirement for more inclusive datasets that reflect the full spectrum of hate-targeted communities.

\paragraph{\textbf{Models}} We acknowledge that the proposed approach reflects certain shortcomings, particularly related to potential biases introduced in the training process. The scores in the explicit and implicit hate categories of MHS-Con (see Table \ref{tab:benchmarkComparison_MHSDBConandBaseline}) reveals that T5-large (FT) consistently outperforms both of variant, $\fvariant$ and  $\svariant$. However, we would like to underscore $\fvariant$’s (M7) balanced performance observed across the three categories as part of the MHS dataset (Table 4), where other baselines (including T5-large [FT]) yield skewed category-specific performances.

To inspect the generalizability of $\svariant$ (M11) over different corpus, we evaluate its performance on the \textit{Harm-P} test set \cite{pramanick-etal-2021-momenta-multimodal}, which contains memes on US Politics, and is realistic. While M11 is trained on the MHS dataset to classify the given instance into one of the following three categories: explicit hate, implicit hate, and benign, the Harm-P dataset uses a different labelling scheme: \textit{Very Harmful}, \textit{Partially Harmful}, and \textit{Not Harmful}. In order to conduct a fair evaluation, we emulated categorization for comparable categories by focusing only on instances labeled as \textit{Not Harmful} and \textit{Very Harmful}, as a proxy for benign and explicit hate, respectively, in \textit{Harm-P}. We observed that M11 effectively identifies \textit{Very-Harmful} (explicit) instances with a F1 score of  $0.86$, but fails to classify \textit{Non-Harmful} (benign) instances (F1 is $0.12$), suggesting potential limitations in its generalizability to different datasets.

\begin{figure}[!t]
\centering
\includegraphics[width=.7\textwidth]{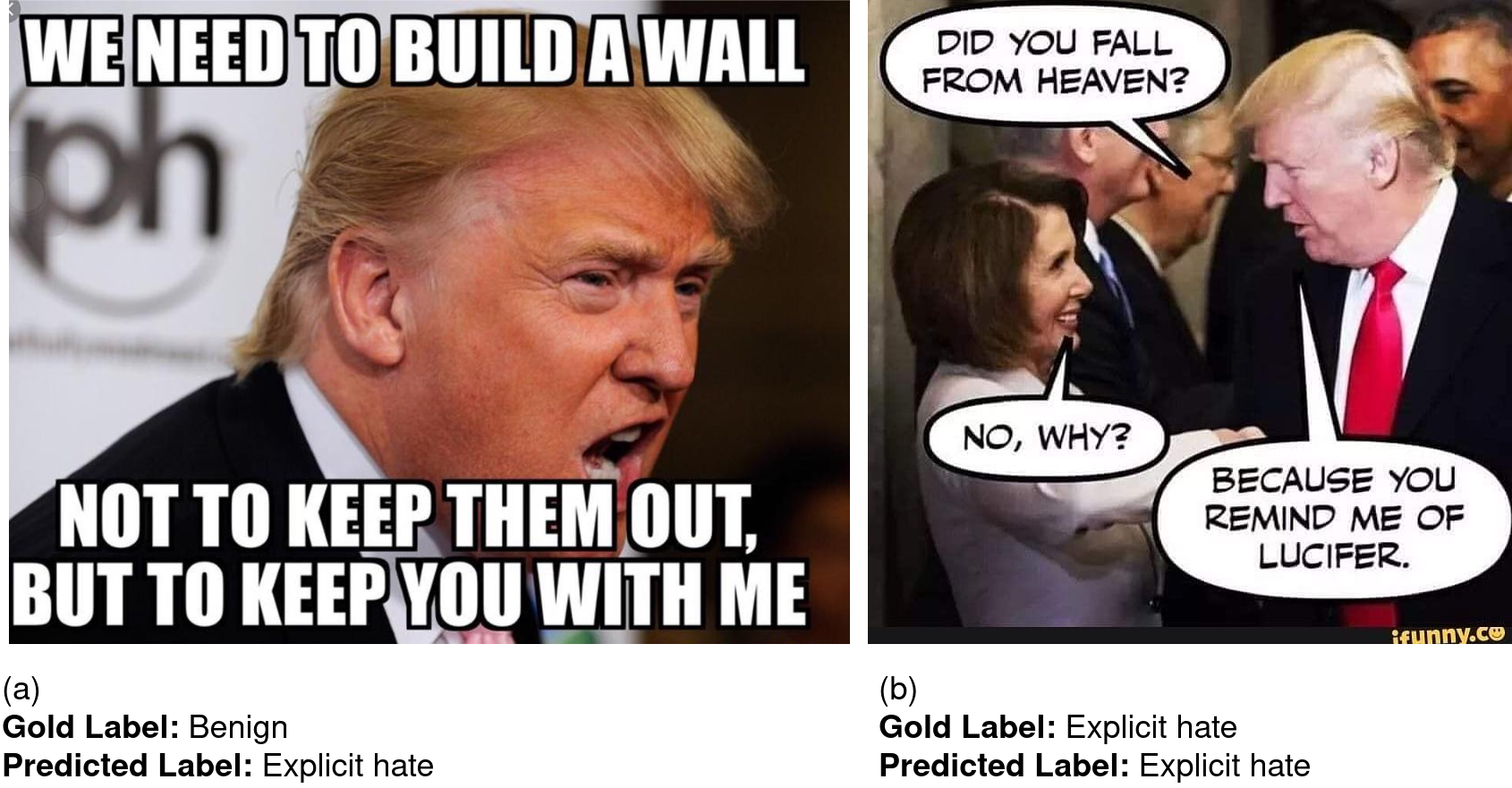} 
\caption{An illustration of training bias in $\svariant$. Subfigure (a) presents an instance featuring President Donald Trump making a seemingly benign statement about building a wall for protection. Despite its non-hateful intent, the model \textit{incorrectly} classifies it as explicit hate. Conversely, in Subfigure (b), President Trump compares a female political person to ‘Lucifer’, which is accurately identified as explicit hate. This contrast highlights potential biases in the training data that may lead to misclassifications.}
\label{fig:training_bias}
\end{figure}

As demonstrated in Figure \ref{fig:training_bias}(a) a meme featuring President Donald Trump appearing to explain the importance of constructing a border wall—not as a measure to prevent immigration but to ensure the safety of citizens, has been incorrectly flagged as hateful. In contrast, in another instance in the Very-Harmful category, involving President Donald Trump addressing a fellow female politician as ‘Lucifer’ is correctly labeled (c.f. Figure \ref{fig:training_bias}(b)), likely reflecting biases present in the training data. The limitation to the generalizability of M11 ($\svariant$) is also evident from the drop in its performance when evaluated on the MHS-Con dataset, after having been trained on MHS (c.f. Table \ref{tab:benchmarkComparison_MHSDBConandBaseline}). It is important to note that the underlying protected categories within Harm-P are individual, community, organisation and society, which represent distinct entities from the ones considered as part of MHS/MHS-Con. Based on these observations, it would be fair to conclude that, given the dependency of M11’s architecture on the required protected categories, the performance on a setting that is different from the one the model is trained on could be sub-optimal.

\section*{Acknowledgments}
The authors acknowledge the financial support of the Anusandhan National Research Foundation (CRG/2023/001351). Tanmoy Chakraborty acknowledges the support of the Google GCP Grant and the Rajiv Khemani Young Faculty Chair Professorship in Artificial Intelligence.

\bibliographystyle{IEEEtranN}
\small
\bibliography{refs}

\newpage
\appendix

\section{Constraints in the Existing Resources}\label{app:limitations_on_existingResources}

\begin{figure}[!t]
\centering
\includegraphics[width=1\textwidth]{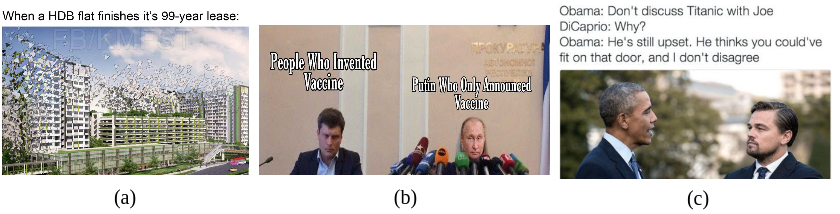} 
\caption{\revised{Examples illustrating shortcomings in existing annotations across three meme datasets: (a) A meme from the TotalDefMeme dataset related to Singapore’s Total Defence Policy, where multiple pillars (e.g., Psychological and Economic) are assigned stance labels without sufficient contextual explanation. (b) An instance from the ExHVV dataset with a brief single-line explanation that does not adequately justify why Vladimir Putin is portrayed as receiving disproportionate attention compared to the vaccine inventors. (c) A sample from the DISARM dataset labeled as “not harmful” without any supporting rationale, highlighting the lack of detailed annotation guidance. }}
\label{fig:limitations_annotation}
\end{figure}

\subsection{\revised{Limitations in Annotation Schemas }}
\textit{TotalDefMeme} contains memes related to Singapore’s \textit{Total Defence Policy} \cite{Prakash_2023}, which consists of six distinct pillars: Military, Civil, Economic, Social, Psychological, and Digital. Each pillar represents an important area of national strategy and contributes to the overall development and security of the country. The memes in the dataset are annotated according to their stance (\textit{support, against, or neutral}) towards the relevant pillar. Figure \ref{fig:limitations_annotation}(a) shows an example of an annotated meme in which psychological and economic defenses are marked as “against” and “neutral”, respectively. It is very difficult for the readers to understand the context unless they are familiar with the implicit information. Likewise, an instance from the ExHVV dataset \cite{Sharma_Agarwal_Suresh_Nakov_Akhtar_Chakraborty_2023} (cf. Figure \ref{fig:limitations_annotation}(b)) contains only a single-line explanation: “people who invented the COVID-19 vaccine are portrayed as being ignored”. This explanation is insufficient to justify why Vladimir Putin is shown as receiving excessive attention, while the actual vaccine inventors are not.  Figure \ref{fig:limitations_annotation}(c) presents an example from the DISARM dataset \cite{sharma-etal-2022-disarm} with a similar limitation. In this case, the sample is labeled as “not harmful”; however, no justification or explanation is provided for this annotation.

\subsection{Limited Ability of VLMs to Interpret Sarcasm}
\begin{figure}[!t]
\centering
\includegraphics[width=1\textwidth]{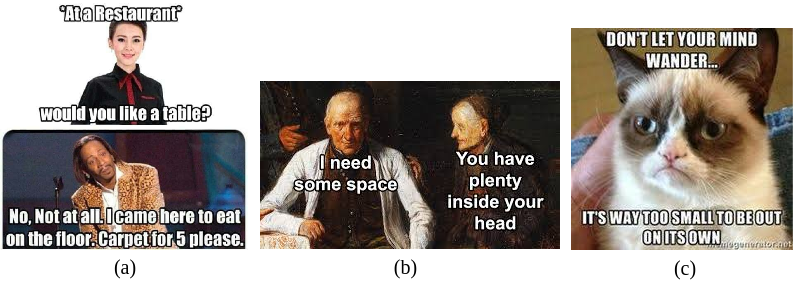} 
\caption{Examples of meme instances that contain sarcasm within their context.}
\label{fig:sarcasm_annotation}
\end{figure}

Here, we discuss the ability of VLMs to understand a context in the presence of sarcasm. Figure \ref{fig:sarcasm_annotation}(a)--(c) show the sarcastic memes that are utilized for evaluation. We employ open-source models, namely llava-1.5-(7b\footnote{\url{https://huggingface.co/llava-hf/llava-1.5-7b-hf/discussions/7}}/13b\footnote{\url{https://huggingface.co/llava-hf/llava-1.5-13b-hf/discussions/7}})-hf and Qwen2.5-VL-7B-Inst\footnote{\url{https://huggingface.co/Qwen/Qwen2.5-VL-7B-Instruct}}, as well as proprietary models, including GPT-4o\footnote{\url{https://chatgpt.com/}} and Gemini 2.5\footnote{\url{https://gemini.google.com/app}}. The generated outcomes are presented below, 

\begin{tcolorbox}[breakable, colback=gray!10, colframe=black, boxrule=0.7pt, arc=3pt, left=6pt, right=6pt, top=6pt, bottom=6pt,fontupper=\small, colupper=black,
    ]
\textbf{Model}:  llava-1.5-7b-hf
\\ \text{}\\
\textbf{Input Image}: Figure \ref{fig:sarcasm_annotation}(a) \\
\textbf{Input Query}: Explain the meme briefly. \\
\textbf{Response}: The meme features a man and a woman in a restaurant setting. The man is sitting ... is making a funny face, and the woman is smiling. The meme is captioned with a humorous statement, suggesting that the man is \textcolor{red}{not interested} in the woman's table. \\
\textbf{Issue(s)}: Model considers the absurd response as a sincere statement, which misses out the sarcastic contradiction. 
\\ \text{}\\
\textbf{Input Image}: Figure \ref{fig:sarcasm_annotation}(b) \\
\textbf{Input Query}: Explain the meme briefly. \\
\textbf{Response}: The meme features \textcolor{red}{two men} sitting next to each other, with one of them saying, "You need more space." The other man is looking at him, \textcolor{blue}{seemingly in disagreement}. The meme is likely a humorous or sarcastic commentary on the concept of \textcolor{red}{personal space or the need for more space in a shared environment}.\\
\textbf{Issue(s)}: Model incorrectly identifies the woman and misinterprets the term `space' at its face value and generalizes it into a commentary on personal space. It causes the model to fail to recognize the sarcasm.. 
\\ \text{}\\
\textbf{Input Image}: Figure \ref{fig:sarcasm_annotation}(c) \\
\textbf{Input Query}: Explain the meme briefly. \\
\textbf{Response}: The meme features a picture of a cat that appears to be looking at the camera, and the caption is a play on words, suggesting that \textcolor{red}{the cat's mind is too small to be left alone}. The meme is likely meant to be humorous and \textcolor{red}{relatable to pet owners who have experienced their pets wandering off or getting into mischief}.\\
\textbf{Issue(s)}: Model misinterprets the terms `mind' and `small' and associates them with pet behavior rather than identifying the sarcastic insult.
\end{tcolorbox}

\begin{tcolorbox}[breakable, colback=gray!10, colframe=black, boxrule=0.7pt, arc=3pt, left=6pt, right=6pt, top=6pt, bottom=6pt, fontupper=\small, colupper=black, fontupper=\small, colupper=black,
]
\textbf{Model}: Qwen2.5-VL-7B-Inst
\\ \text{}\\
\textbf{Input Image}: Figure \ref{fig:sarcasm_annotation}(a) \\
\textbf{Input Query}: Explain the meme briefly. \\
\textbf{Response}: This meme humorously depicts a scenario  at a restaurant where a customer is asked if they would like a table. \textcolor{blue}{The joke plays on the absurdity of the request}, suggesting that the customer wants to \textcolor{red}{eat directly on the carpet rather than using a table}, adding a layer of unexpected humor to the situation.\\
\textbf{Issue(s)}: Model misses the point that the man orders `5 carpets' for food, which creates a sarcastic remark.
\\ \text{}\\
\textbf{Input Image}: Figure \ref{fig:sarcasm_annotation}(b) \\
\textbf{Input Query}: Explain the meme briefly. \\
\textbf{Response}: This meme humorously depicts \textcolor{red}{two older men} in a conversation, with 
but the man on the right is playfully suggesting that the man on the left has a \textcolor{red}{lot of thoughts or ideas inside his head}, implying that he might be \textcolor{red}{overthinking} or need to calm down. \\
\textbf{Issue(s)}: It incorrectly labels the woman as a man and misinterprets the term `space' for overthinking rather than stupidity that generates the humor.
\\ \text{}\\
\textbf{Input Image}: Figure \ref{fig:sarcasm_annotation}(c) \\
\textbf{Input Query}: Explain the meme briefly. \\
\textbf{Response}: This meme features Grumpy Cat, a popular internet meme 
... implying \textcolor{blue}{a need for constant supervision or control} over one\'s thoughts. The overall tone of the meme is playful and slightly sarcastic.\\
\textbf{Issue(s)}: N/A
\end{tcolorbox}

\begin{tcolorbox}[breakable, colback=gray!10, colframe=black, boxrule=0.7pt, arc=3pt, left=6pt, right=6pt, top=6pt, bottom=6pt, fontupper=\small, colupper=black,
    ]
\textbf{Model}: GPT-4o
\\ \text{}\\
\textbf{Input Image}: Figure \ref{fig:sarcasm_annotation}(a) \\
\textbf{Input Query}: Explain the meme briefly. \\
\textbf{Response}: The meme jokes about a \textcolor{blue}{common restaurant question}—“Would you like a table?”—by responding with exaggerated sarcasm. The humor comes from mocking \textcolor{blue}{how unnecessary or obvious} the question seems, since people go to restaurants specifically to sit at tables, so the sarcastic response highlights the perceived redundancy of the server’s polite greeting.\\
\textbf{Issue(s)}: Although the model identifies that the primary humour arises from the obvious nature of the waitress’s question, it does not acknowledge the additional critical aspect where the man sarcastically refers to the carpet as if it were a food item. 
\\ \text{}\\
\textbf{Input Image}: Figure \ref{fig:sarcasm_annotation}(b) \\
\textbf{Input Query}: Explain the meme briefly. \\
\textbf{Response}: The meme uses a classical-style painting of two older \textcolor{red}{men} having a serious conversation. The joke plays on the \textcolor{blue}{double meaning} of “space”: the first person 
... person is \textcolor{blue}{empty-headed}. The humor comes from this clever wordplay and the unexpected insult delivered in a calm, refined setting.\\
\textbf{Issue(s)}: It fails to identify the woman in the context.
\\ \text{}\\
\textbf{Input Image}: Figure \ref{fig:sarcasm_annotation}(c) \\
\textbf{Input Query}: Explain the meme briefly. \\
\textbf{Response}: The meme features a grumpy-looking cat 
... from the sarcastic twist on the \textcolor{blue}{common phrase} “let your mind wander,” which usually encourages creativity or daydreaming. Instead, the joke implies the person isn’t very smart, suggesting \textcolor{blue}{their mind is “too small” to handle wandering by itself}.\\
\textbf{Issue(s)}: N/A
\end{tcolorbox}

\begin{tcolorbox}[breakable, colback=gray!10, colframe=black, boxrule=0.7pt, arc=3pt, left=6pt, right=6pt, top=6pt, bottom=6pt, fontupper=\small, colupper=black,
]
\textbf{Model}: Gemini 2.5
\\ \text{}\\
\textbf{Input Image}: Figure \ref{fig:sarcasm_annotation}(a) \\
\textbf{Input Query}: Explain the meme briefly. \\
\textbf{Response}: This meme uses sarcasm to mock a common, albeit redundant, question asked by restaurant hosts. The humor relies on the stating-the-obvious trope, where the customer highlights the \textcolor{blue}{silliness} of the initial question by suggesting a \textcolor{blue}{ridiculous alternative} to a standard dining setup.\\
\textbf{Issue(s)}: It correctly labels the obviousness of the question asked by the waitress but fails to highlight that the man sarcastically talks about the carpet as if it is something to eat.
\\ \text{}\\
\textbf{Input Image}: Figure \ref{fig:sarcasm_annotation}(b) \\
\textbf{Input Query}: Explain the meme briefly. \\
\textbf{Response}:   This meme relies on a deadly \textcolor{blue}{insult} delivered 
... implying that he is \textcolor{blue}{empty-headed or lacking in intelligence}. The humor comes from the savage subversion of a common serious request into a blunt critique of the man's intellect.\\
\textbf{Issue(s)}: N/A
\\ \text{}\\
\textbf{Input Image}: Figure \ref{fig:sarcasm_annotation}(c) \\
\textbf{Input Query}: Explain the meme briefly. \\
\textbf{Response}: This meme uses the legendary Grumpy Cat to deliver a 
...suggests that the \textcolor{blue}{target's intellect is so "small" and underdeveloped that it couldn't possibly survive "out on its own" without supervision}. \\
\textbf{Issue(s)}: N/A
\end{tcolorbox}

\subsection{\revised{Limited Ability of VLMs to Interpret Irony}}
\begin{figure}[!t]
\centering
\includegraphics[width=0.85\textwidth]{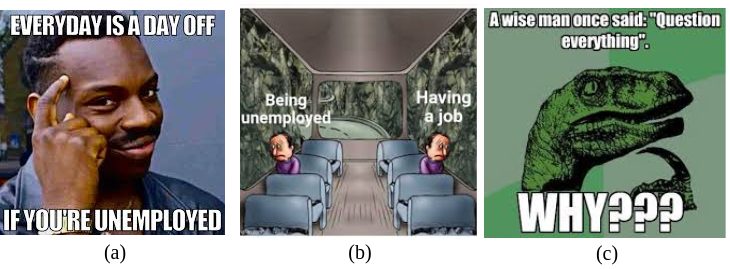} 
\caption{Examples of meme instances that contain irony within their context.}
\label{fig:irony_annotation}
\end{figure}

Here, we examine the ability of the previously introduced VLMs to understand contextual meaning in the presence of irony. Figures \ref{fig:irony_annotation}(a)-–(c) present the input memes used for evaluation. The corresponding generated outputs are shown below.

\begin{tcolorbox}[breakable, colback=gray!10, colframe=black, boxrule=0.7pt, arc=3pt, left=6pt, right=6pt, top=6pt, bottom=6pt, fontupper=\small, colupper=black,
]
\textbf{Model}:  llava-1.5-7b-hf\\

\textbf{Input Image}: Figure \ref{fig:irony_annotation}(a) \\
\textbf{Input  Query}: Explain the meme briefly. \\
\textbf{Response}: The meme features a man with a shaved head ... making a funny face. He is giving a \textcolor{red}{thumbs up} and appears to be smiling. The man is also \textcolor{red}{wearing a gold necklace} that  \textcolor{red}{adds to the humor} of the situation. The meme is captioned with a humorous message, suggesting that every day is a day off if you're unemployed.\\
\textbf{Issue(s)}: Model incorrect interprets the gesture and includes hallucinated detail. Moreover, the joke is not merely about unemployment being humorous; rather, it is based on flawed logical reasoning.
\\ \text{}\\
\textbf{Input Image}: Figure \ref{fig:irony_annotation}(b) \\
\textbf{Input Query}: Explain the meme briefly. \\
\textbf{Response}: The meme features a cartoon of \textcolor{red}{a man} sitting on a bus. The man appears to be sad or upset, possibly due to his unemployment. The image is a visual representation of \textcolor{red}{the emotional struggles that come with unemployment} and the desire to find a job.\\
\textbf{Issue(s)}: While it considers the emotion associated with unemployment, it ignores the emotional distress experienced even after having a job, the crux of the meme.
\\ \text{}\\
\textbf{Input Image}: Figure \ref{fig:irony_annotation}(c) \\
\textbf{Input Query}: Explain the meme briefly. \\
\textbf{Response}: The meme features a cartoon of a dinosaur with a question mark on its face, suggesting that it is asking a question. The image is captioned with a quote from a wise man, which reads, "A wise man once said, 'Question everything.'" \textcolor{red}{The meme is meant to encourage critical thinking and questioning} of assumptions. \\
\textbf{Issue(s)}: Model fails to recognize the recursive irony based humor.
\end{tcolorbox}

\begin{tcolorbox}[breakable, colback=gray!10, colframe=black, boxrule=0.7pt, arc=3pt, left=6pt, right=6pt, top=6pt, bottom=6pt, fontupper=\small, colupper=black,
]
\textbf{Model}: Qwen2.5-VL-7B-Inst 
\\ \text{}\\
\textbf{Input Image}: Figure \ref{fig:irony_annotation}(a) \\
\textbf{Input Query}: Explain the meme briefly. \\
\textbf{Response}: This meme humorously ... day off because \textcolor{blue}{they don\'t have to go to work}. The image features a person making a \textcolor{red}{"peace" sign with their hand}, which is often associated with a relaxed or carefree attitude. The text overlay adds a playful and relatable commentary on the challenges of unemployment.\\
\textbf{Issue(s)}: Model incorrectly describes the gesture as a “peace sign.” Additionally, it reduces the joke to a literal statement and fails to capture the core irony of the meme.
\\ \text{}\\
\textbf{Input Image}: Figure \ref{fig:irony_annotation}(b) \\
\textbf{Input Query}: Explain the meme briefly. \\
\textbf{Response}: This meme humorously illustrates the difference between being unemployed and having a job. It shows \textcolor{blue}{two identical} cartoon characters sitting in a bus, but \textcolor{blue}{with different captions} above them ... Despite the identical appearance of the characters, the context suggests that the character on the \textcolor{red}{right feels more comfortable or content, implying that having a job provides a sense of stability and security compared to being unemployed}.\\
\textbf{Issue(s)}: Model misinterprets the core message of the input meme by overemphasizing the literal expectation of job stability instead of recognizing the irony. 
\\ \text{}\\
\textbf{Input Image}: Figure \ref{fig:irony_annotation}(c) \\
\textbf{Input Query}: Explain the meme briefly. \\
\textbf{Response}: This meme features an image of Philosoraptor, a dinosaur with a human-like head and a thoughtful expression. The meme playfully challenges the idea of questioning everything by asking why one should question everything, implying that \textcolor{blue}{it might be unnecessary or even counterproductive to do so}.\\
\textbf{Issue(s)}: N/A
\end{tcolorbox}

\begin{tcolorbox}[breakable, colback=gray!10, colframe=black, boxrule=0.7pt, arc=3pt, left=6pt, right=6pt, top=6pt, bottom=6pt, fontupper=\small, colupper=black,
]
\textbf{Model}: GPT-4o
\\ \text{}\\
\textbf{Input Image}: Figure \ref{fig:irony_annotation}(a) \\
\textbf{Input Query}: Explain the meme briefly. \\
\textbf{Response}: This meme uses the popular image of a man ... paired with the caption “Everyday is a day off if you’re unemployed.” The humor comes from \textcolor{blue}{ironic logic}: while having every day off sounds positive ... is generally undesirable. The joke plays on \textcolor{blue}{twisting a negative situation} into something that sounds cleverly optimistic, highlighting sarcastic or flawed reasoning.\\
\textbf{Issue(s)}: N/A
\\ \text{}\\
\textbf{Input Image}: Figure \ref{fig:irony_annotation}(b) \\
\textbf{Input Query}: Explain the meme briefly. \\
\textbf{Response}: This meme shows two \textcolor{blue}{identical-looking sad} people sitting in the same gloomy bus. The joke suggests that \textcolor{blue}{both situations can feel equally miserable, implying that having a job doesn’t automatically guarantee happiness or satisfaction}. The humor comes from \textcolor{blue}{the irony that two seemingly opposite life situations are portrayed as emotionally the same}, poking fun at the idea that employment alone solves life’s problems.\\
\textbf{Issue(s)}: N/A
\\ \text{}\\
\textbf{Input Image}: Figure \ref{fig:irony_annotation}(c) \\
\textbf{Input Query}: Explain the meme briefly. \\
\textbf{Response}: This meme features the “Philosoraptor” character, a dinosaur ... philosophical questions. The joke plays on \textcolor{blue}{literal interpretation and self-reference—if you’re supposed to question everything, then you should also question the advice itself}. The humor comes from this paradoxical loop, highlighting how blindly following an advice can lead to absurd or endless questioning.\\
\textbf{Issue(s)}: N/A
\end{tcolorbox}

\begin{tcolorbox}[breakable, colback=gray!10, colframe=black, boxrule=0.7pt, arc=3pt, left=6pt, right=6pt, top=6pt, bottom=6pt, fontupper=\small, colupper=black,
    ]
\textbf{Model}: Gemini 2.5
\\ \text{}\\
\textbf{Input Image}: Figure \ref{fig:irony_annotation}(a) \\
\textbf{Input Query}: Explain the meme briefly. \\
\textbf{Response}: This meme features the popular "Roll Safe" character, known for ... a mock "brilliant" idea. The humor lies in the \textcolor{blue}{use of absurd or flawed logic to re-frame a negative situation as a positive one}. It suggests that the "benefit" of being unemployed is that you never have to work. By \textcolor{blue}{presenting a stressful life circumstance as a clever life hack}, the meme highlights the irony of finding a technical "win" in a situation that is actually quite difficult.\\
\textbf{Issue(s)}: N/A
\\ \text{}\\
\textbf{Input Image}: Figure \ref{fig:irony_annotation}(b) \\
\textbf{Input Query}: Explain the meme briefly. \\
\textbf{Response}: The image depicts two passengers sitting in a bus. Both sides ...
... By showing both characters looking equally miserable, the meme humorously suggests \textcolor{blue}{a "no-win" situation} where \textcolor{blue}{neither the stress of work nor the uncertainty of unemployment provides any happiness or relief}.\\
\textbf{Issue(s)}: N/A
\\ \text{}\\
\textbf{Input Image}: Figure \ref{fig:irony_annotation}(c) \\
\textbf{Input Query}: Explain the meme briefly. \\
\textbf{Response}: This meme features Philosoraptor ... paradoxical or philosophical questions. The humor here comes from  ... \textcolor{blue}{simultaneously following the advice and challenging its validity}. It captures the \textcolor{blue}{humorous irony} of a statement that, \textcolor{blue}{when acted upon, immediately targets itself}.\\
\textbf{Issue(s)}: N/A
\end{tcolorbox}

\subsection{\revised{Limitations of VLMs in Fine-Grained Question Answering}}

Here, we evaluate the ability of the previously introduced VLMs to understand contextual semantics and respond to queries. Figure \ref{fig:MHS_DBExample} presents the input memes used for evaluation, and the corresponding generated outputs are shown below,

\begin{tcolorbox}[breakable, colback=gray!5, colframe=black, boxrule=0.7pt, arc=3pt, left=6pt, right=6pt, top=6pt, bottom=6pt, fontupper=\small, colupper=black
]
\textbf{Input Image}: Figure \ref{fig:MHS_DBExample} \\
\textbf{Model}:  llava-1.5-7b-hf\\
 \text{}\\

\textbf{Input Query}: \textcolor{red}{What is the primary activity of the people in the image?} \\
\textbf{Response}:  \textcolor{red}{The primary activity of the people in the image is working on a farm, specifically cutting down sugar cane.}\\
\textbf{Issue(s)}: Irrelevant query-response pair.
\\ \text{}\\

\textbf{Input Query}: \textcolor{red}{What is the purpose of the people in the image?}\\
\textbf{Response}: \textcolor{red}{The image captures the hard work and teamwork required to cultivate and harvest this crop}.\\
\textbf{Issue(s)}: Irrelevant query-response pair.
\\ \text{}\\

\textbf{Input Query}: \textcolor{red}{What is the time of day in the image?} \\
\textbf{Response}: \textcolor{red}{The image shows a group of people working in a field during the day}. \\
\textbf{Issue(s)}:  Irrelevant query-response pair.
\\ \text{}\\

\textbf{Input Query}: \textcolor{red}{What is the weather in the image?} \\
\textbf{Response}: \textcolor{red}{The image shows ... it is not possible to determine the weather from the image alone}. \\
\textbf{Issue(s)}: Irrelevant query-response pair.
\\ \text{}\\

\textbf{Input Query}: What is the main focus of the image?\\
\textbf{Response}: The main focus of the image is a \textcolor{red}{group of people working together to cut down a large amount of sugar cane}. \\
\textbf{Issue(s)}: Fails to identify the hateful-humor w.r.t. the reference of slavery within the context.
\\ \text{}\\

\textbf{Input Query}: What is the context of the image? \\
\textbf{Response}: The context of the image is a painting or drawing ... harvesting sugar cane. The people are depicted as \textcolor{red}{hard-working farmers}, and the scene is described as a "best game of farming simulator." \\
\textbf{Issue(s)}: Fails to detect the hateful historical reference to slavery within the context.
\\ \text{}\\

\textbf{Input Query}: What is the historical context of the image?\\
\textbf{Response}: The historical context of the image could be related to the use of machetes as a tool for agriculture ... which could be \textcolor{red}{indicative of a simpler time or a region with limited access to modern technology}. \\
\textbf{Issue(s)}: Fails to detect the hateful historical reference to slavery within the context.
\\ \text{}\\

\textbf{Input Query}: What is the cultural significance of the image? \\
\textbf{Response}:  The image depicts ... depicted as African Americans, which highlights the racial discrimination and exploitation that occurred in the sugar industry during the time of slavery.\\
\textbf{Issue(s)}: N/A
\\ \text{}\\

\textbf{Input Query}: What is the intended emotion evoked by the image? \\
\textbf{Response}:  The image depicts ...  \textcolor{red}{The image may also evoke feelings of nostalgia or a sense of history, as it portrays a traditional or historical method of harvesting cane or grass}.\\
\textbf{Issue(s)}: Fails to detect the hateful humor of the context.
\\ \text{}\\

\textbf{Input Query}: What is the message behind the image? \\
\textbf{Response}:  The image depicts ... \textcolor{red}{The fact that the men are working together to accomplish a common goal highlights the value of cooperation and the role of manual labor in various industries}.\\
\textbf{Issue(s)}: Fails to detect the hateful historical reference of slavery within the context.
\end{tcolorbox}

\begin{tcolorbox}[breakable, colback=gray!10, colframe=black, boxrule=0.7pt, arc=3pt, left=6pt, right=6pt, top=6pt, bottom=6pt, fontupper=\small, colupper=black,
]
\textbf{Input Image}: Figure \ref{fig:MHS_DBExample} \\\textbf{Model}:  Qwen2.5-VL-7B-Inst
\\ \text{}\\

\textbf{Input Query}: What historical period does the image reference with 'Farming Simulator 1848'?\\
\textbf{Response}: The image references the historical period of the mid-19th century, specifically the year 1848.\\
\textbf{Issue(s)}: N/A
\\ \text{}\\
\textbf{Input Query}: Why might the depiction of people working in a field be considered problematic or offensive?\\
\textbf{Response}: The depiction of people working in a field can be considered problematic or offensive if it is used to make light of historical events or to trivialize the struggles faced by individuals during that time period. Such imagery can \textcolor{red}{perpetuate stereotypes and overlook the real-life hardships experienced by those who labored under oppressive conditions}.\\
\textbf{Issue(s)}: Fails to detect the hateful historical reference to slavery within the context.\\
\text{}\\
\textbf{Input Query}: What message or intent could be behind using the term 'Best Game' in relation to a historical farming scenario?\\
\textbf{Response}: The use of the term \'Best Game\' in relation to a historical farming scenario, as depicted in the image, may intend to highlight the contrast between \textcolor{red}{modern gaming experiences and the realities of historical agricultural labor. It could be interpreted as a critique of contemporary gaming for its lack of historical accuracy or sensitivity towards the conditions faced by farmers in the past}.\\
\textbf{Issue(s)}: Fails to detect the implicit hate.
\\ \text{}\\

\textbf{Input Query}: How does the juxtaposition of modern gaming terminology with historical imagery contribute to the overall message?\\
\textbf{Response}: The juxtaposition of ... \textcolor{red}{highlights the evolution of farming practices over time. This comparison can be seen as a commentary on how technology has transformed agricultural methods, suggesting that even in the past, farming was complex and required simulation-like strategies. It also serves to underscore the continuity of human endeavors in agriculture despite significant changes in technology and societal structures}.\\
\textbf{Issue(s)}: Fails to detect the implicit hate.

\end{tcolorbox}

\begin{tcolorbox}[breakable, colback=gray!10, colframe=black, boxrule=0.7pt, arc=3pt, left=6pt, right=6pt, top=6pt, bottom=6pt, fontupper=\small, colupper=black,
]
\textbf{Input Image}: Figure \ref{fig:MHS_DBExample} \\
\textbf{Model}:  GPT-4o
\\ \text{}\\
\textbf{Input Query}: What is the theme of the image?\\
\textbf{Response}: The image appears to be a \textcolor{red}{historical illustration with a humorous caption related to farming}.\\
\textbf{Issue(s)}: Fails to detect the implicit hateful humor of the context.\\
\text{}\\
\textbf{Input Query}: What caption is given to the image?\\
\textbf{Response}: The caption reads, "the best game of all time, farming simulator 1848".\\
\textbf{Issue(s)}: N/A\\ 
\text{}\\
\textbf{Input Query}: What is depicted in the illustration?\\
\textbf{Response}: The illustration seems to show people engaged in farming or agricultural work, possibly depicting a historical scene.\\
\textbf{Issue(s)}: N/A\\ 
\text{}\\
\textbf{Input Query}: How does the caption relate to the image?\\
\textbf{Response}: The caption humorously refers to the illustration as \textcolor{red}{an early or historical version of a "farming simulator" game}.\\
\textbf{Issue(s)}: Fails to identify the implicit hate in the context.\\ 
\text{}\\
\textbf{Input Query}: What might "farming simulator 1848" imply?\\
\textbf{Response}: It humorously suggests that the scene is an \textcolor{red}{old or original form of farming simulation, comparing real-life farming in 1848 to modern simulation games}.\\
\textbf{Issue(s)}: Fails to interpret the dark-humor behind the phrase.\\

\text{}\\
\textbf{Input Query}: Why is the caption considered humorous?\\
\textbf{Response}: The humor comes from juxtaposing a serious historical farming scene with the concept of a modern, entertaining video game simulator.\\
\textbf{Issue(s)}: N/A
\end{tcolorbox}

\begin{tcolorbox}[breakable, colback=gray!10, colframe=black, boxrule=0.7pt, arc=3pt, left=6pt, right=6pt, top=6pt, bottom=6pt,  fontupper=\small, colupper=black,
]
\textbf{Input Image}: Figure \ref{fig:MHS_DBExample} \\
\textbf{Model}:  Gemini 2.5\\
\text{}\\

\textbf{Input Query}: What is depicted in the central image?\\
\textbf{Response}: The central image shows four figures, appearing to be men, engaged in agricultural labor, possibly hoeing or scything in a field. They appear to be people of color, and the style suggests an older illustration or lithograph. One figure on the far right wears a wide-brimmed hat and a red shirt.\\
\textbf{Issue(s)}: N/A\\
\text{}\\
\textbf{Input Query}: What text is present in the image?\\
\textbf{Response}: At the top, the text reads "the best game of all time". At the bottom, it says "farming simulator 1848".\\
\textbf{Issue(s)}: N/A\\
\text{}\\
\textbf{Input Query}: What is the likely historical context suggested by the year '1848' and the imagery?\\
\textbf{Response}:  The year 1848, combined with the depiction of people of color engaged in intense manual agricultural labor, strongly suggests a historical context of forced labor or slavery. This was a period when slavery was still legal and widespread in many parts of the world, particularly in plantation economies of the Americas.
\\
\textbf{Issue(s)}: N/A\\
\text{}\\
\textbf{Input Query}:  What is the overall tone or intent of this image, considering all its elements?\\
\textbf{Response}: The overall tone is satirical and employs dark humor. It aims to create a provocative or shocking effect by juxtaposing a serious historical reality (forced labor/slavery) with the lighthearted, often enjoyable context of a modern video game simulator.\\
\textbf{Issue(s)}: N/A \\ 
\text{}\\
\textbf{Input Query}: What potential interpretations or controversial aspects might this image have? \\
\textbf{Response}: This image is highly controversial and can be interpreted as deeply offensive. It trivializes the immense suffering, dehumanization, and historical injustice associated with slavery and forced labor by turning it into a "game." \\
\textbf{Issue(s)}: N/A

\end{tcolorbox}

\begin{table}[t!]
\small
\centering
\caption{Common phrases or keywords for various protected groups in the $\dataone$ dataset.}
\label{tab:hate_category}
  
\begin{tabular}{p{3.5cm} p{12.5cm}}
\toprule
    \textbf{Protected Categories} & \textbf{Keywords} 
    \\ \midrule
    Islam & Islam, Prophet, Sheikh, Quran, hijab, burkha, burqa, Allah, Muhammad, Moslem, Buslim \\
    Black Community & Black, nigger, nigga, African, ape, Gorilla, slave, Cotton African-American, monkey\\
    Jewish Community & Adolf, Hitler, Anne, Frank, Jew, camp, gas, Nazi, Semitism, Poland, polish, concentration, bake, oven\\
    Disabile People & Disable, down, autistic, autism, midget, mental, disorder, mute, blind, syndrome, deaf, amputee, dwarf\\
    Women in general & Woman, women, girl, female, misogyny, pussy, prostitute, babe, blonde, dishwasher,  sandwich,  blowjob, bj\\
    Sexual orientation & Trans, gay, lesbian, homosexual, homo, sexuality, twisted sex, LGBT, tranny, queer  \\
    Immigration & Immigration, Mexico, Mexican, Hispanic, refugee, xenophobia\\
    White community & crack, caucasian \\
    Miscellaneous & Hinduism: India, Hindu. Christianity: Jesus, Christ, Church, pope\\
    \bottomrule
\end{tabular}%
\end{table}
\section{ Construction of Datasets } \label{app:dataset_construction}
The construction and annotation process of $\dataone$ and $\datatwo$ spans over \textit{four} distinct phases: (i) Data Acquisition and Pre-processing, (ii) Annotator Onboarding and Calibration, (iii) Annotation-on-call, and (iv) Quality Assurance and Consolidation.

\paragraph{ \textbf{Step 1}: Data Acquisition and Pre-processing.}

Initially, we marked a total of \textit{nine} categories of protected groups in the  $\dataone$ dataset. Table \ref{tab:hate_category} illustrates the common keywords or phrases associated with each category. Keywords are collected from observations made during the preliminary annotation phases. The corresponding meme instances for $\dataone$ dataset are primarily gathered from \textit{Google Image} and web-scraping the online platforms - \textit{Instagram}\footnote{\url{https://www.instagram.com/}} and \textit{Reddit}\footnote{\url{https://www.reddit.com}}. Next, we retain only one instance from the set of duplicates using dupeguru\footnote{\url{http://github.com/arsenetar/dupeguru}}. In addition, we filter out memes that have at least one of the following characteristics: (a) unclear or low-quality background images, (b) the overlaid text is in a non-English language, and (c) the content is unimodal, that is, either textual or visual. Next, the textual content of each meme was extracted using an open-source Optical Character Recognition (OCR) tool, Tesseract\footnote{\url{https://github.com/tesseract-ocr/tesseract}} followed by a manual correction by annotators w.r.t. the original memes to ensure textual accuracy.

\begin{figure}[!t]
\centering
\includegraphics[width=1\textwidth]{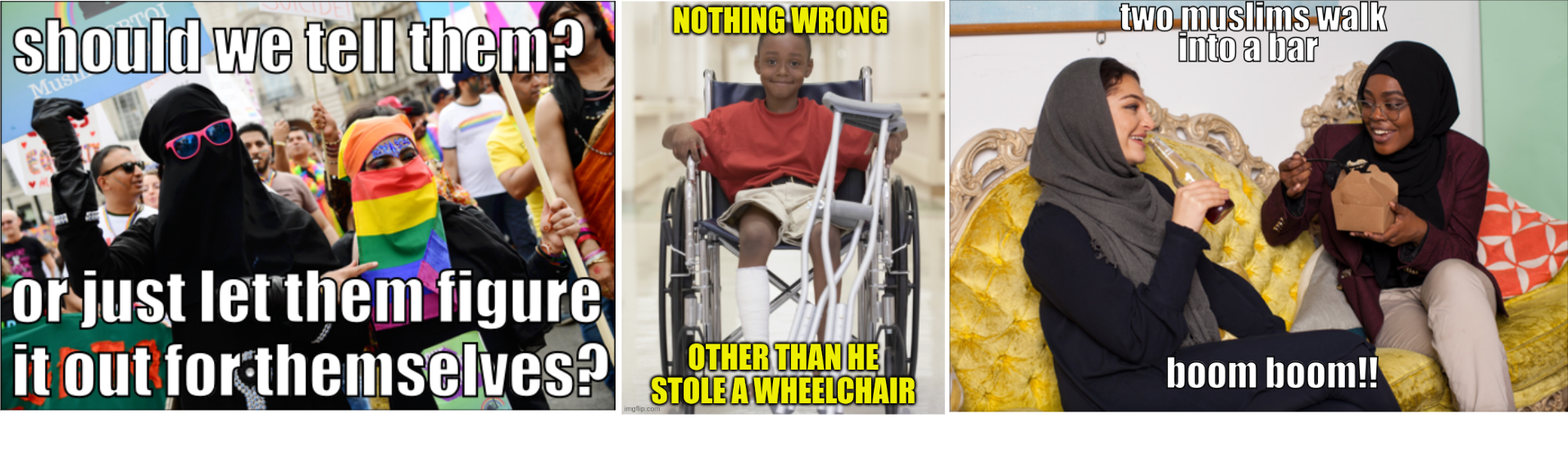} 
\caption{An illustration of instances involving multiple protected groups: In subfigure (a) both \textit{Islam} and the \textit{LGBTQ} community are visually represented; however, the context considers \textit{Islam} as the primary target. Subfigure (b) depicts both \textit{African-American} and \textit{Disabled} identities, yet the derogatory implication is directed mainly to the \textit{African-American} individual. Lastly, in subfigure (c) although both \textit{Women} and individuals associated with \textit{Islam} are featured, the context reveals that \textit{Islam} is the focal point of the hateful context.}
\label{fig:ambiguous_target}
\end{figure}

\begin{figure}[!h]
\centering
\includegraphics[width=1\textwidth]{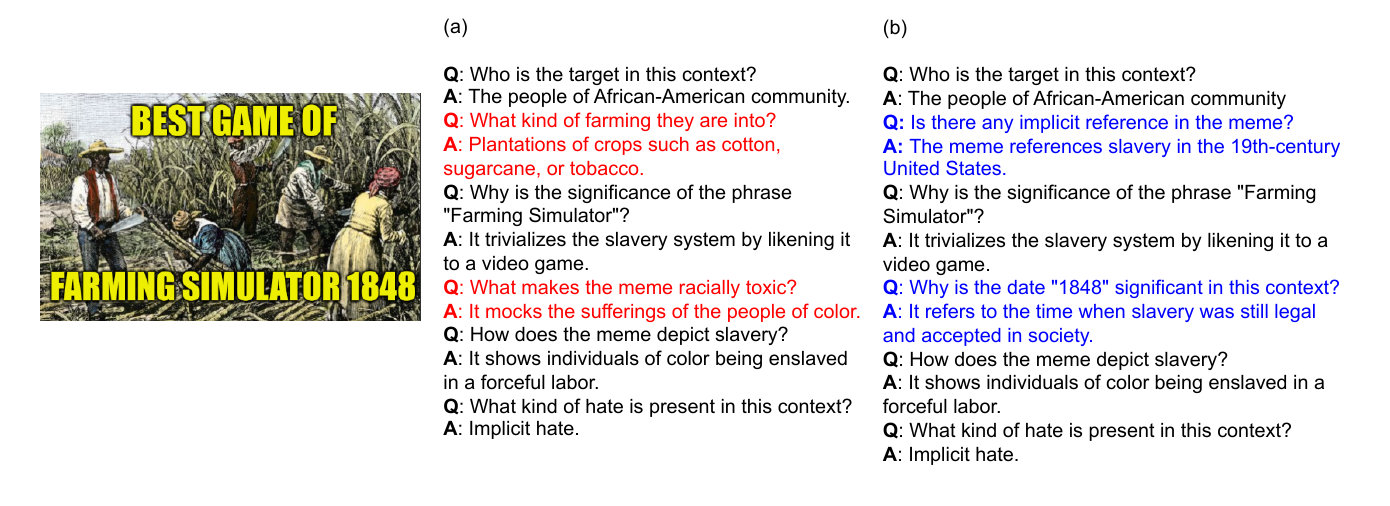} 
\caption{An illustration of the Q\&A annotation process. Subfigure (a) shows Q\&A pairs highlighted in \textcolor{red}{red} and black, indicating disagreements and agreements, respectively, among annotators regarding the interpretation of the meme. Subfigure (b) displays Q\&A pairs highlighted in \textcolor{blue}{blue}, representing additional relevant suggestions from other annotators that are subsequently verified and approved by the original annotator. Note that the Q\&A pairs shown in black are the undisputed ones.}
\label{fig:annotation_process}
\end{figure}
\paragraph{ \textbf{Step 2}: Annotator Onboarding and Calibration.}
In this phase, a sample set of pre-annotated memes, along with the corresponding annotation guidelines are given to the annotators in order to familiarize themselves with the task and preliminary insights. The pre-annotated dataset is curated by the in-house research team, having \textit{five} instances from explicit hate, implicit hate, and benign categories. According to the annotation guidelines, a label is considered explicit if the textual or visual content conveys hate directly and unambiguously towards a target.  In contrast, a label is implicit when the detection of hate requires background knowledge and a careful integration of multiple pieces of distinct information. Lastly, a label is benign if there is no hate. The annotators are asked to identify the primary target group in the presence of multiple protected groups. Figure \ref{fig:ambiguous_target} presents an example potentially with multiple targeted groups. In Figure \ref{fig:ambiguous_target}.a, both \textit{Islam} and the \textit{LGBTQ} community are depicted; however, the context targets \textit{Islam}. Similarly, Figure \ref{fig:ambiguous_target}(b) features representations of both the \textit{African-American} and \textit{Disabled} communities, yet the \textit{African-American} community is the focus of the hateful context. In Figure \ref{fig:ambiguous_target}(c), while both \textit{Women} and \textit{Islam} are referenced, the hateful content is directed against \textit{Islam}. In case of dispute regarding the selection of the primary target group, the group with maximum support is retained.

\paragraph{\textbf{Stage 3: } Annotation-on-call.}
In this stage, the annotators are provided with both the OCR-extracted text and the corresponding meme instances to label each of them into one of the following \textit{three} categories: explicit, implicit, or benign. Moreover, annotators are tasked to provide explanatory insights for each meme in the form of question-and-answer (Q\&A) pairs. The group of annotators comprised \textit{three} individuals aged between $25$ and $35$ with proficiency in English. Annotators were employed based on their demonstrated efficiency in English and prior experience in content moderation. To ensure balanced coverage across all categories, each of the annotators was assigned \textit{one-third} of the meme instances from each category. While annotators could directly assign a category label, the instructions for Q\&A pairs required annotators to formulate them in a meaningful way that simultaneously reveals the underlying context. For example, if an individual of a particular ethnicity is referred to using a derogatory term such as `old farm equipment', the corresponding question-and-answer (Q\&A) pairs are expected to explain both the intended target and the relevant contextual or historical background.

\paragraph{\textbf{Stage 4: } Quality Assurance and Consolidation.}
Upon completion, each of them received $50$ randomly selected samples from the other two annotators for cross-verification. In this phase, the instruction was to verify both the assigned labels and associated Q\&A pairs. The inter-annotator agreement is assessed with Cohen’s Kappa - the resulting average Kappa scores are $0.76$, $0.71$, $0.85$ for explicit hate, implicit hate and benign categories, respectively.

In case of a disagreement for a category label, we opted for the majority voting strategy to select a label. Most of the disagreements are reported in the implicit hate category. In a harmful instance where subtle hate is present but lacks any explicit hint or a scope of in-depth reasoning, this caused disagreement between annotators. For the Q\&A pairs, we filter out pairs that are marked as unnecessary, irrelevant or disagreed. Please note that only undisputed pairs are preserved.
 Figure \ref{fig:annotation_process} demonstrates an example of the Q\&A annotation process. The Q\&A pairs marked with \textcolor{red}{red} colour in Figure \ref{fig:annotation_process} (a) represent instances where annotators disagreed in their interpretations of the corresponding meme. Conversely, the Q\&A pairs highlighted with \textcolor{blue}{blue} colour in Figure \ref{fig:annotation_process} (b) reflect additional relevant suggestions from other annotators that are subsequently approved by the original annotator. Note that the Q\&A pairs in black colour are the ones that all annotators agreed on.

\section{Methodology}

\begin{figure}[!t]
\centering
\includegraphics[width=1\textwidth]{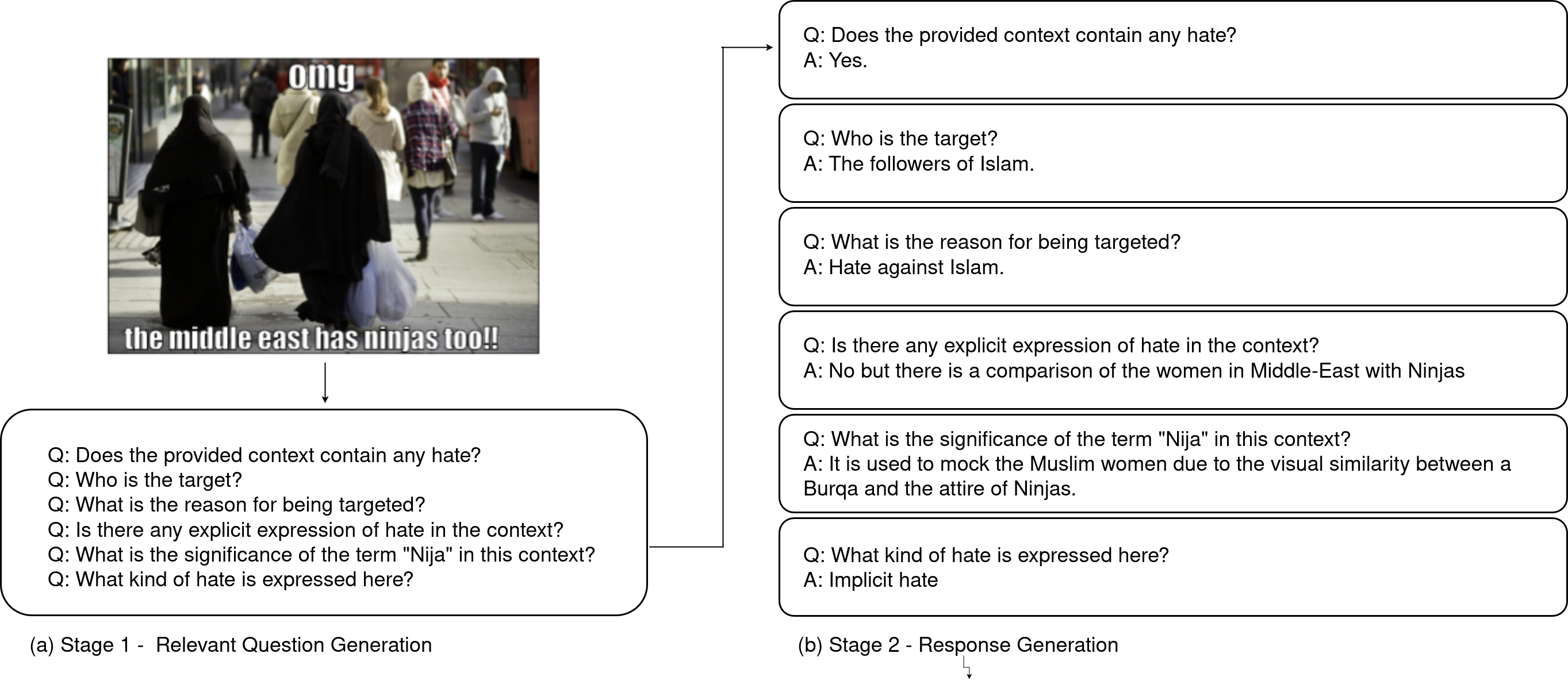} 
\caption{A visual illustration of each intermediate stage in $\fvariant$. Subfigure (a) illustrates the first stage, Relevant Question Generation (RQG): $\fvariant$ formulates context-revealing and critical queries essential for subsequent analysis. Subfigure (b) demonstrates the second stage, Response Generation (RG), in which each question generated during the RQG stage is addressed sequentially.}
\label{fig:SAFE_MEME_QA_execution}
\end{figure}

\subsection{Additional details on $\fvariant$} \label{app:safeMemeQA}
In this section, we discuss the intermediate stages of $\fvariant$ with examples. There are primarily two distinct stages in $\fvariant$: (a) Relevant Question Generation (RQG) and (b) Response Generation (RG).

\paragraph{\textbf{Stage 1}: Relevant Question Generation} This stage is tasked with generating questions that are contextually relevant. It takes the image, $I$ and the overlaid text, $T$, as inputs and generates a set of questions, $Y_{query}$. Figure \ref{fig:SAFE_MEME_QA_execution}(a) illustrates such an example. The input context satirically portrays women of Islamic faith wearing black burqas through a comparison with ninjas, thereby invoking a stereotypically and culturally biased analogy. In this scenario, RQG generates the following questions: (i) Does the provided context contain any hate?, (ii) Who is the target? (iii) What is the reason for being targeted?, (iv) Is there any explicit expression of hate in the context?, (v) What is the significance of the term "Nija" in this context?, (vi) What kind of hate is expressed here? Appropriate responses to each question will reveal the hidden link between the burqa and the ninja, identify the target group of the sarcasm, and classify the type of hate.

\paragraph{\textbf{Stage 2}: Response Generation} This stage is responsible for generation of response for each of the questions in $Y_{query}$, generated in Stage 1 i.e. in RQG. It takes the image $I$ , the overlaid text $T$ and a \textit{single} query $y_{i}$ from $Y_{query}$ to generate the corresponding response. The expected behavior of RG is illustrated in Figure \ref{fig:SAFE_MEME_QA_execution}(b). RG generates responses sequentially with respect to each corresponding query $y_{i}$.

\begin{figure}[!t]
\centering
\includegraphics[width=1\textwidth]{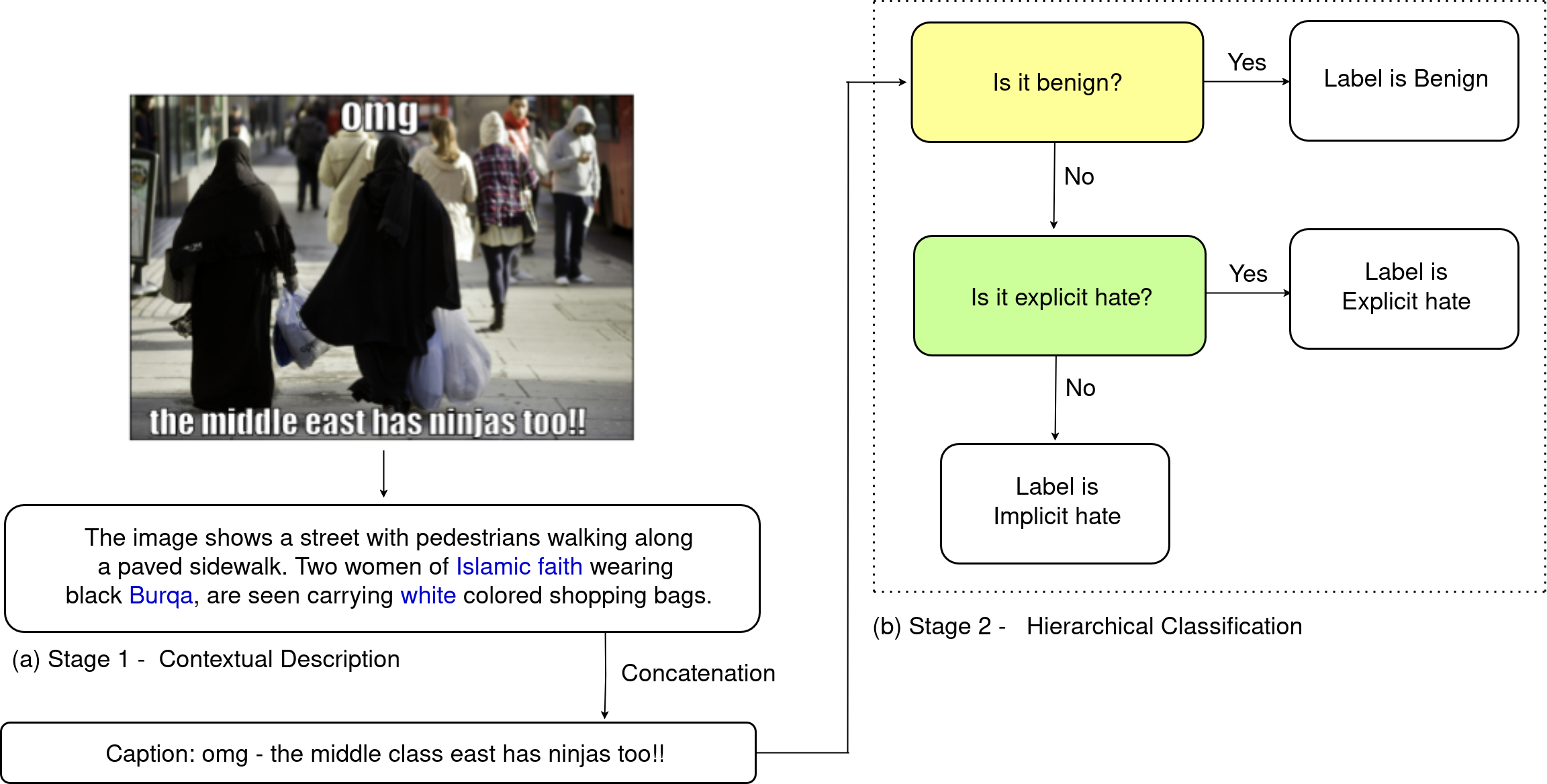} 
\caption{A visual illustration of each intermediate stage in $\svariant$. Subfigure (a) illustrates the first stage,  Contextual Description (CG): $\svariant$ formulates context-revealing description. Note that the critical contextual cues generated in CG are highlighted in \textcolor{blue}{blue}. Subfigure (b) demonstrates the second stage,  Hierarchical Classification (HC), in which the contextual description from the previous stage is appended with the image caption and forwarded to the classifiers. The first classifier (marked in \textcolor{good_yellow}{\textbf{yellow}}) determines whether the input exhibits hateful content. If it is hateful, then a second classifier (marked in \textcolor{good_green}{\textbf{green}}) further categorizes it as either explicit or implicit hate.}
\label{fig:SAFE_MEME_H_execution}
\end{figure}

\subsection{Additional details on $\svariant$} \label{app:safeMemeH}

In this section, we explain the stages of $\svariant$ with examples. There are two stages in $\svariant$: (a)  Contextual Description Generation (CDG) and (b) Hierarchical Classification (HC).

\paragraph{\textbf{Stage 1}: Contextual Description Generation} This stage of $\svariant$ generates a comprehensive contextual description of the input meme. It requires the meme, $I$, along with its overlaid text, $T$ to produce a brief but useful contextual description, $Y_{\text{desc}}$, containing significant contextual cues. Figure \ref{fig:SAFE_MEME_H_execution}(a) illustrates an example in which the output indicates that the women are wearing black burqas and belong to the Muslim community. The input prompt used in Stage 1 is provided below.

\begin{tcolorbox}[enhanced, colback=gray!10, colframe=black, boxrule=0.7pt, arc=3pt, left=6pt, right=6pt, top=6pt, bottom=6pt]
\revised{\textbf{Input Prompt}: Generate a contextual description for the input image.}
\end{tcolorbox}

\paragraph{ \textbf{Stage 2}: Hierarchical Classification} This stage utilizes a hierarchical classification approach: initially, it identifies if the input instance is hateful; if classified as hateful, it further categorizes it as either implicit or explicit hate. If the instance is classified as benign, no further classification is performed. Both classifiers require the contextual description $Y_{\text{desc}}$ from the previous stage to be concatenated with $T$ to generate an output label (cf. \ref{fig:SAFE_MEME_H_execution}(b)). The input prompts used in both levels of classification are provided below,

\begin{tcolorbox}[breakable, colback=gray!10, colframe=black, boxrule=0.7pt, arc=3pt, left=6pt, right=6pt, top=6pt, bottom=6pt, fontupper=\small, colupper=black]
\revised{\textbf{For First-Level Classification} \\
\text{} \\
\textbf{Input Prompt}: Is the input instance hateful or benign? Response HATEFUL if it contains hate, else BENIGN w.r.t. the given caption and image description. \\
\textbf{Image Caption}: [CAPTION] \\
\textbf{Image Description}: [GDESC] \\
\text{} \\
\textbf{For Second-Level Classification} \\
\text{}\\
\textbf{Input Prompt}: What kind of hate content is present in the context w.r.t. the given caption and image description. \\
\textbf{Image Caption}: [CAPTION] \\
\textbf{Image Description}: [GDESC]} \\
\end{tcolorbox}

\section{Additional Details on Variants of $\model$}\label{app:modelAblation_onMHS}

\noindent \paragraph{\textbf{On \dataone}} As shown in Table \ref{tab:benchmarkComparison_FVandSVonMHS_DB}, the M7 variant of $\fvariant$ outperforms other variants. It scores $0.494$, $0.571$, and $0.554$ in \textit{explicit}, \textit{implicit}, and \textit{benign} categories, respectively. In \textit{explicit hate}, M7 has a minimmum lead of $2.2\%$ (against M1 and M6) and a maximum lead of $5.9\%$ (against M0). Similarly, for \textit{implicit hate}, M7 has a minimum lead of $3.9\%$ (against M2) and a maximum lead of $12.3\%$ (against M6). A slightly different scenario is observed in the \textit{benign} category, with M3 achieving the highest score of $0.557$. However, M7 exhibits only a minimal lag of $0.3\%$. In the overall assessment, M7 scores $0.539$, which creates a minimum lead of $4.7\%$ (against M6) and a maximum lead of $11.4\%$ (against M0).

Among the variants of $\svariant$, M11 with an F1 score of $0.546$ outperforms others. Interestingly, M11 demonstrates performance that is either superior to or comparable with its fully fine-tuned counterpart, M10. For example, in the \textit{explicit hate} category, M11 scores $0.587$ while M10, M12, and M13 are able to score $0.494$, $0.53$, and $0.531$; M11 is ahead of M10 by $9.3\%$. For the \textit{implicit hate} category, M9 is superior to others with an F1 score of $0.696$. M9 is ahead of M10 and M11 by $6\%$ each, as they have the same score of $0.636$. For the \textit{benign} category, while M12 achieves the highest score of $0.426$, M10 and M11 trail by margins of $6.3\%$ and $0.9\%$, with a score of $0.363$ and $0.417$; notice that M11 demonstrates a $5.4\%$ improvement over M10.

\noindent \paragraph{\textbf{On \datatwo}} Here, we discuss the intra-variant performance corresponding to $\fvariant$ and $\svariant$ on \datatwo\. Among the variants of $\fvariant$, the performance metrics for models M1, M3, and M7 are analyzed. Across all three categories, M7 demonstrates superior performance compared to M1 and M3. Specifically, for the \textit{explicit hate} category, M1 and M3 achieve scores of $0.553$ and $0.513$, respectively, while M7 attains a higher score of $0.602$. In the \textit{implicit hate} category, M1 and M3 show comparable performance with scores of $0.412$ and $0.411$, respectively, whereas M7 outperforms them with a score of $0.499$. For the \textit{benign} category, M7 achieves a score of $0.757$, surpassing the scores of M1 and M3, which are $0.685$ and $0.604$, respectively. Overall, M7 surpasses M1 and M3 by margins of $5.64\%$ and $11\%$, respectively.

In the case of $\dataone$, it is observed that M11 outperforms outperforms M10, its fully fine-tuned T5-base counterpart. Similar trends are evident in the case of $\datatwo$. M11 demonstrates superior performance compared to M10 across all categories except for \textit{explicit hate}. In the \textit{explicit hate} category, M11 achieves a score of $0.459$, whereas M10 attains $0.572$. Conversely, in other categories (\textit{implicit} and \textit{benign}), M11 attains scores of $0.488$ and $0.548$, respectively, outperforming M10, which scores $0.44$ and $0.36$, respectively. Regarding overall assessment (macro F1) M10 lags by $4.1\%$. M13 also demonstrates a $2.8\%$ performance improvement over its fully fine-tuned counterpart, M12.

\section{Additional Analysis of T5-large (FT)'s performance on $\datatwo$}\label{app:analysisOfT5_onMHS}

To analyze the competitive performance of the text-only T5-large (FT) baseline on $\datatwo$, we present a qualitative error analysis across three distinct scenarios. The purpose of this analysis is to understand the model’s decision-making process and to identify whether its predictions are based on genuine linguistic understanding or on spurious correlations. Here, we present three types of observations: instances where (i) the model performs well without requiring visual context, (ii) it fails due to lexical bias towards certain demographic groups, leading to incorrect classification, and (iii) it correctly identifies the sentiment and classifies it. Through this analysis, we highlight why the text-only model, T5-large (FT), performs competitively on confounded memes.

\begin{figure}[!t]
\centering
\includegraphics[width=\textwidth]{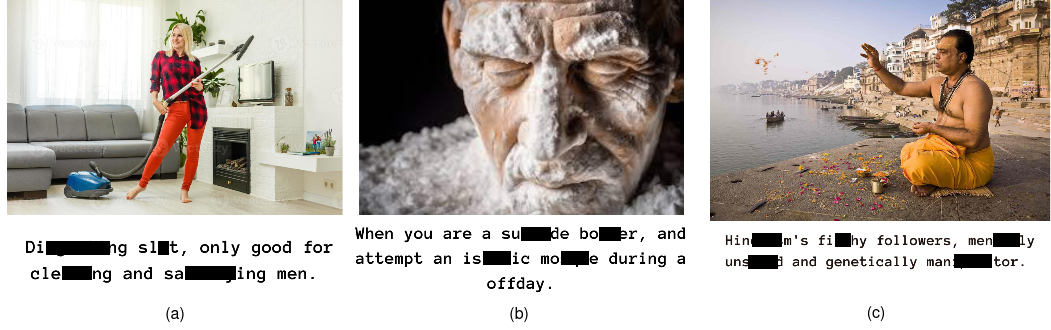} 
\caption{Sample test outcomes where the baseline model, T5-large (FT), correctly classifies the instances as explicit hate. In all three cases Fig \ref{fig:t5OnEEFromMHSConf}(a–c), the textual content contains clear and direct offensive language that enables T5-large (FT) for competitive predictions without any cues from visual context. For ethical considerations, sensitive offensive expressions and target group names have been masked in the figure.}
\label{fig:t5OnEEFromMHSConf}
\end{figure}

In the explicit category, T5-large (FT) achieves a precision of 0.625, a recall of 0.672, and an F1-score of 0.648, a fairly balanced performance between precision and recall (cf. Table \ref{tab:benchmarkComparison_MHSDBConandBaseline}). It suggests that the model has a reasonable semantic understanding of explicit hate. Figure \ref{fig:t5OnEEFromMHSConf}(a–c) shows instances that are  correctly predicted by T5-large (FT). The text in these examples contains direct slurs and aggressive expressions, which are inherently offensive, irrespective of the visual context. Since the textual content in the explicit hate instances has a high density of offensive language, it becomes sufficiently informative for T5-large (FT) to make competitive predictions without relying on visual inputs.

\begin{figure}[!t]
\centering
\includegraphics[width=0.9\textwidth]{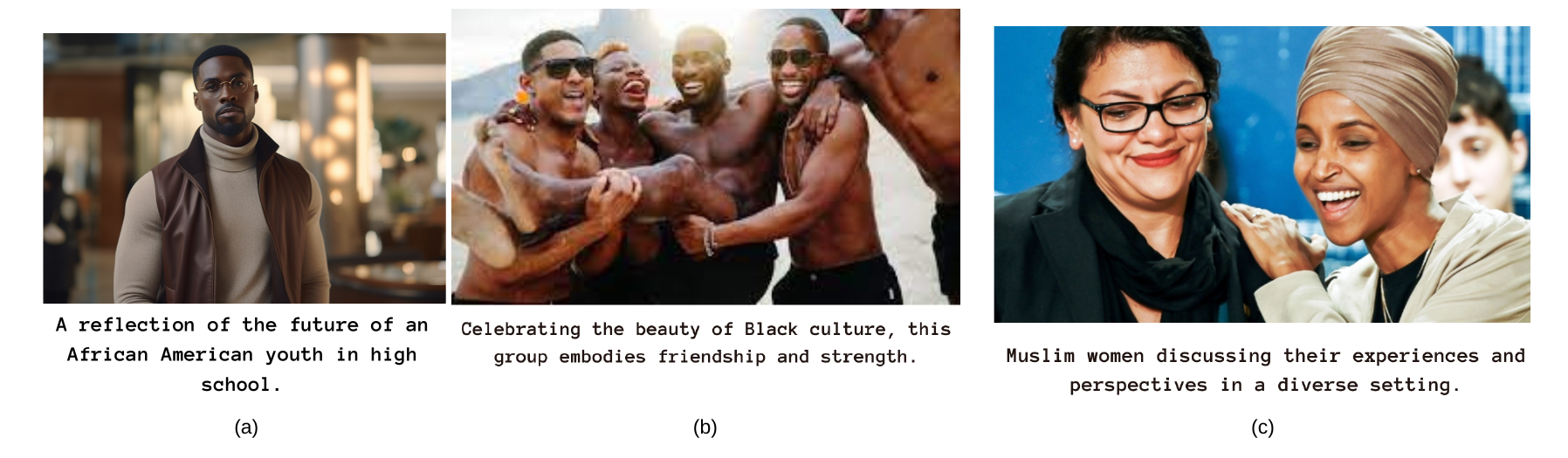} 
\caption{\revised{Sample test outcomes where the baseline model, T5-large (FT), incorrectly classifies benign instances as implicit hate. In all three cases of Fig. \ref{fig:t5OnBIFromMHSConf}(a–c), the content is non-offensive and contextually neutral or positive. However, the presence of target group identifiers or group-related associations leads T5-large (FT) to misinterpret the context as potentially hateful. It highlights  a high sensitivity of T5-large(FT) to group mentions and its tendency to over-predict implicit hate. }}
\label{fig:t5OnBIFromMHSConf}
\end{figure}
\revised{In all three cases shown in Fig. \ref{fig:t5OnBIFromMHSConf}(a–c), the content is non-offensive and contextually neutral or positive. However, the presence of target group identifiers or group-related references leads T5-large(FT) to incorrectly interpret the content as potentially hateful. This indicates that the model is highly sensitive to mentions of social groups and tends to over-predict implicit hate. The misclassification in Fig. \ref{fig:t5OnBIFromMHSConf}(b) further shows a clear lexical bias in the T5-large(FT) model. The mere presence of demographic terms such as “African American,” “Black,” and “Muslim” shifts the prediction toward Implicit Hate, even when the overall sentiment of the text is clearly positive. It results in a higher number of false positives, which lowers the precision (0.441). Meanwhile, this sensitivity towards implicit hate enables the model to identify a large portion of actual implicit hate instances, contributing to a relatively higher recall (0.602). Consequently, the model demonstrates a tendency to over-predict implicit hate, as reflected in the moderate F1-score (0.509)}.

\begin{figure}[!t]
\centering
\includegraphics[width=\textwidth]{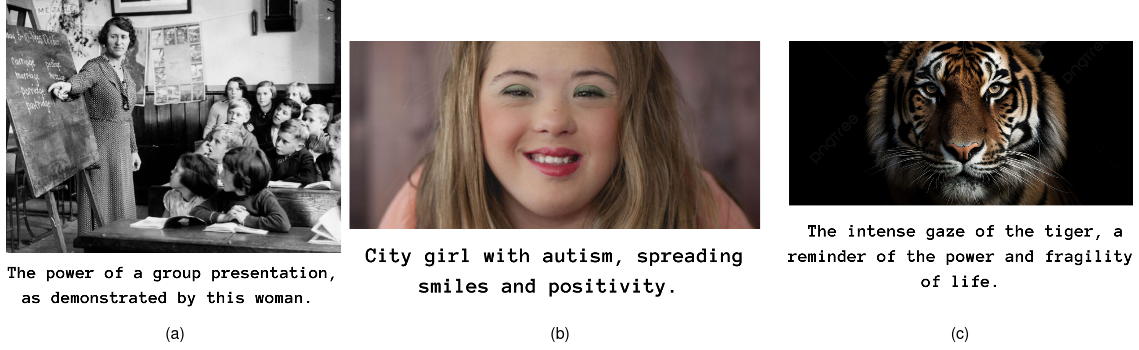} 
\caption{Sample test outcomes where the baseline model, T5-large (FT), correctly classifies the instances as benign. In all three cases of Fig. \ref{fig:t5OnBBFromMHSConf}(a–c), the content does not include any target group references or associations that could be interpreted as hate-related. Instead, the examples convey neutral or strongly positive context, allowing the model to make correct predictions.}
\label{fig:t5OnBBFromMHSConf}
\end{figure}

The examples in Figure \ref{fig:t5OnBBFromMHSConf}(a–c) show that T5-large (FT) correctly classifies clearly benign instances, particularly when target group references or associations are absent. The content in these cases is either neutral or strongly positive, which helps the model make accurate predictions. However, the model appears to exhibit selective behavior for benign category, performing well only when such explicit cues are absent. This leads to a relatively high precision (0.824) but a low recall (0.447), indicating that many benign instances are not correctly identified. As a result, the overall F1-score remains moderate (0.580).

\end{document}